\documentclass[10pt,twocolumn,letterpaper]{article}

\usepackage{iccv}
\usepackage{times}
\usepackage{epsfig}
\usepackage{graphicx}
\usepackage{amsmath}
\usepackage{amssymb}
\usepackage{enumitem}

\usepackage[vlined, ruled, linesnumbered]{algorithm2e}
\usepackage[justification=centering]{subfig}

\usepackage[breaklinks=true,bookmarks=false]{hyperref}
\usepackage[margin=4pt,font=footnotesize,labelfont=bf,labelsep=endash,tableposition=top]{caption}

\def\cI{{\mathcal I}}

\def\cL{{\mathcal L}}

\def\mbR{{\mathcal R}}
\def\bX{{\mathbf X}}
\def\bY{{\mathbf Y}}

\def\cI{{\mathcal I}}

\def\cL{{\mathcal L}}
\def\cF{{\mathcal F}}
\def\cC{{\mathcal C}}
\def\cD{{\mathcal D}}

\def\mbE{{\mathbb{E}}}

\def\mbR{{\mathbb{R}}}

\iccvfinalcopy 

\def\iccvPaperID{****} 

\ificcvfinal\fi
\begin{document}

\title{Regularizing Proxies with Multi-Adversarial Training for Unsupervised Domain-Adaptive Semantic Segmentation}

\author{Tong Shen$^1$, Dong Gong$^2$, Wei Zhang$^1$, Chunhua Shen$^2$, Tao Mei$^1$\\
$^1$JD AI Research, China\\
$^2$The University of Adelaide, Australia\\
{\tt\small tshen.st@outlook.com, edgong01@gmail.com, wzhang.cu@gmail.com}\\
{\tt\small chunhua.shen@adelaide.edu.au, tmei@live.com}
}

\maketitle

\begin{abstract}
Training a semantic segmentation model requires a large amount of pixel-level annotation, hampering its application at scale.
With computer graphics, we can generate almost unlimited training data with precise annotation. However, a deep model trained with synthetic data usually cannot directly generalize well to realistic images due to domain shift. 

It has been observed that highly confident labels for the unlabeled real images may be predicted relying on the labeled synthetic data. To tackle the unsupervised domain adaptation problem, we explore the possibilities to generate high-quality labels as proxy labels to supervise the training on target data. Specifically, we propose a novel proxy-based method using multi-adversarial training. We first train the model using synthetic data (source domain). Multiple discriminators are used to align the features between the source and target domain (real images) at different levels. Then we focus on obtaining and selecting high-quality proxy labels by incorporating both the confidence of the class predictor and that from the adversarial discriminators. Our discriminators not only work as a regularizer to encourage feature alignment but also provide an alternative confidence measure for generating proxy labels. Relying on the generated high-quality proxies, our model can be trained in a ``supervised manner" on the target domain. On two major tasks, GTA5 $\rightarrow$ Cityscapes and SYNTHIA $\rightarrow$ Cityscapes, our method achieves state-of-the-art results, outperforming the previous by a large margin.
\end{abstract}

\section{Introduction}

\begin{figure}[t]
\centering
   \includegraphics[width=1.0\linewidth]{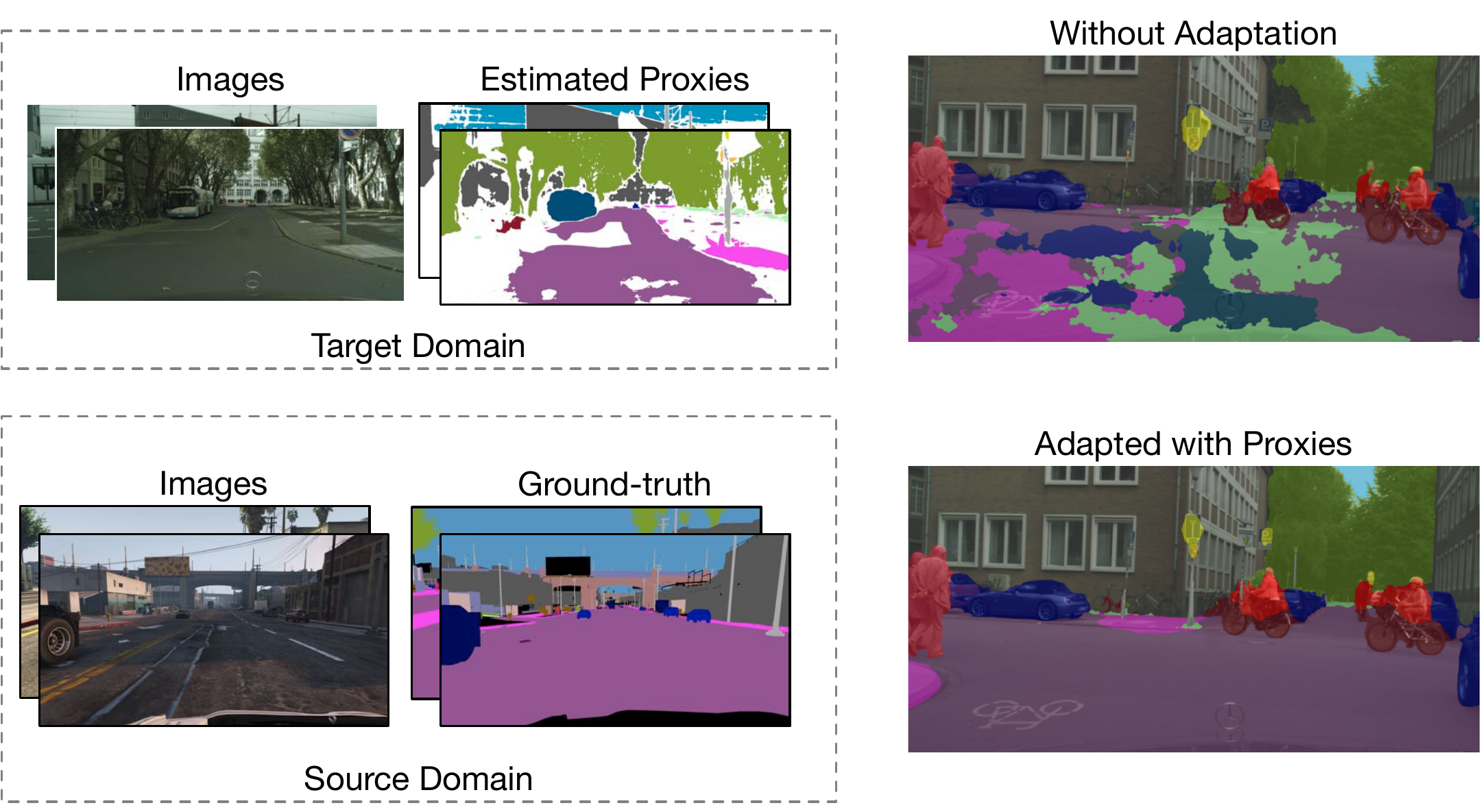}
   \caption{The ground-truth is provided in the source domain while there is no annotation in the target domain. Our proposed method can generate high-quality proxy labels for the unlabeled target data. As shown in the right, adapted with the proxy labels, the model can produce  more reliable and consistent predictions in the target domain.}
\label{fig:intro}
\end{figure}

Semantic segmentation aims to predict a dense labelling map for the input image by assigning each pixel a categorical label. It is crucial for image understanding and many specific applications such as autonomous driving \cite{geiger2012we}. Benefiting from intensively  annotated training samples, supervised deep neural networks have shown 
success 
for semantic segmentation \cite{long2015fully,refinenet, deeplabv3p, psp}. However, obtaining high-quality pixel-wise annotations can be very time-consuming and costly. 

\par
To greatly reduce the cost of data annotation, simulated scenes are used to automatically generate annotated synthetic datasets (\eg, GTA5 \cite{richter2016playingGTA} and SYNTHIA \cite{ros2016synthia}) for semantic segmentation. 
Although the enormous synthetic datasets can thoroughly cover diverse scenes and objects,  models trained on them 
are not able to generalize well to 
real-world data due to the domain shift between the source (synthetic) data and the target (real) data \cite{zhang2017curriculum}. 


\par
Unsupervised domain-adaptive semantic segmentation methods are investigated towards learning a model from the labelled \emph{source} data that can freely adapt to the novel unlabelled \emph{target} domain. Some general approaches obtain domain invariant representations by minimizing some measure of the domain shift \cite{sun2016return,tzeng2014deep,tzeng2017adversarial}. Inspired by this, apart from the pixel-wise semantic labeling loss on source data, a series of domain-adaptive semantic segmentation methods \cite{fcan, fcnwild, Tsai_adaptseg_2018} seek to minimize the feature discrepancy between  both domains via confusing an adversarial domain discriminator that is trained to distinguish the two domains. 
Although the adversarial training methods have achieved impressive results \cite{fcan}, they 
neglect 
the pixel-wise categorical particularity in the target domain while aligning the global distributions of the two domains. 

\par
The other orthogonal line of methods \cite{cbst,chitta2018adaptive} seek to assign \emph{proxy labels} (\ie, pseudo labels) to the unlabeled target images using the network trained on the source data, referred to as \emph{proxy-label} (or self-training) methods \cite{ruder2018strong}. 
Unlike the global adversarial adapter \cite{fcan,tzeng2017adversarial}, proxy labels can
offer 
explicit pixel-wise semantic supervision to the unlabelled target domain images. 
Proxies for the target data are usually generated according to the prediction of the cross-domain trained networks, which may induce 
non-negligible 
errors in the proxy annotations even if the prediction is very confident.
Some recent works \cite{cbst,chitta2018adaptive} seek to tackle this issue by scheduling the training samples via curriculum training strategies.

\par
In this paper, we propose a novel proxy label based method for unsupervised domain-adaptive semantic segmentation, 
allowing us to effectively adapt supervision of the source domain data to the unlabelled target domain data, as shown in Figure \ref{fig:intro}. Our model is first trained in an adversarial scheme with multiple discriminators to align features between the two domains. Motivated by the fact that the discriminators not only work as a regularizer to align features, but also provide another measure of the image indicating whether a particular region is domain-invariant. We would have higher confidence on those domain-invariant regions because the supervision in the source domain can be transferred through them more easily. By incorporating both the confidence from the standard segmentation classifier and the adversarial confidence from the discriminators, we are able to select more reliable proxy labels. Relying on the generated high-quality proxy labels, the model can be further boosted to adapt to the new domain.
The main contributions are summarized as below:
\begin{itemize}[itemsep=-1pt,topsep=-2pt]
\item We propose a novel proxy-based method for unsupervised domain-adaptive semantic segmentation. By integrating the benefits of the adversarial training into the proxy label generation, the proposed algorithm can effectively adapt the supervision from the source domain to the  target domain
of unlabelled real data. 
\item We propose a multi-adversarial training strategy to obtain high-quality proxies. Multiple adversarial discriminators 
at different levels of the network are applied to improve the quality of cross-domain predictions directly. More importantly, the outputs of multiple discriminators are used as an alternative signal for measuring the prediction confidence, leading to high-quality proxies. 
\item Relying on the generated high-quality proxies, the model can be boosted to adapt to the target domain. On two unsupervised domain-adaptive semantic segmentation tasks, GTA5 $\rightarrow$ Cityscapes \cite{Cordts2016Cityscapes} and SYNTHIA $\rightarrow$ Cityscapes, the proposed algorithm obtains the new state-of-the-art results.
\end{itemize}

\subsection{Related Work}
\textbf{Semantic Segmentation} Semantic segmentation is one of the key problems in computer vision and has been studied widely. Most current methods are based on fully convolutional networks (FCNs) \cite{fcn}, which train a model mapping an image to a semantic mask in an end-to-end fashion. There are  a series of dilation based methods \cite{deeplabv2, deeplabv3p, dilatedconv, psp} where dilated convolution kernels are used to retain the resolution of the features. There are also a group of encoder-decoder based methods \cite{segnet, refinenet, unet, deconv}. Those methods usually contain an encoder that gradually downsamples the features and captures high-level information, and a decoder that fuses the high-level information with low level features to recover details. 

Apart from fully supervised methods, there is a line of research that focuses on training models in a semi- or weakly supervised setting. Different types of supervision are used such as bounding boxes \cite{boxsup}, points \cite{point} and scribbles \cite{scribble}. Using the weakest supervision, image-level labels, to train models is also widely explored \cite{bidi, wssVideo, stc, wssSemi, constrainWSS}. These weakly and semi-supervised methods share some similar motivation with our research topic in the sense that one of our goals is to use synthetic data to 
spare 
enormous annotation effort.

\textbf{Domain-Adaptive Semantic Segmentation}
Domain adaptation methods aim to address the problem of domain shift between the source domain and the target domain. Some methods \cite{DA_BP, Long2015, NIPS2016_6110, Tzeng2015, tzeng2017adversarial} have been proposed to solve the issue in image classification problems. The main idea is to align the features and thus reduce the difference between the source and target domain distributions. The domain shift problem also exists in dense prediction problems such as semantic segmentation. A family of methods \cite{Tsai_adaptseg_2018, fcan, fcnwild, cycada, Sankaranarayanan18} tackle the problem by applying adversarial learning. The underlying insight  is as follows.
By confusing an adversarial domain discriminator that is used to distinguish the two domains, the features discrepancy can be minimized to make the samples from the two domain more similar. Adversarial learning is also applied to semi-supervised settings \cite{bmvcsemi}, where unlabelled data are also from the same domain.  The other line of methods \cite{cbst, chitta2018adaptive, zhang2017curriculum} focus on a self-training or curriculum learning framework. Curriculum methods adopt an easy-to-hard strategy to gradually mine and select confident samples using the knowledge learned from the annotated source domain \cite{chitta2018adaptive, zhang2017curriculum} and improve the performance . An iterative self-training procedure can also be used to alternatively generate proxy labels for the target images and retrain the network with these labels \cite{cbst}.

\section{Methodology}

\begin{figure*}[t!]
\centering
   \includegraphics[width=0.8\linewidth]{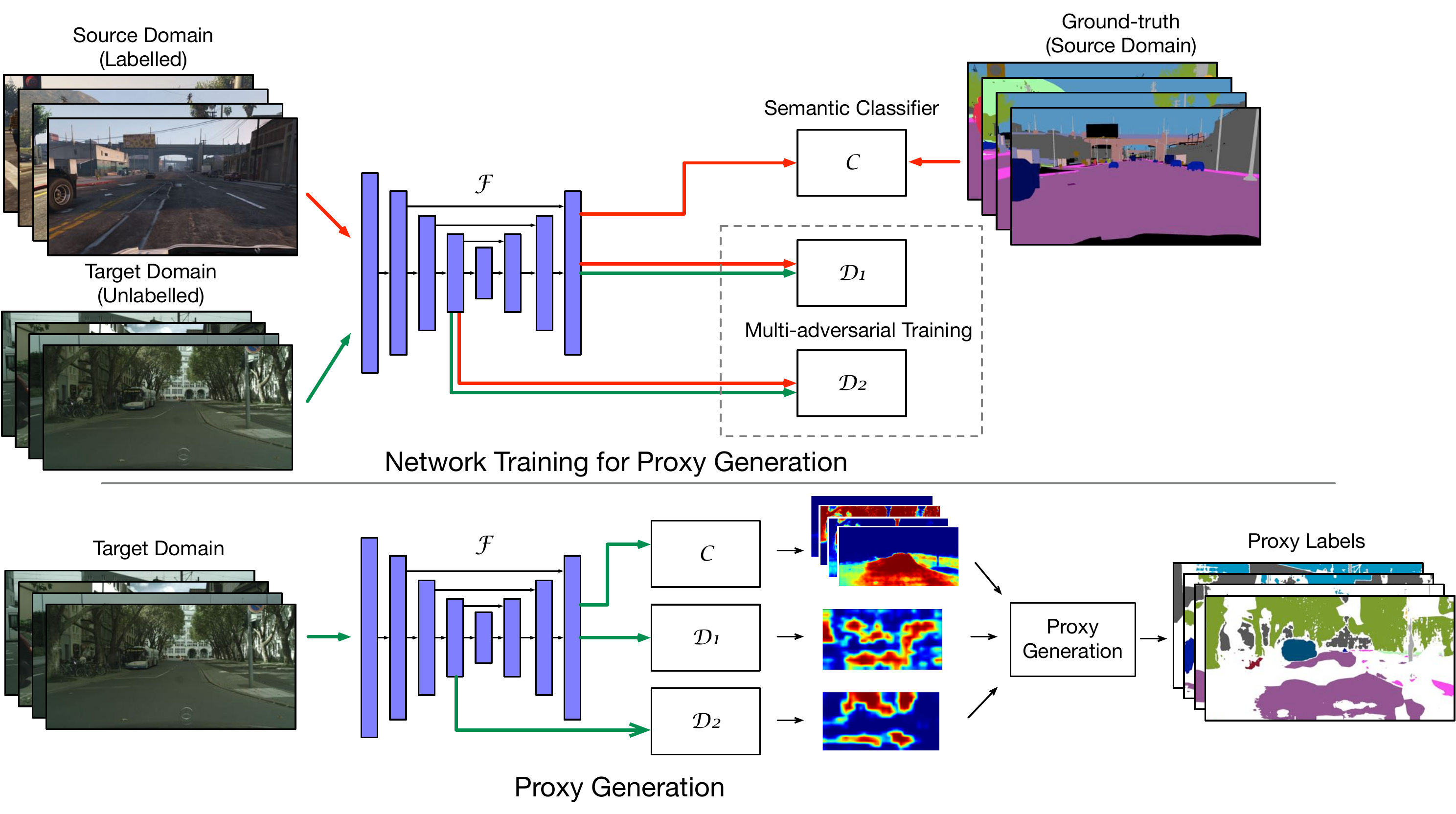}
   \caption{The pipeline of our proposed proxy generation method. The data flows of the source and target images are denoted by red and green lines respectively. In the training phase, images from both domains are fed into the feature backbone $\cF$ to produce both high level and mid level features. A classifier $\cC$ is used to supervise the training using the ground-truth provide in the source domain. Two discriminators $\cD_1$ and $\cD_2$ are used to minimize the feature discrepancy between the domains on different levels, which makes features more invariant to the domains.  In the proxy generation step, the target images are fed into the model and generate several confidence maps produced by $\cF$, $\cD_1$ and $\cD_2$. Those maps are passed into a proxy generation module to generate the final proxy label. Those high-quality proxy labels will can be used to largely boost the performance of the segmentation model on the target domain.}


\label{fig:pipeline}
\end{figure*}


Unsupervised domain adaptation is to adapt the supervision (or the learned representations) from the source domain to the unlabelled target domain. Let $\mathcal{X}_s$ and $\mathcal{X}_t$ denote the source and target domain of the images, respectively, which usually suffer from domain shift. 
For domain adaptive semantic segmentation, we have access to \emph{labelled} source domain data $\mathcal{I}_s=\{(\bX^{(m)}_s, \bY^{(m)}_s)\}^M_{m=1}$ and \emph{unlabelled} target domain data $\mathcal{I}_t=\{\bX^{(n)}_t\}^N_{n=1}$ where $\bX_s^{(m)}\in\mathcal{X}_s$ and $\bX_t^{(n)}\in\mathcal{X}_t$ denote the images from the source and target domain, and $\bY^{(m)}_s$ denotes the pixel-level mask for each image in the source domain dataset $\mathcal{I}_s$. 
Given both datasets $\mathcal{I}_s$ and $\mathcal{I}_t$, our goal is to train a model $f(\bX; \mathbf{\Theta})$ relying on the supervision information in $\mathcal{I}_s$. Given an image in the target domain, $\bX \in \mathcal{X}_t$, the model can obtain satisfactory predictions. Note that the unlabelled images $\bX_t^{(n)}$ in the target domain $\mathcal{I}_t$ are also used in the training. 

\subsection{Overview of the Proposed Framework}

Since there is no supervision in the target domain, we propose to generate proxy labels first and use the proxies to supervise the training in the target domain. Specifically, given $\mathcal{I}_s$ and $\mathcal{I}_t$, we tackle the problem by obtaining the proxy labels $\{\hat{\bY}^{(n)}_t\}^N_{n=1}$ relying on $\cI_s$ and $\cI_t$ and then train the semantic segmentation network $f(\bX; \mathbf{\Theta})$ on the proxy-labelled target dataset $\hat{\cI}_t=\{(\bX^{(n)}_t, \hat{\bY}^{(n)}_t)\}^N_{n=1}$. Therefore, the key is to generate high-quality proxy labels and mainly focus on this part.


\par
As illustrated in Figure \ref{fig:pipeline}, for generating the proxy labels, two main steps are involved in the proposed scheme -- \emph{multi-adversarial training} for feature alignment and \emph{proxy label generation} regularized by adversarial confidence. In the first step, we adopt an adversarial learning paradigm to train a network consisting of a feature backbone $\cF$, a semantic classifier $\cC$ and two discriminators, $\cD_1$ and $\cD_2$. Unlike previous methods \cite{cbst,chitta2018adaptive} that generate proxy labels using the network merely trained using the source data, the adversarial training leads to higher-quality proxies. In addition, inspired by deep supervision techniques \cite{psp}, which helps optimize the learning process, we use two discriminators to help minimize the feature discrepancy between two domains, referred to as multi-adversarial training. Apart from discriminator $\cD_1$, which is responsible for high level features, another discriminator $\cD_2$ is added to align mid level features.


In the training process, shown in the upper part of Figure \ref{fig:pipeline}, images from both the source and target domains are fed into $\cF$ to extract features. The flows of the source and target domain inputs are denoted by red and green lines respectively. Since the pixel-level ground-truth is available only in the source domain, the segmentation loss is only applied to the source images. For the two discriminators, the adversarial training is applied where $\cF$ tries to confuse the discriminators and the discriminators try to distinguish the features from the two domains.


\par
After the training process, we generate proxy labels for the unlabelled target domain images using the trained $\cF$, $\cC$, $\cD_1$ and $\cD_2$, as shown in the lower part of Figure \ref{fig:pipeline}. Given any image $\bX_t^{(n)}\in \cI_t$, the classifier $\cC$ predicts a normalized scoremap $\mathbf{P}_t^{(n)}\in \mbR^{HW\times L}$ that indicates the confidence to assign each pixel to a particular class out of the total $L$ classes, which is usually used to obtain the proxy labels for $\bX_t^{(n)}$ \cite{cbst}. However, simply using the scoremap is not sufficient as some wrong but confident predictions would lead to erroneous and noisy proxy labels. Considering that the two discriminators can determine whether a region is more likely from the source domain or the other, the outputs of the discriminators provide an alternative signal to check the confidence and correctness of the predictions on the target domain images. 
By incorporating the outputs of $\cC$, $\cD_1$ and $\cD_2$, our method can generate high-quality proxy labels. The regions with low confidence are set as ignored and will not affect the training (See the white-colored regions in the proxy label maps in Figure \ref{fig:pipeline}).


The estimated high-quality proxy labels for the target images will be used to train the network in a ``supervised manner''.
We will describe each part more specifically in the following sections.

\subsection{Multi-Adversarial Training}
Recall that our framework is composed of a feature backbone $\cF$, a classifier $\cC$ and two discriminators, $\cD_1$ and $\cD_2$. $\cF$ extracts features shared by both domains. Generally, if $\cF$ can produce domain-invariant features for the two domains, the classifier $\cC$ learned with the supervision on the source domain can also be used to predict satisfactory results on the unlabelled target domain.
In our setting, we adopt an encoder-decoder architecture for $\cF$, which has been widely used in advanced segmentation models \cite{unet, deeplabv3p, refinenet, segnet}. The output has resolution of $1/4$ of the input image. 
For the classifier, the segmentation loss $\cL_\text{seg}$ for the source domain dataset $\cI_s$ is defined as:
\begin{equation}
\mathcal{L}_\text{seg}=\sum_{m=1}^{M}\sum_{j=1}^{H \times W}\sum_{l=1}^{L}-\bY_{s}^{(m,j,l)}\log ( ~\cC(\cF(\bX_{s}^{(m)}))^{(j,l)} ),
\end{equation}
where $m$, $j$ and $l$ are used to index the $l$-th channel of the prediction at $j$-th position in $m$-th image. $\bY_{s}^{(m,j,l)}=1$ if the pixel is annotated with class $l$ and 0 otherwise.

\par
Considering that we will align both the high-level and some low-level features produced by $\cF(\cdot)$ via the multi-adversarial discriminators, we define $\cF_l(\cdot)$ to denote that subnetwork of $\cF(\cdot)$ that produces the intermediate features at the level lower than the final layer. 
For the target images, $\cF$ is trained to confuse the discriminators, making them believe the features are from the source domain. The discriminators are a binary classifier, where 0 means the sample is from the target domain while 1 represents the source domain. 
By letting $p_{s}(\bX)$ and $p_{t}(\bX)$ denote the data distributions of source domain dataset $\cI_s$ and target domain data $\cI_t$, we define the adversarial losses on two different levels as:
\begin{equation}
\begin{aligned}
  \mathcal{L}_{\text{adv1}} & = \mbE_{\bX_s\sim p_s(\bX_s)} [ \frac{1}{Z}\sum_{j=1}^{Z}\log(\cD_1(\cF(\bX_{s}))^{(j)} ]\\
  & ~+ \mbE_{\bX_t\sim p_t(\bX_t)} [ \frac{1}{Z}\sum_{j=1}^{Z}\log(1-\cD_1(\cF(\bX_{t}) )^{(j)} ], \\
\end{aligned}
\end{equation}
\begin{equation}
\begin{aligned}
  \mathcal{L}_{\text{adv2}} & = \mbE_{\bX_s\sim p_s(\bX_s)} [ \frac{1}{Z}\sum_{j=1}^{Z}\log(\cD_2(\cF_l(\bX_{s}))^{(j)} ]\\
  & ~+ \mbE_{\bX_t\sim p_t(\bX_t)} [ \frac{1}{Z}\sum_{j=1}^{Z}\log(1-\cD_2(\cF_l(\bX_{t}) )^{(j)} ],  \\
\end{aligned}
\end{equation}
where $Z$ denotes the numbers of the spatial ``pixels'' in the output of the discriminators; $j$ is used to index them.

\par
Overall, the multi-adversarial training is conducted by solving the following minimax optimization problem:
\begin{equation}
\min_{\cF,\cC}\max_{\cD_{1},\cD_{2}}~\lambda_\text{seg}\mathcal{L}_\text{seg}+\lambda_{1}\mathcal{L}_\text{adv1}+\lambda_{2}\mathcal{L}_\text{adv2}, 
\end{equation}
where $\cF$, $\cC$, $\cD_{1}$ and $\cD_{2}$ are used to denote the parameters of these networks for simplification; $\lambda_\text{seg}$, $\lambda_1$ and $\lambda_2$ are weighting constants.

\begin{figure}[t]
\begin{center}
   \includegraphics[width=1.0\linewidth]{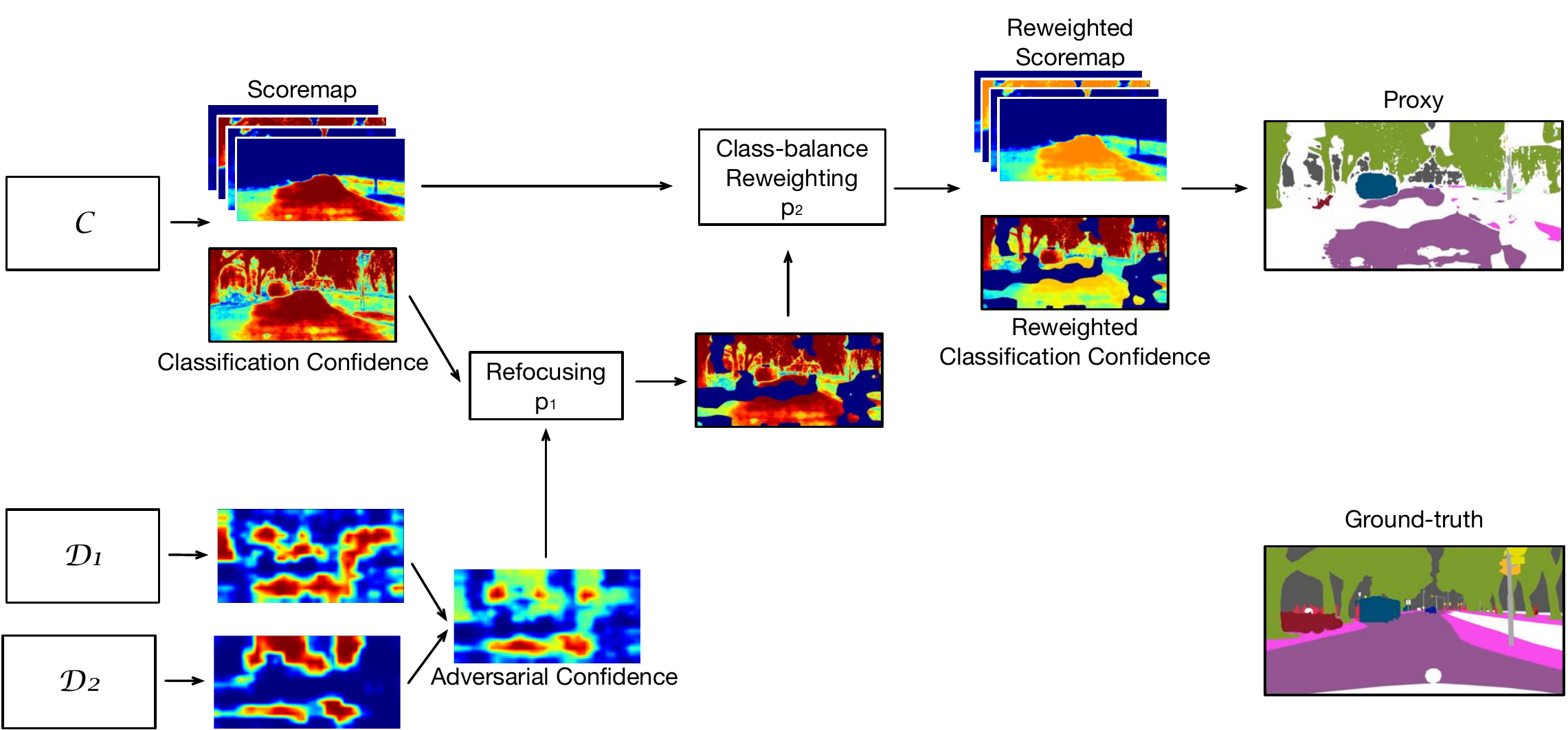}
\end{center}
   \caption{Illustration of proxy label generation. The classification confidence map given by the classifier $\cC$ is regularized by the adversarial confidence map produced by the discriminators $\cD_1$ and $\cD_2$. By further using a class-balance strategy, the scoremap is reweighted and the proxy labels are generated based on the reweighted scoremap and classification confidence map.} 
\label{fig:proxy_gen}
\end{figure}
\subsection{Proxy Generation Regularized by Multi-Adversarial Training}

After obtaining $\cF$, $\cC$, $\cD_1$ and $\cD_2$, we generate the proxy labels for the target image as illustrated in Figure \ref{fig:proxy_gen}. 
The general idea is to use both the classification confidence from the classifier $\cC$ and the adversarial confidence from the discriminators, $\cD_1$ and $\cD_2$, to select high confidence samples as the proxy. 



\textbf{Classification Confidence}
The classifier $\cC$ is trained with labeled data from the source domain. Given any image $\bX_t^{(n)}$ from the target domain, $\cC$ can produce a scoremap $\mathbf{P}_t^{(n)}$ representing the probability of assigning each pixel to a particular category. When taking the maximum value across all the categories for each pixel, we can obtain a \emph{classification confidence} map indicating the prediction confidence for the pixels. Naturally, the proxies can be selected from the pixels with high confidence by using the classification confidence map as the criteria. 
However, it is not reliable enough since the highly confident but wrong predictions are harmful to training; and some potentially beneficial samples with low confidence would be ignored. 
This motivates us to use the extra measurement of the predictions. 

\textbf{Adversarial Confidence}
In the process of adversarial training, the discriminators are trained to distinguish the features from the two domains. The feature backbone $\cF$ is trained to confuse the discriminators $\cD_1$ and $\cD_2$, for learning domain-invariant features. 
For the target domain image, each discriminator can give a confidence map in which the large values indicate indistinguishable regions that are more likely from the source domain; the small values indicate that the regions are easy to be recognized as in target domain \cite{fcan}. Considering that $\cF$ and $\cC$ are trained with the supervision from the source domain data, we can interpreter the outputs of the discriminators as an alternative confidence map to measure whether a specific pixel can be benefited by the supervision signal adapted from the source domain and then the quality of predictions, referred to as adversarial confidence map. 
We take the outputs of the two discriminators and normalize them to the same scale. An averaged adversarial confidence map is obtained as $\textbf{A}^{(n)}\in\mbR^{H\times W}$. The two discriminators not only jointly promote the feature alignment, but also cooperatively provide the confidence for proxy generation. A similar idea \cite{bmvcsemi} is used to be applied for semi-supervised learning, which simply uses a fixed threshold to generate a temporal label map in an online fashion, however. The proposed can generate more reliable proxy labels by taking account of the multi-adversarial training and statistics of the whole training set.


\textbf{Confidence Refocusing}
To incorporate the adversarial confidence into the classification confidence, we threshold $\textbf{A}^{(n)}$ such that we refocus on the regions that are domain-invariant. This procedure is controlled by a parameter $p_1$ that defines the proportion of selected pixels. To achieve this, we iterate all the images in the target dataset and store all the vectorized confidence maps, $[\text{vec}(\textbf{A}^{(1)}), ..., \text{vec}(\textbf{A}^{(n)})]$. A threshold $t_1$ is calculated as $p_1$ percentile of all the scores. After this step, only the domain invariant regions are highlighted for further processing while the regions whose adversarial score is lower than $t_1$ are ignored in the later steps. 

\textbf{Class-balance Reweighting}
As discussed in \cite{cbst}, domain gaps may cause different domain-transfer difficulties that make some easy-to-transfer samples have high confidence. Those samples might dominate the process of proxy selection, making it difficult for some samples belonging to a hard-to-transfer class to be selected. To alleviate this issue, we adopt a class-balance strategy similar to \cite{cbst} to balance the domain transfer among different classes. The process is controlled by another parameter $p_2$ that defines the proportion of selected samples in this step. Similar to the previous step, we iterate all the image in the target dataset and analyze the scoremaps. Since it is operated in a class-wise scheme, all the scores are sorted independently for each category. In other words, we want to find a vector $\textbf{t}_2$ of length $L$ containing an independent threshold for each category. Please note that we only focus on the domain invariant regions, therefore only the samples whose score is higher than $t_1$ are considered in this step. Then all the scoremaps $\textbf{P}^{(n)}_t$ are reweighted by dividing $\textbf{t}_2$ within the corresponding channel.

\textbf{Proxy Generation}
After reweighting, the pixels whose response is larger than $1$ indicate high confidence. We are interested in the pixels where both the classification confidence and adversarial confidence agree. Therefore, the proxy labels are generated as the follows:
\begin{equation}
\widehat{\bY}{}^{(n,j,l)}=\begin{cases}
1, & \text{if}\;l={\text{argmax}_{l}}\frac{\textbf{P}_{t}^{(n,j,l)}}{\textbf{t}_2^{(l)}},\\
 & {\textbf{P}_{t}^{(n,j,l)}}>{\textbf{t}_2^{(l)}}\;\text{and}\;\textbf{A}^{(n,j)}>t_1,\\
0, & \text{otherwise},
\end{cases}
\end{equation}
where $\textbf{P}_{t}^{(n,j,l)}$ represents the score of $l$-th channel at $j$-th pixel of $n$-th image in the target dataset and $\hat{\bY}{}^{(n,j,l)}$ is the corresponding label. $\textbf{t}_2^{(l)}$ is $l$-th value of the threshold vector $\textbf{t}_2$.

The pixels that are not assigned with any class are ignored in the proxies and will not affect the training.

\subsection{Network Architecture}
\begin{figure}[t!]
\begin{center}
   \includegraphics[width=0.8\linewidth]{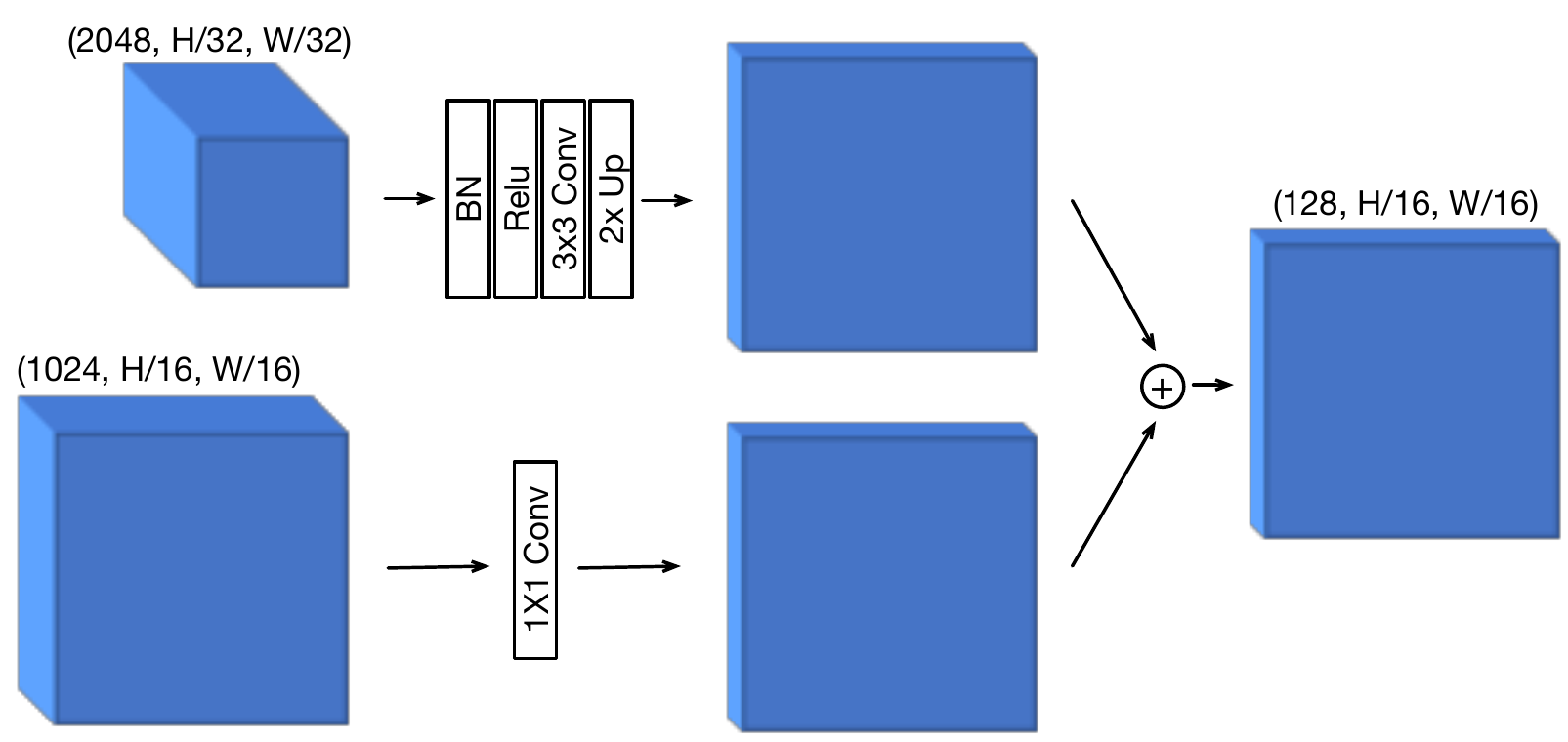}
\end{center}
   \caption{Configuration of our fusion block. The low-resolution feature maps are fed into a series of units and upsampled to the same size of the high-resolution feature maps. Both the features are adapted to have the same channels and then merged by element-wise summation.}
\label{fig:block}
\end{figure}

The feature backbone $\cF$ has an encoder-decoder architecture. The encoder is a classification network, \eg VGG16 or Resnet101, pretrained on ImageNet \cite{imagenet}. The decoder has a light-weight structure. We upsample the low-resolution high-level features and add lower-level features to it. To make the dimension consistent, we reduce the number of channels to 128 for all the feature maps. Taking Resnet101 as an example, Figure \ref{fig:block} shows how the features are fused. The upper feature maps are from the block Res5c and the lower ones are from the block Res4b22. The final output of the decoder has a resolution of $1/4$. In order to enlarge the field of view, we also use dilation for the 3x3 Conv layer. 

For the classifier $\cC$, we use an Atros Spatial Pyramid Pooling (ASPP) \cite{deeplabv2} module with the rates 1, 2, 4 and 8 to aggregate different scales. For the two discriminators, we adopt a patch based structure by using a small fully convolutional network, similar to \cite{Tsai_adaptseg_2018, pix2pix2017}. For a specific region, $\cD_1$ and $\cD_2$ are used to determine if it is from the source domain or the target domain. The discriminators consist of several convolutional layers followed by a Leaky ReLU layer \cite{leakyrelu}. There are also average pooling layers with strides in between to downsample the features. In our setting, we control the output resolution to be $1/32$ of the input image size. Therefore, the two discriminators have a bit difference in term of downsampling times because they are applied to different sizes of features, $\cF(\cdot)$ and $\cF_l(\cdot)$.

\section{Experiments}

\begin{table*}[t]
 \centering 
         \scalebox{0.6031102}{
         \begin{tabular}{  r|r|ccccccccccccccccccc|c }
\hline 
 & Method & Road & SW & Build & Wall & Fence & Pole & TL & TS & Veg. & Terrain & Sky & PR & Rider & Car & Truck & Bus & Train & Motor & Bike & mIoU\tabularnewline
\hline 
VGG16 & Src only \cite{fcnwild} & 31.9 & 18.9 & 47.7 & 7.4 & 3.1 & 16 & 10.4 & 1.0 & 76.5 & 13.0 & 58.9 & 36.0 & 1.0 & 67.1 & 9.5 & 3.7 & 0.0 & 0.0 & 0.0 & 21.2\tabularnewline
 & FCN wild \cite{fcnwild} & 70.4 & 32.4 & 62.1 & 14.9 & 5.4 & 10.9 & 14.2 & 2.7 & 79.2 & 21.3 & 64.6 & 44.1 & 4.2 & 70.4 & 8.0 & 7.3 & 0.0 & 3.5 & 0.0 & 27.1\tabularnewline
\hline 
VGG16 & Src only \cite{zhang2017curriculum} & 18.1 & 6.8 & 64.1 & 7.3 & 8.7 & 21.0 & 14.9 & 16.8 & 45.9 & 2.4 & 64.4 & 41.6 & 17.5 & 55.3 & 8.4 & 5.0 & 6.9 & 4.3 & 13.8 & 22.3\tabularnewline
 & Curr. DA \cite{zhang2017curriculum} & 74.9 & 22.0 & 71.7 & 6.0 & 11.9 & 8.4 & 16.3 & 11.1 & 75.7 & 13.3 & 66.5 & 38.0 & 9.3 & 55.2 & 18.8 & 18.9 & 0.0 & 16.8 & 16.6 & 28.9\tabularnewline
\hline 
VGG16 & Src only \cite{cycada} & 26.0 & 14.9 & 65.1 & 5.5 & 12.9 & 8.9 & 6.0 & 2.5 & 70.0 & 2.9 & 47.0 & 24.5 & 0.0 & 40.0 & 12.1 & 1.5 & 0.0 & 0.0 & 0.0 & 17.9\tabularnewline
 & CyCADA \cite{cycada} & 85.2 & 37.2 & 76.5 & 21.8 & 15.0 & 23.8 & 22.9 & 21.5 & 80.5 & 31.3 & 60.7 & 50.5 & 9.0 & 76.9 & 17.1 & 28.2 & 4.5 & 9.8 & 0.0 & 35.4\tabularnewline
\hline 
Resnet101 & Src only \cite{Tsai_adaptseg_2018} & 75.8 & 16.8 & 77.2 & 12.5 & 21.0 & 25.5 & 30.1 & 20.1 & 81.3 & 24.6 & 70.3 & 53.8 & 26.4 & 49.9 & 17.2 & 25.9 & 6.5 & 25.3 & 36.0 & 36.6\tabularnewline
 & AdaptSeg \cite{Tsai_adaptseg_2018} & 86.5 & 36.0 & 79.9 & 23.4 & 23.3 & 23.9 & 35.2 & 14.8 & 83.4 & 33.3 & 75.6 & 58.5 & 27.6 & 73.7 & 32.5 & 35.4 & 3.9 & 30.1 & 28.1 & 42.4\tabularnewline
\hline 
VGG16 & Src only \cite{cbst} & 64.0 & 22.1 & 68.6 & 13.3 & 8.7 & 19.9 & 15.5 & 5.9 & 74.9 & 13.4 & 37.0 & 37.7 & 10.3 & 48.2 & 6.1 & 1.2 & 1.8 & 10.8 & 2.9 & 24.3\tabularnewline
 & CBST \cite{cbst} & 90.4 & 50.8 & 72 & 18.3 & 9.5 & 27.2 & 28.6 & 14.1 & 82.4 & 25.1 & 70.8 & 42.6 & 14.5 & 76.9 & 5.9 & 12.5 & 1.2 & 14.0 & 28.6 & 36.1\tabularnewline
\hline 
WideRes-38 & Src only \cite{cbst} & 70.0 & 23.7 & 67.8 & 15.4 & 18.1 & 40.2 & 41.9 & 25.3 & 78.8 & 11.7 & 31.4 & 62.9 & 29.8 & 60.1 & 21.5 & 26.8 & 7.7 & 28.1 & 12 & 35.4\tabularnewline
 & CBST \cite{cbst} & 89.6 & \textbf{58.9} & 78.5 & 33.0 & 22.3 & \textbf{41.4} & \textbf{48.2} & \textbf{39.2} & 83.6 & 24.3 & 65.4 & 49.3 & 20.2 & 83.3 & 39 & \textbf{48.6} & \textbf{12.5} & 20.3 & 35.3 & 47.0\tabularnewline
\hline 
VGG16 & Src only & 53.3 & 20.4 & 73.3 & 10.7 & 18.9 & 20.8 & 12.9 & 7.7 & 76.0 & 6.5 & 57.0 & 44.0 & 18.4 & 40.5 & 25.8 & 14.3 & 1.2 & 17.5 & 11.6 & 27.9\tabularnewline
 & Ours & 87.3 & 42.7 & 79.6 & 27.5 & 24.6 & 28.8 & 16.8 & 7.7 & 84.7 & 24.1 & 66.6 & 56.1 & 26.8 & 80.6 & 22.1 & 13.6 & 0.0 & 24.9 & 27.1 & 39.0\tabularnewline
\hline 
Resnet101 & Src only & 64.3 & 16.4 & 78.6 & 21.8 & 22.4 & 28.2 & 32.6 & 17.8 & 80.3 & 12.7 & 70.7 & 53.9 & 26.7 & 70.1 & \textbf{34.8} & 28.4 & 10.5 & 29.7 & 24.6 & 38.1\tabularnewline
 & Ours & \textbf{90.7} & 49.3 & \textbf{84.8} & \textbf{36.0} & \textbf{36.3} & 41.3 & 46.7 & 29.7 & \textbf{87.3} & \textbf{34.6} & \textbf{82.5} & \textbf{68.1} & \textbf{38.4} & \textbf{85.3} & 25.9 & 29.2 & 1.7 & \textbf{46.1} & \textbf{44.6} & \textbf{50.4}\tabularnewline
\hline 
\end{tabular}
    }
   
    \caption{Experimental results on GTA5 to Cityscapes.}
    \vspace{-0.2cm}
    \label{tab:gta2cs}
\end{table*}

\begin{table*}[t]
\centering 
\scalebox{0.65}{
\begin{tabular}{ r |r |cccccccccccccccc|c|c }
\hline 
 & Method & Road & SW & Build & Wall{*} & Fence{*} & Pole{*} & TL & TS & Veg. & Sky & PR & Rider & Car & Bus & Motor & Bike & mIoU & mIoU{*}\tabularnewline
\hline 
VGG16 & Src only \cite{fcnwild} & 6.4 & 17.7 & 29.7 & 1.2 & 0.0 & 15.1 & 0.0 & 7.2 & 30.3 & 66.8 & 51.1 & 1.5 & 47.3 & 3.9 & 0.1 & 0.0 & 17.4 & 20.2\tabularnewline
 & FCN wild \cite{fcnwild} & 11.5 & 19.6 & 30.8 & 4.4 & 0.0 & 20.3 & 0.1 & 11.7 & 42.3 & 68.7 & 51.2 & 3.8 & 54.0 & 3.2 & 0.2 & 0.6 & 20.2 & 22.1\tabularnewline
\hline 
VGG16 & Src only \cite{zhang2017curriculum} & 5.6 & 11.2 & 59.6 & 8.0 & 0.5 & 21.5 & 8.0 & 5.3 & 72.4 & 75.6 & 35.1 & 9.0 & 23.6 & 4.5 & 0.5 & 18.0 & 22.0 & 27.6\tabularnewline
 & Curr. DA \cite{zhang2017curriculum} & 65.2 & 26.1 & 74.9 & 0.1 & 0.5 & 10.7 & 3.5 & 3.0 & 76.1 & 70.6 & 47.1 & 8.2 & 43.2 & 20.7 & 0.7 & 13.1 & 29.0 & 34.8\tabularnewline
\hline 
Resnet101 & Src only \cite{Tsai_adaptseg_2018} & 55.6 & 23.8 & 74.6 & - & - & - & 6.1 & 12.1 & 74.8 & 79 & 55.3 & 19.1 & 39.6 & 23.3 & 13.7 & 25.0 & - & 38.6\tabularnewline
 & AdaptSeg \cite{Tsai_adaptseg_2018} & \textbf{84.3} & \textbf{42.7} & 77.5 & - & - & - & 4.7 & 7.0 & 77.0 & 82.5 & 54.3 & \textbf{21.0} & 72.3 & 32.2 & 18.0 & 32.3 & - & 46.6\tabularnewline
\hline 
VGG16 & Src only \cite{cbst} & 17.2 & 19.7 & 47.3 & 1.1 & 0.0 & 19.1 & 3.0 & 9.1 & 71.8 & 78.3 & 37.6 & 4.7 & 42.2 & 9.0 & 0.1 & 0.9 & 22.6 & 26.2\tabularnewline
 & CBST \cite{cbst} & 69.6 & 28.7 & 69.5 & 12.1 & 0.1 & 25.4 & 11.9 & 13.6 & 82.0 & 81.9 & 49.1 & 14.5 & 66.0 & 6.6 & 3.7 & 32.4 & 35.4 & 40.7\tabularnewline
\hline 
WideRes-38 & Src only \cite{cbst} & 32.6 & 21.5 & 46.5 & 4.8 & 0.1 & 26.5 & 14.8 & 13.1 & 70.8 & 60.3 & 56.6 & 3.5 & 74.1 & 20.4 & 8.0 & 13.1 & 29.2 & 33.6\tabularnewline
 & CBST \cite{cbst} & 53.6 & 23.7 & 75.0 & \textbf{12.5} & 0.3 & \textbf{36.4} & \textbf{23.5} & 26.3 & \textbf{84.8} & 74.7 & \textbf{67.2} & 17.5 & \textbf{84.5} & 28.4 & 15.2 & \textbf{55.8} & 42.5 & 48.4\tabularnewline
\hline 
VGG16 & Src only & 13.1 & 14.6 & 63.7 & 1.9 & 0.0 & 8.8 & 0.2 & 0.0 & 64.6 & 74.3 & 41.5 & 1.9 & 31.9 & 12.5 & 0.1 & 0.6 & 20.6 & 24.5\tabularnewline
 & Ours & 76.6 & 31.1 & 79.6 & 0.1 & 0.1 & 29.5 & 0.0 & 9.1 & 81.8 & 78.2 & 54.6 & 20.9 & 73 & 28.8 & 2.1 & 23.3 & 36.8 & 43.0 \tabularnewline
\hline 
Resnet101 & Src only & 43.7 & 19.2 & 76.2 & 9.4 & 0.5 & 26.2 & 7.4 & 12.5 & 77.6 & 78.5 & 53.0 & 13.3 & 52.7 & 37.4 & 14.7 & 12.0 & 33.4 & 38.3\tabularnewline
 & Ours & 74.6 & 32.7 & \textbf{82.5} & 6.5 & \textbf{2.2} & 32.7 & 21.5 & \textbf{27.0} & 81.9 & \textbf{83.5} & 59.8 & 13.1 & 82.2 & \textbf{53.4} & \textbf{30.2} & 36.9 & \textbf{45.0} & \textbf{52.3}\tabularnewline
\hline 
\end{tabular}}

\caption{Experimental results on SYNTHIA to Cityscapes.}
\vspace{-0.2cm}
\label{tab:syn2cs}
\end{table*}

\begin{figure*}[t]
    \centering
    \scalebox{1}{
    \begin{tabular}{cccc}
         \subfloat{\includegraphics[width=0.2\linewidth]{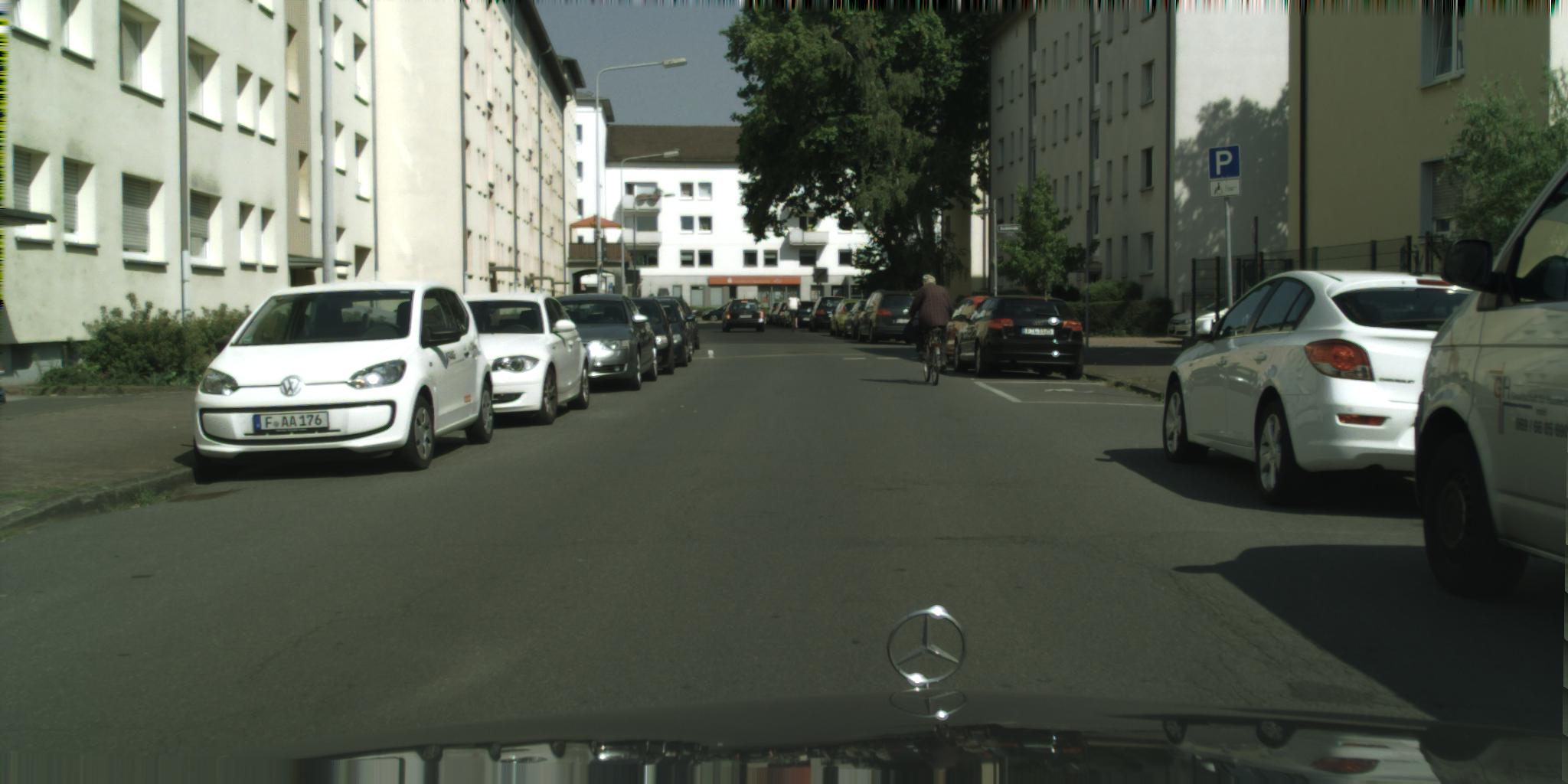}} &
         \subfloat{\includegraphics[width=0.2\linewidth]{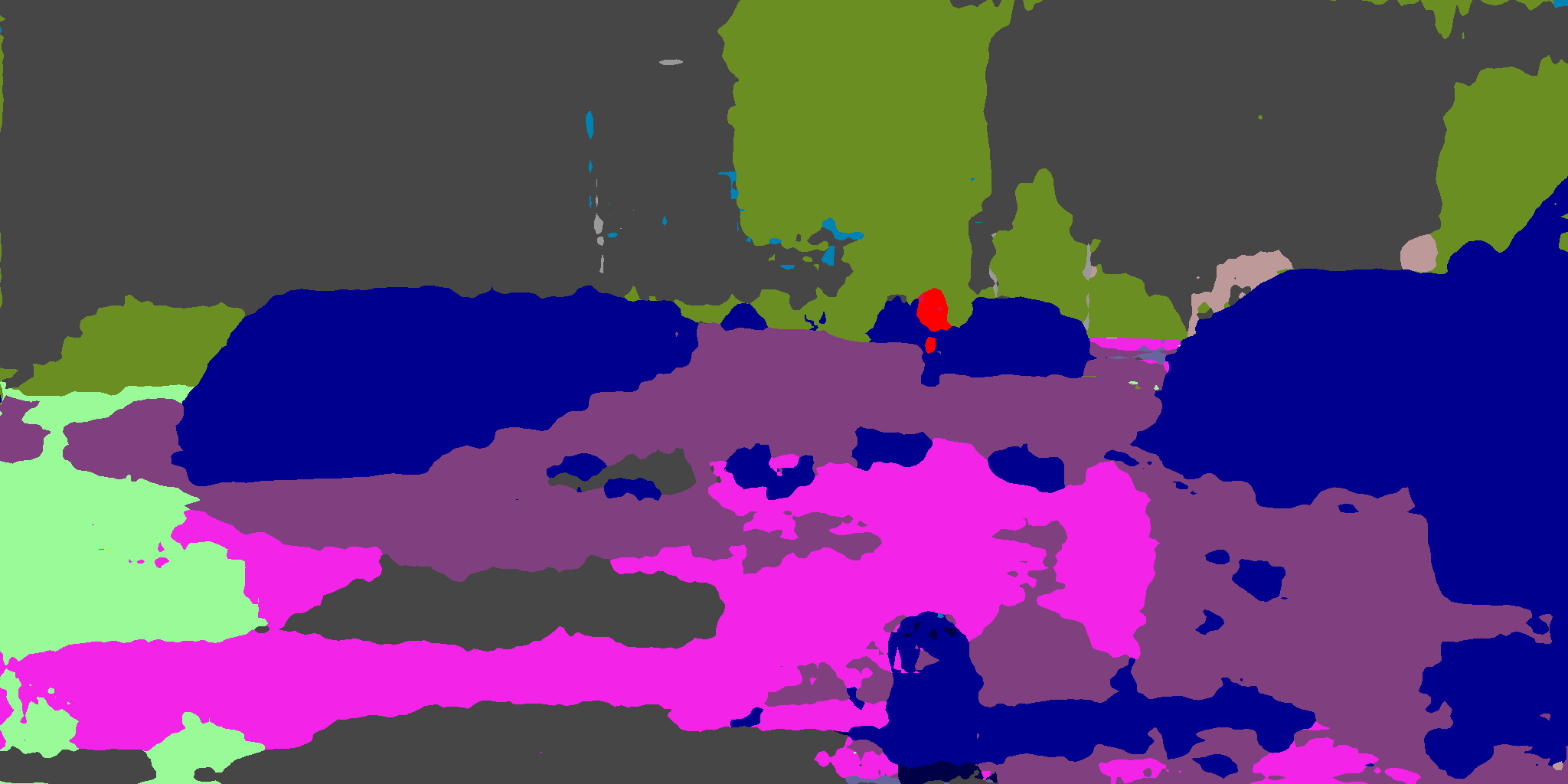}} &
         \subfloat{\includegraphics[width=0.2\linewidth]{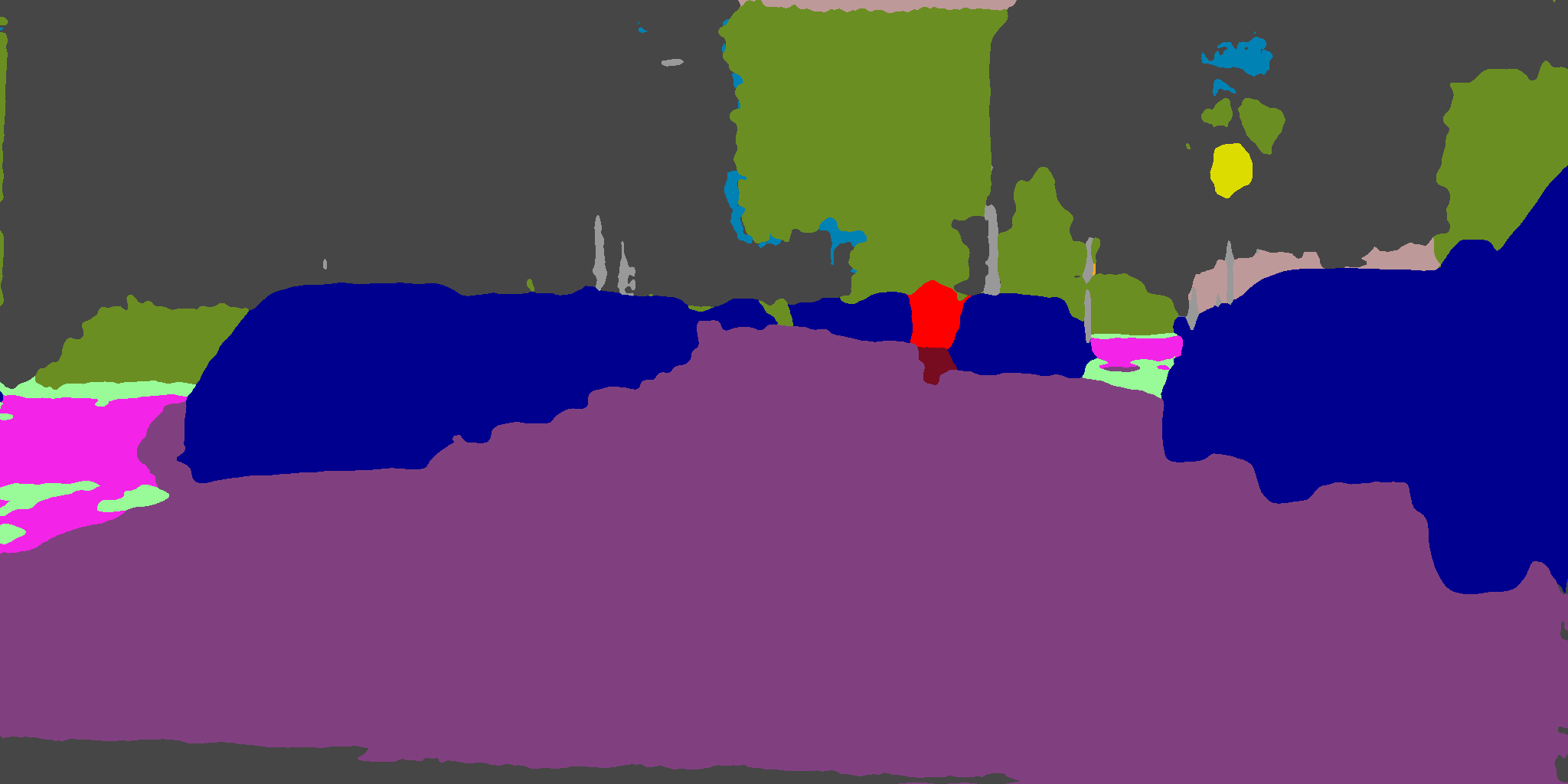}} &
         \subfloat{\includegraphics[width=0.2\linewidth]{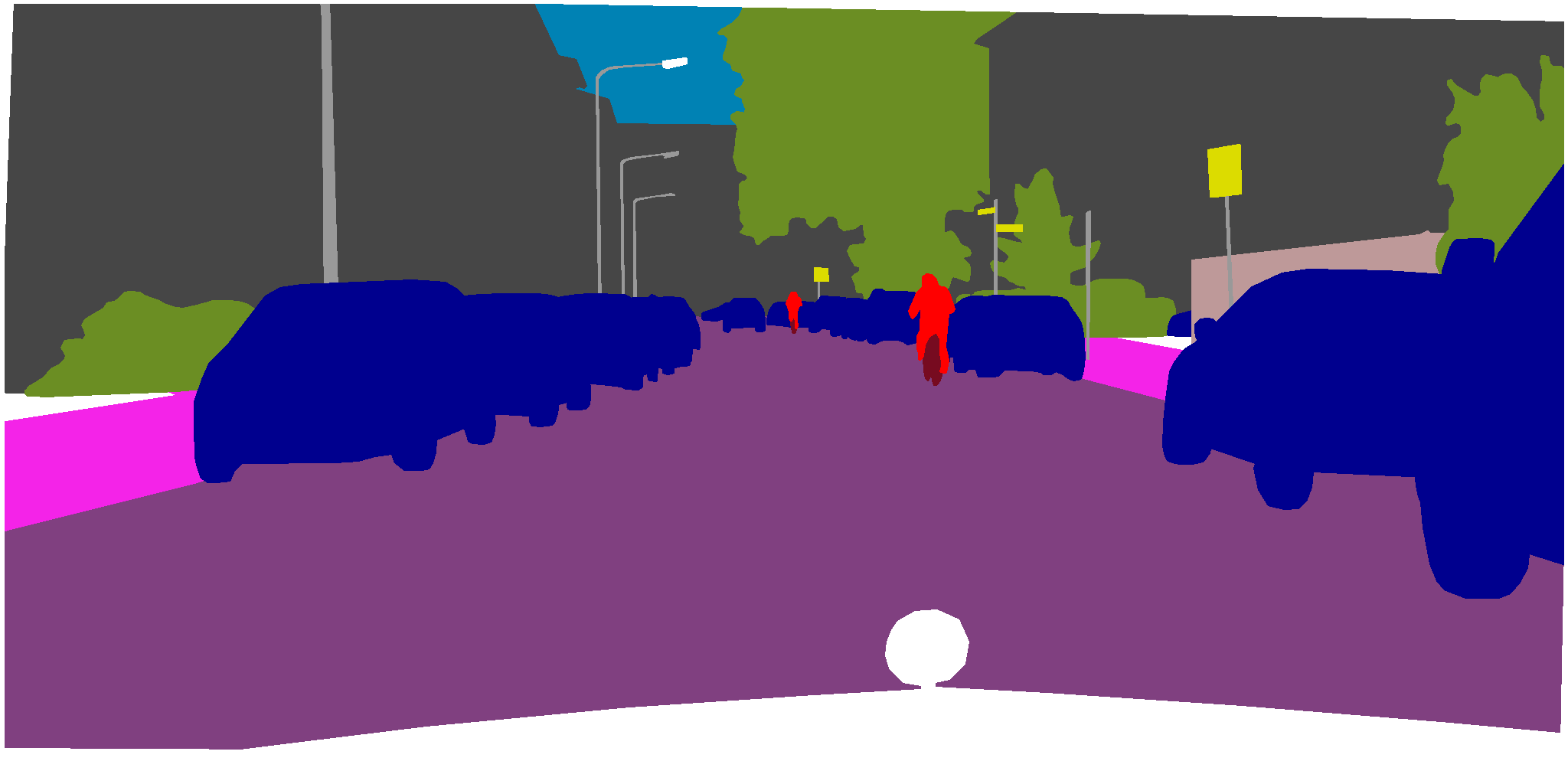}} \\
         
         \subfloat{\includegraphics[width=0.2\linewidth]{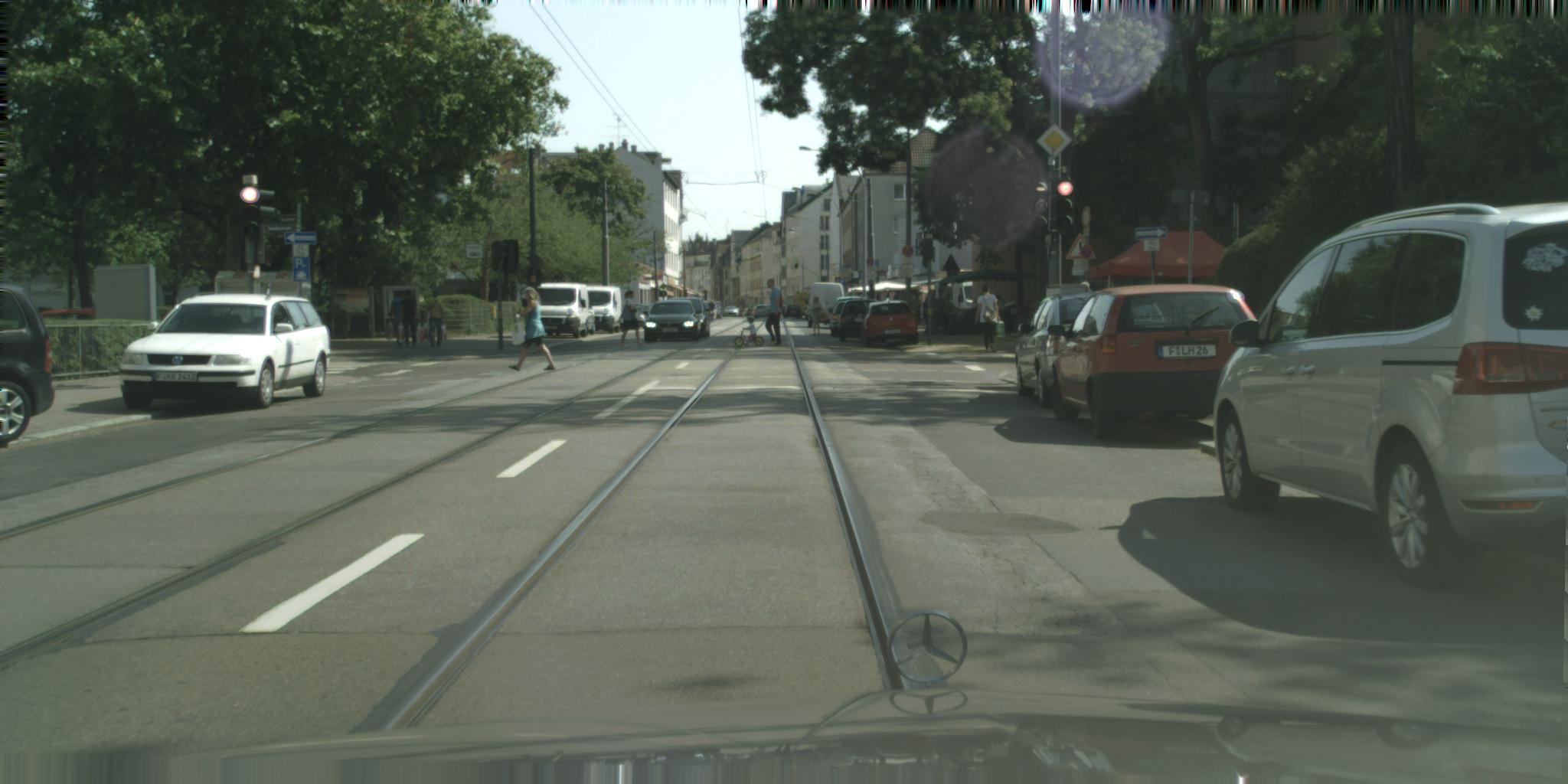}} &
         \subfloat{\includegraphics[width=0.2\linewidth]{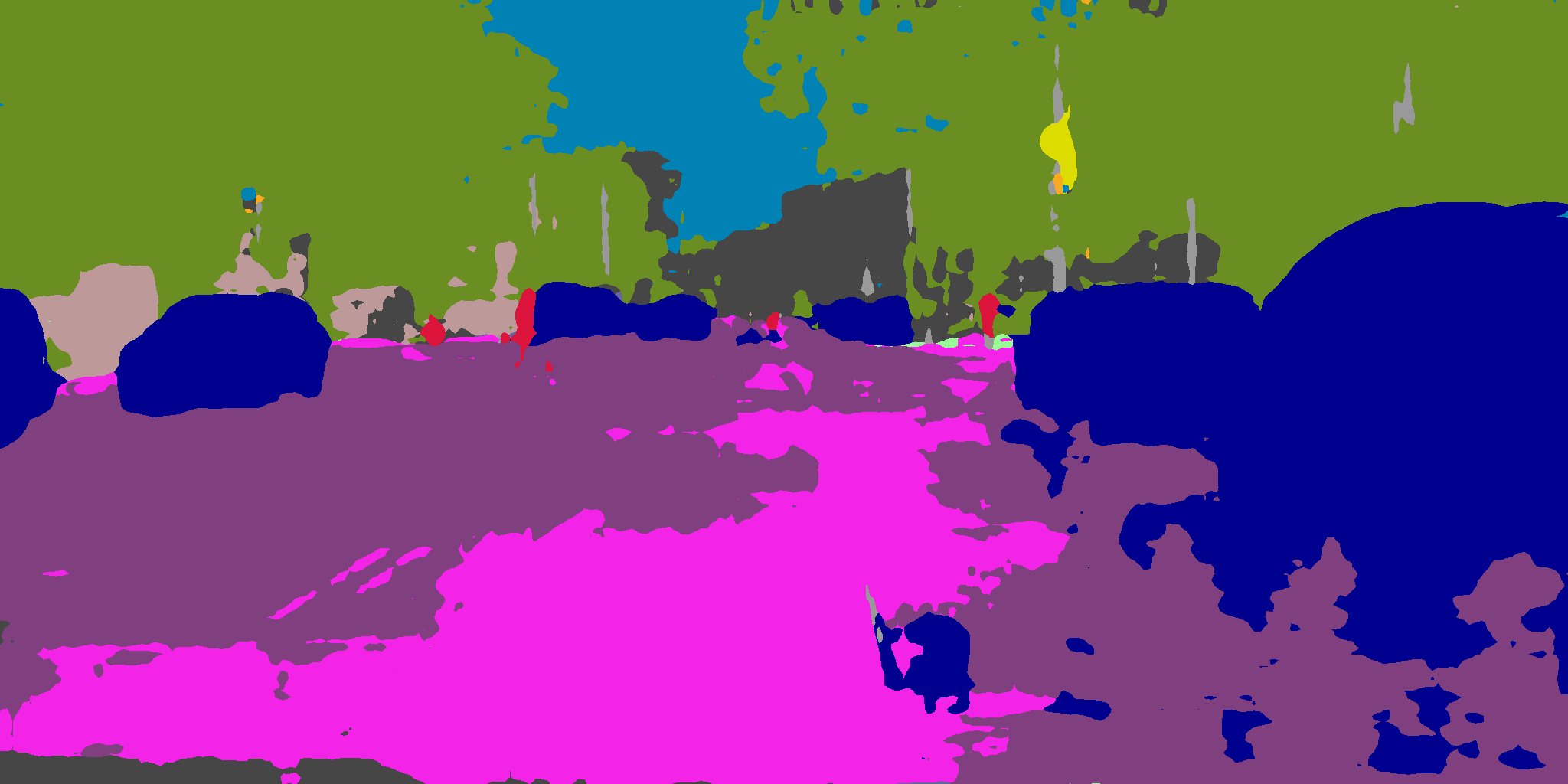}} &
         \subfloat{\includegraphics[width=0.2\linewidth]{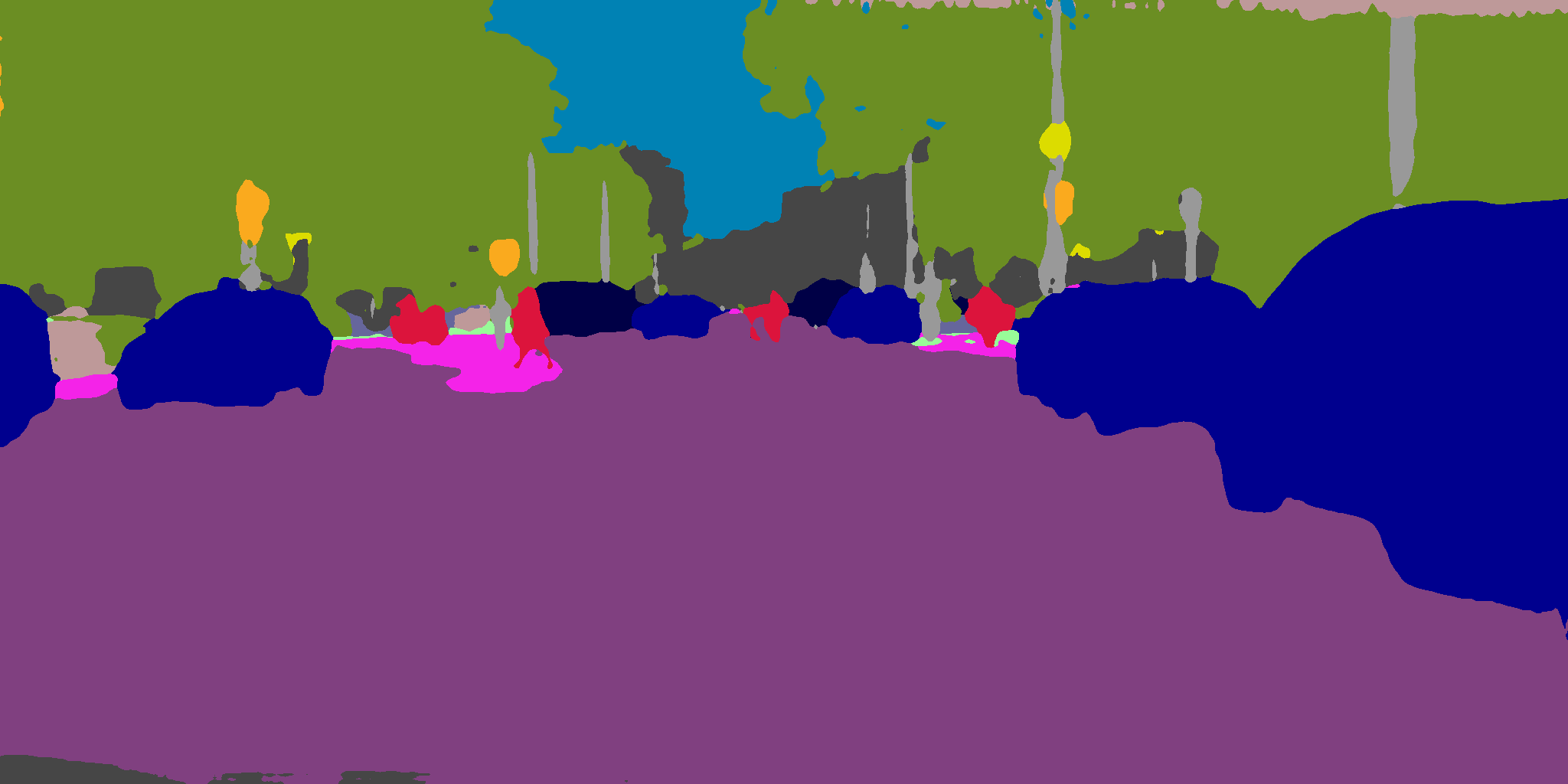}} &
         \subfloat{\includegraphics[width=0.2\linewidth]{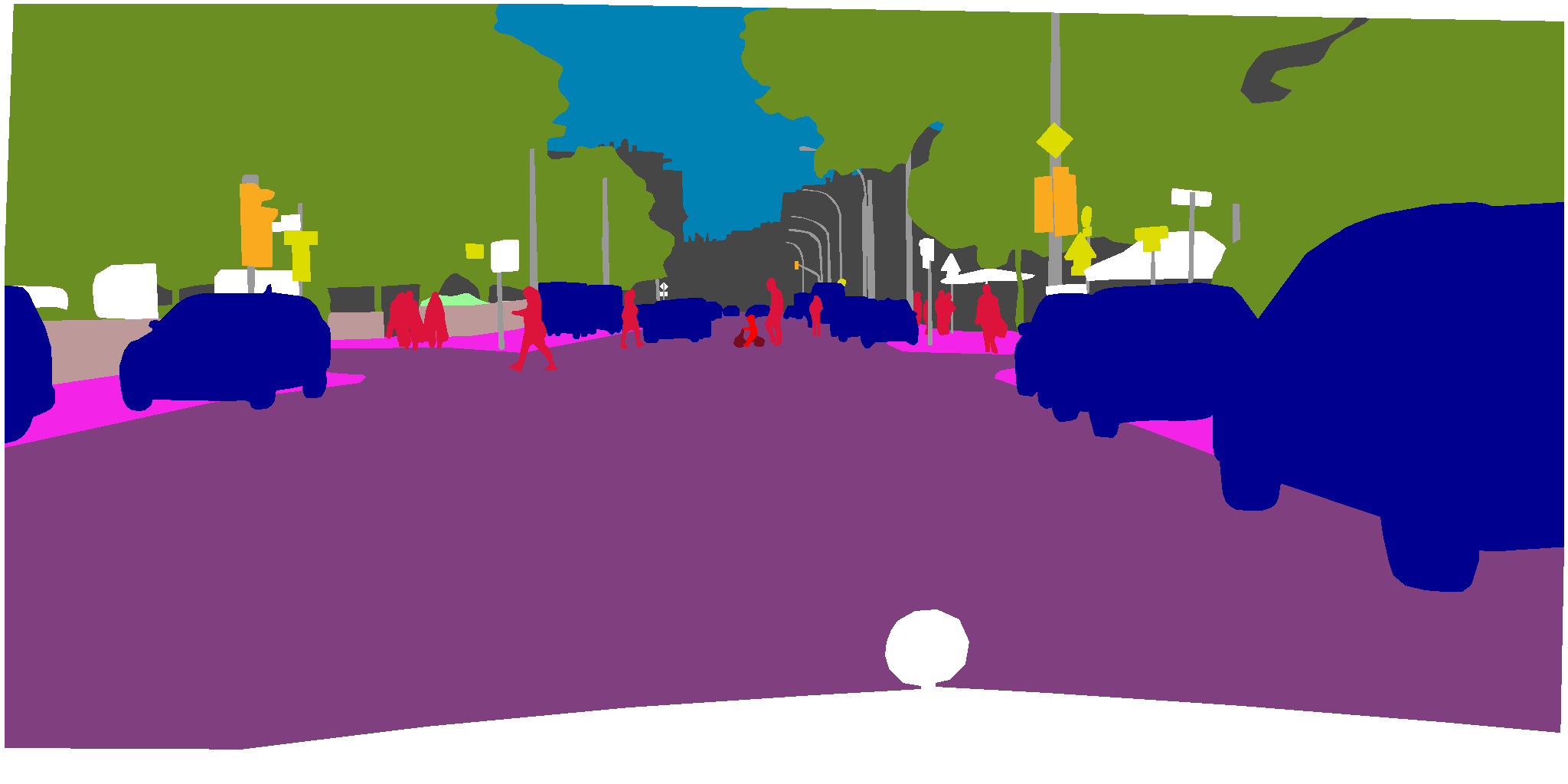}} \\
         
         \subfloat{\includegraphics[width=0.2\linewidth]{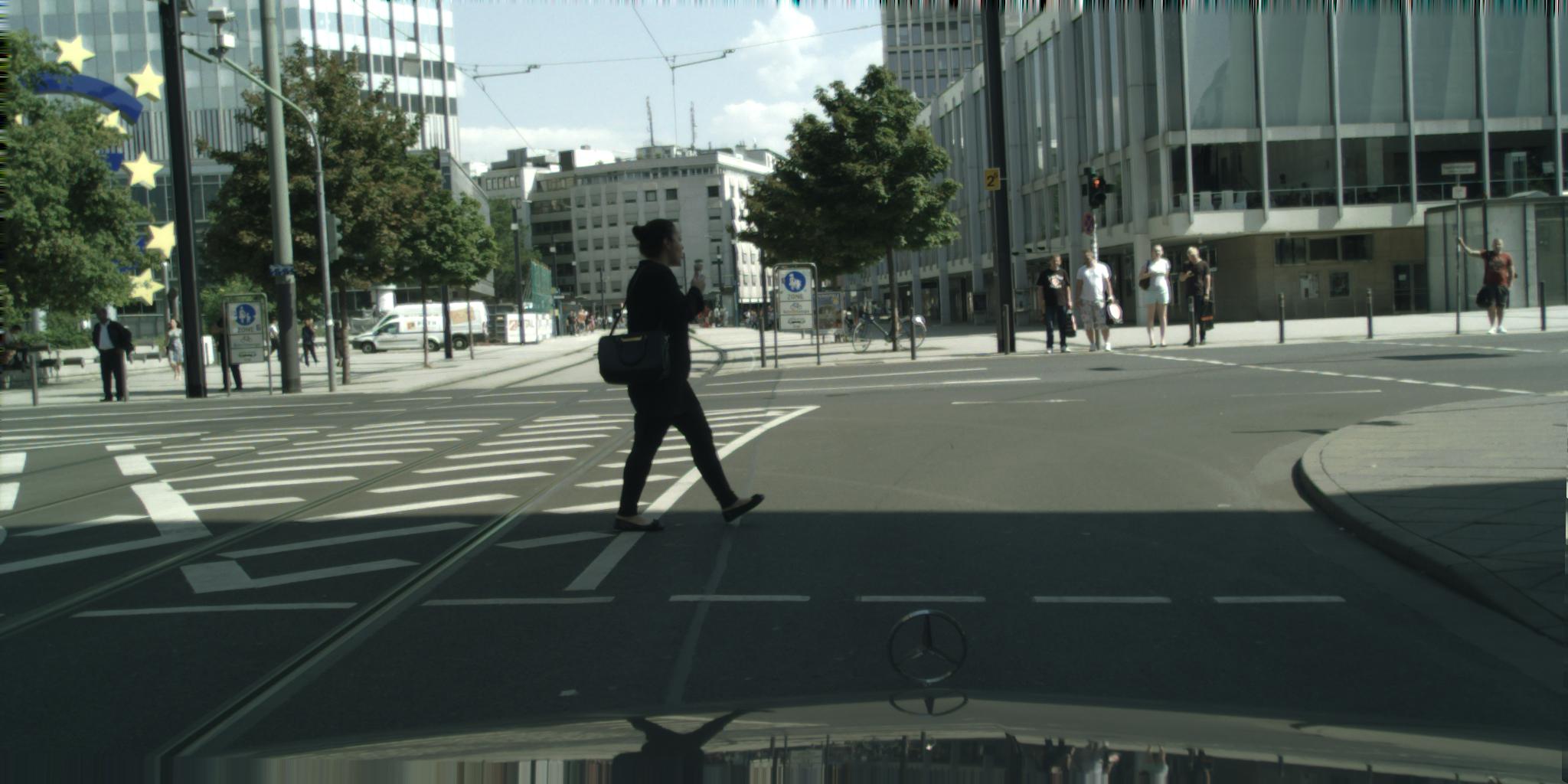}} &
         \subfloat{\includegraphics[width=0.2\linewidth]{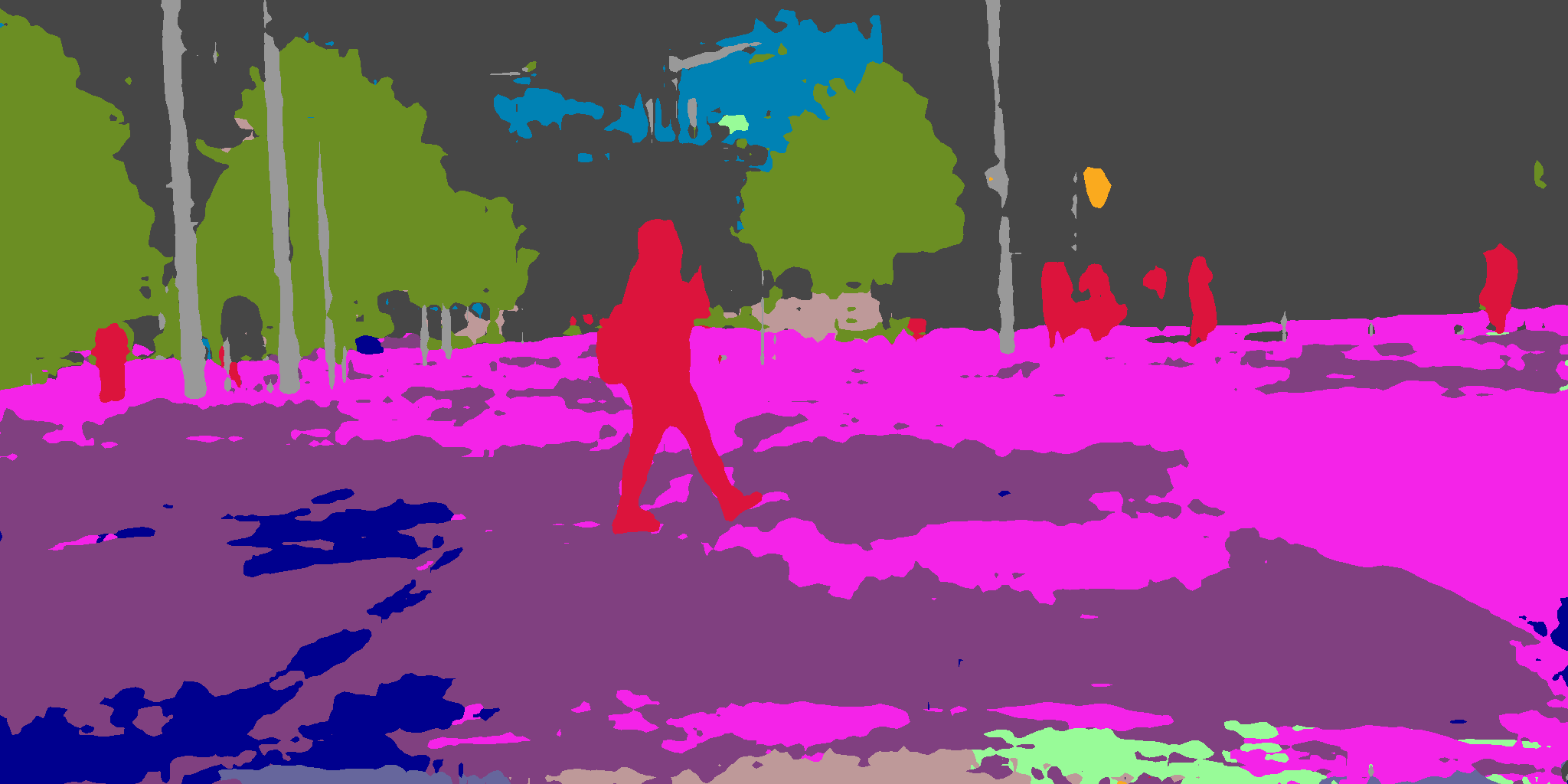}} &
         \subfloat{\includegraphics[width=0.2\linewidth]{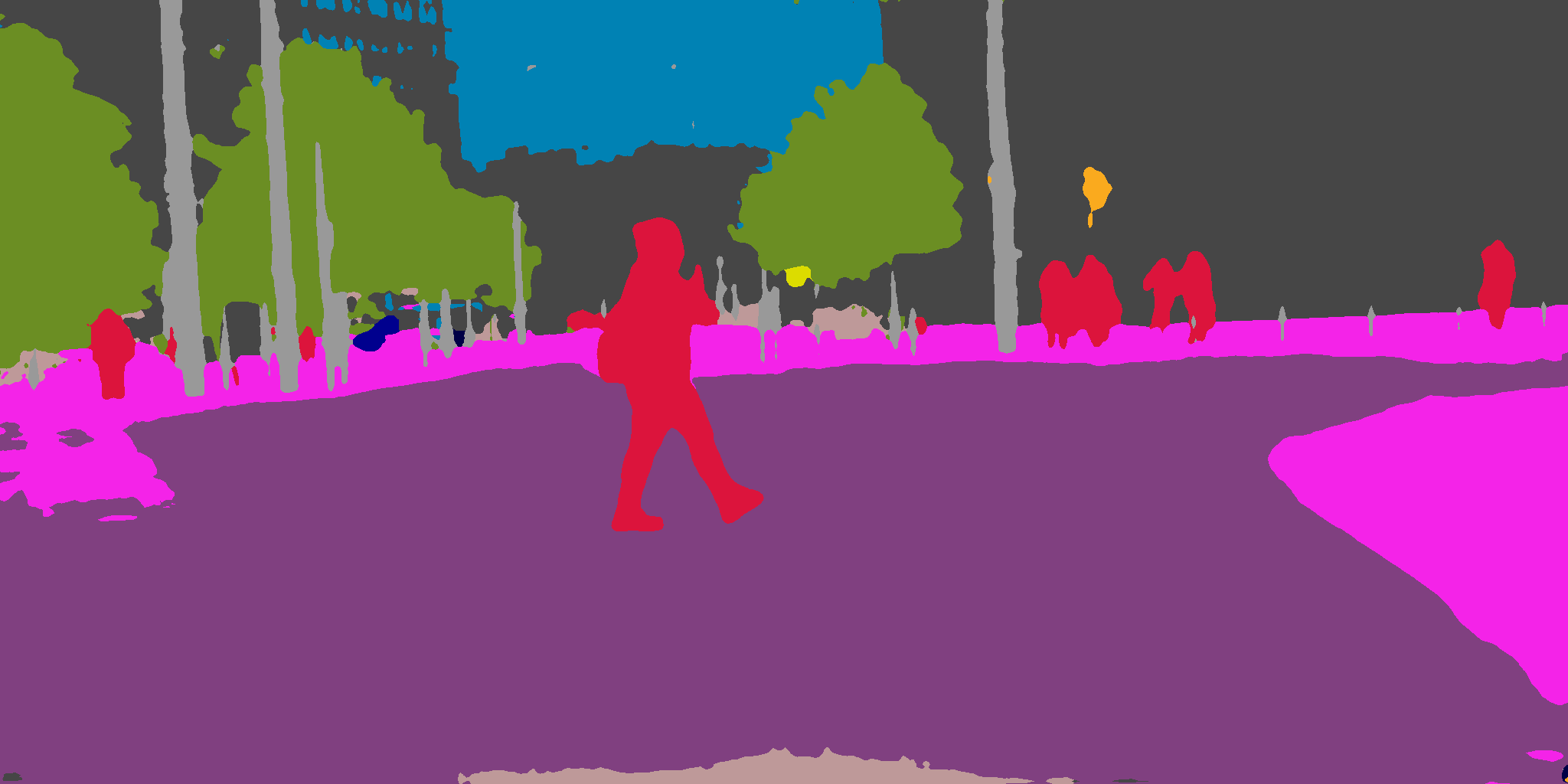}} &
         \subfloat{\includegraphics[width=0.2\linewidth]{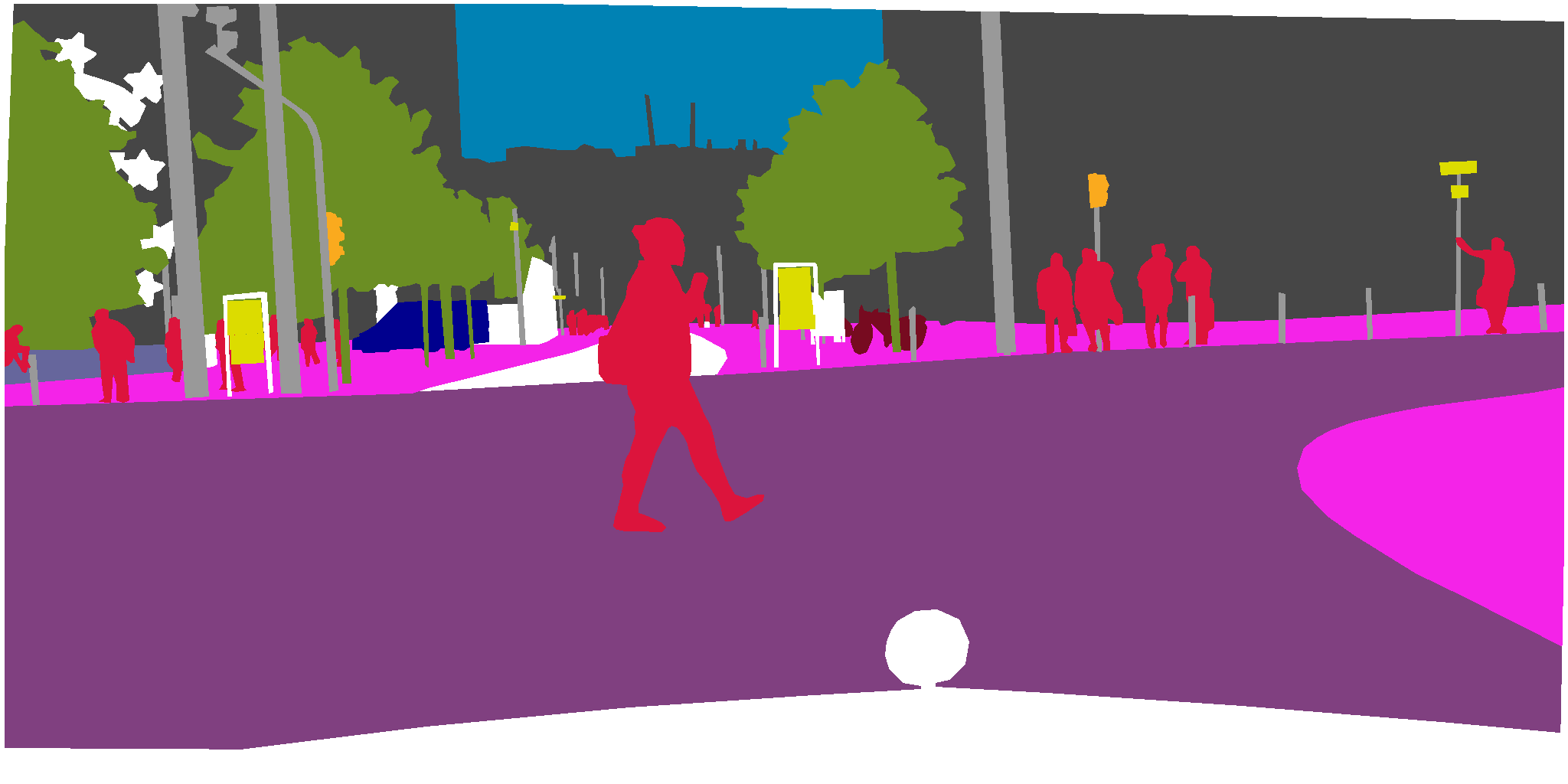}} \\
         
         \subfloat{\includegraphics[width=0.2\linewidth]{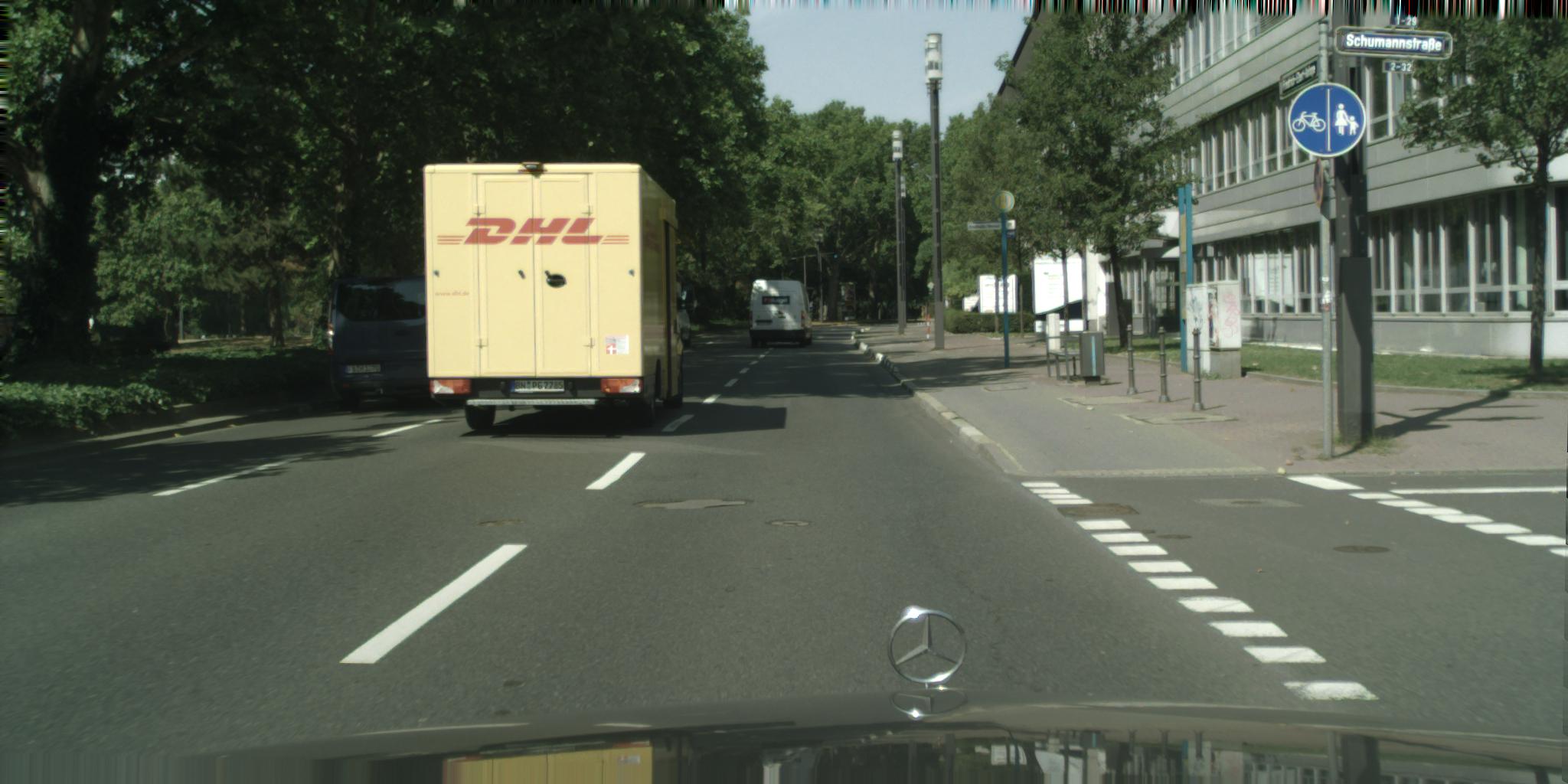}} &
         \subfloat{\includegraphics[width=0.2\linewidth]{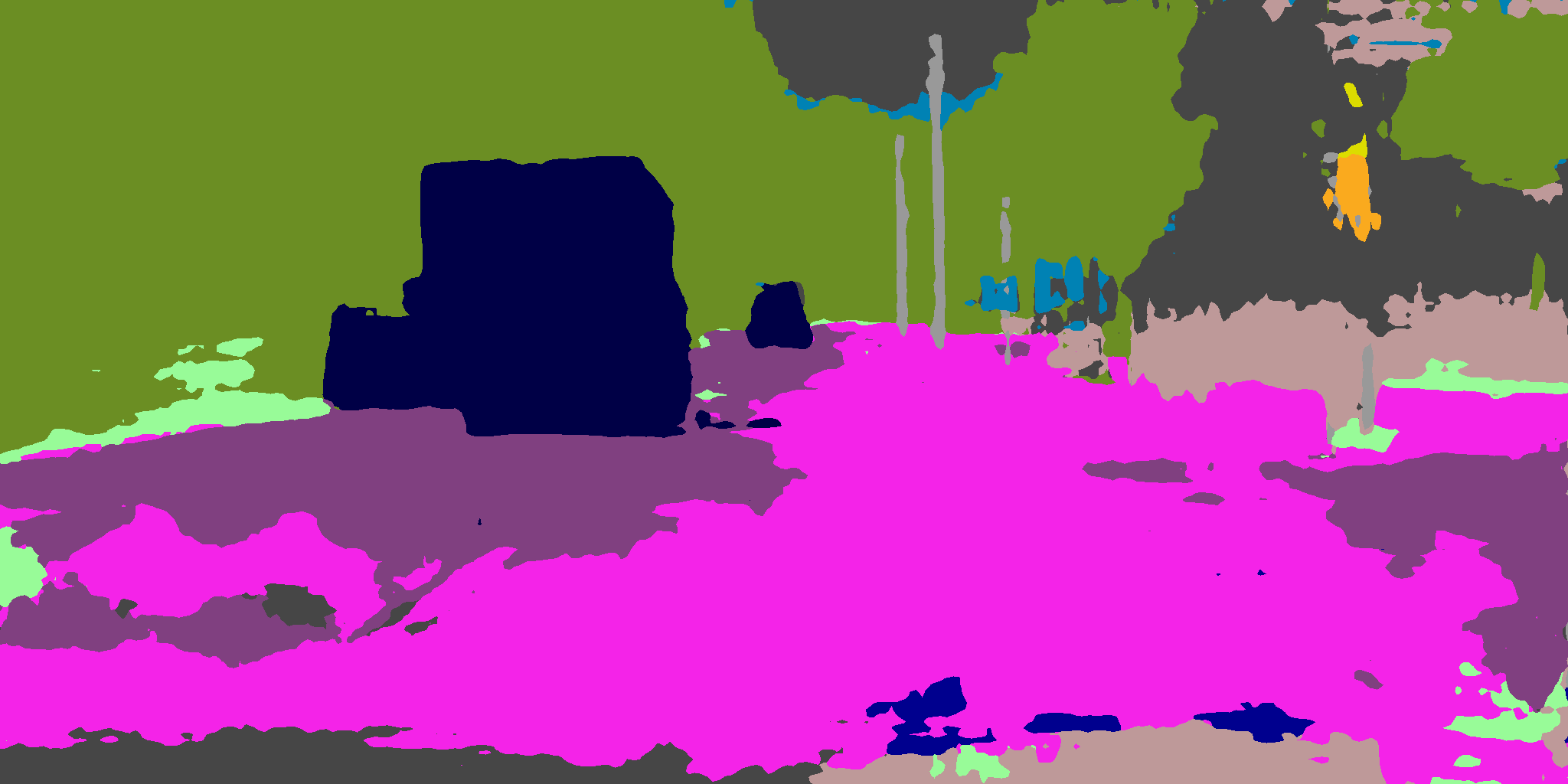}} &
         \subfloat{\includegraphics[width=0.2\linewidth]{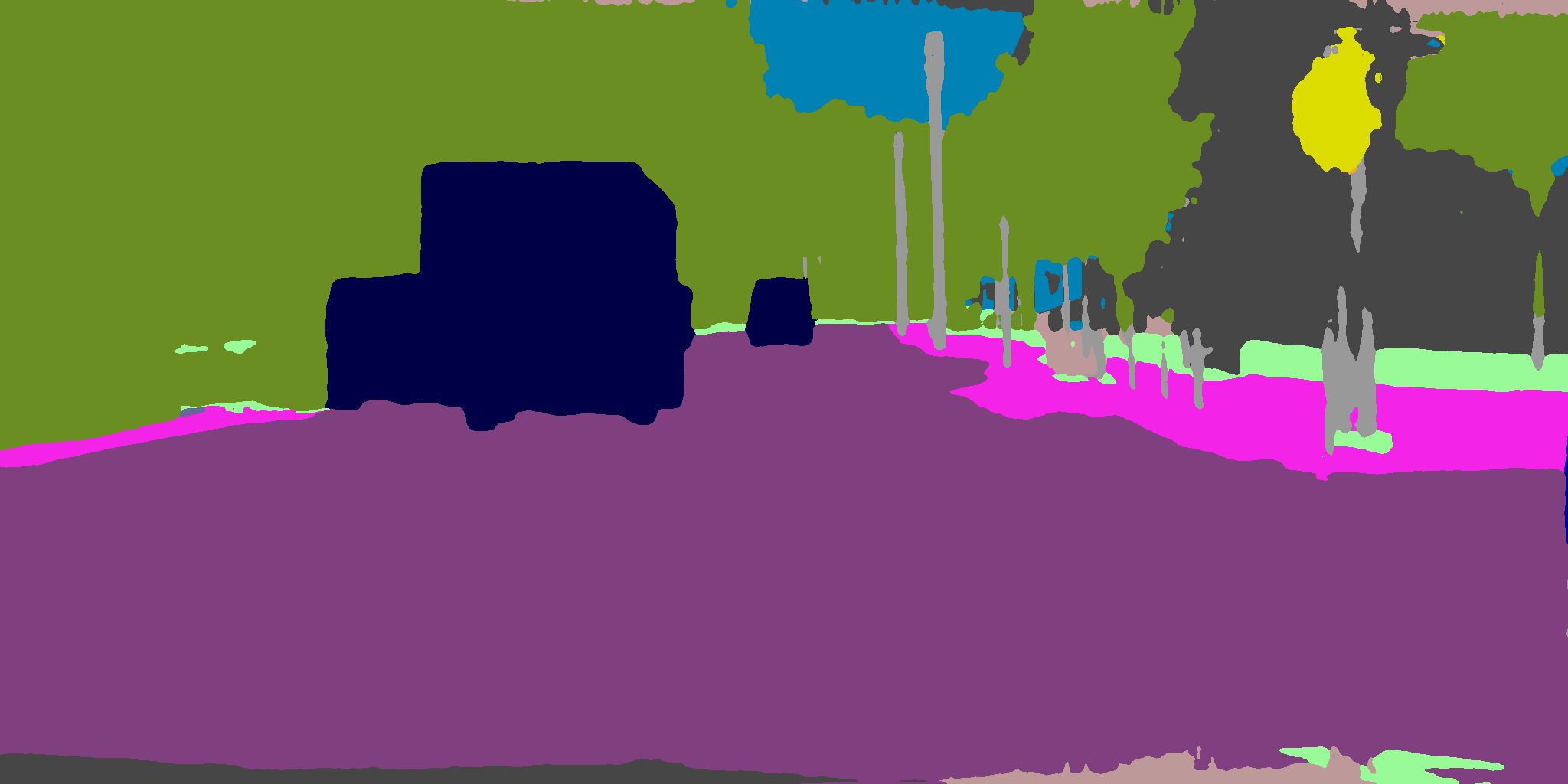}} &
         \subfloat{\includegraphics[width=0.2\linewidth]{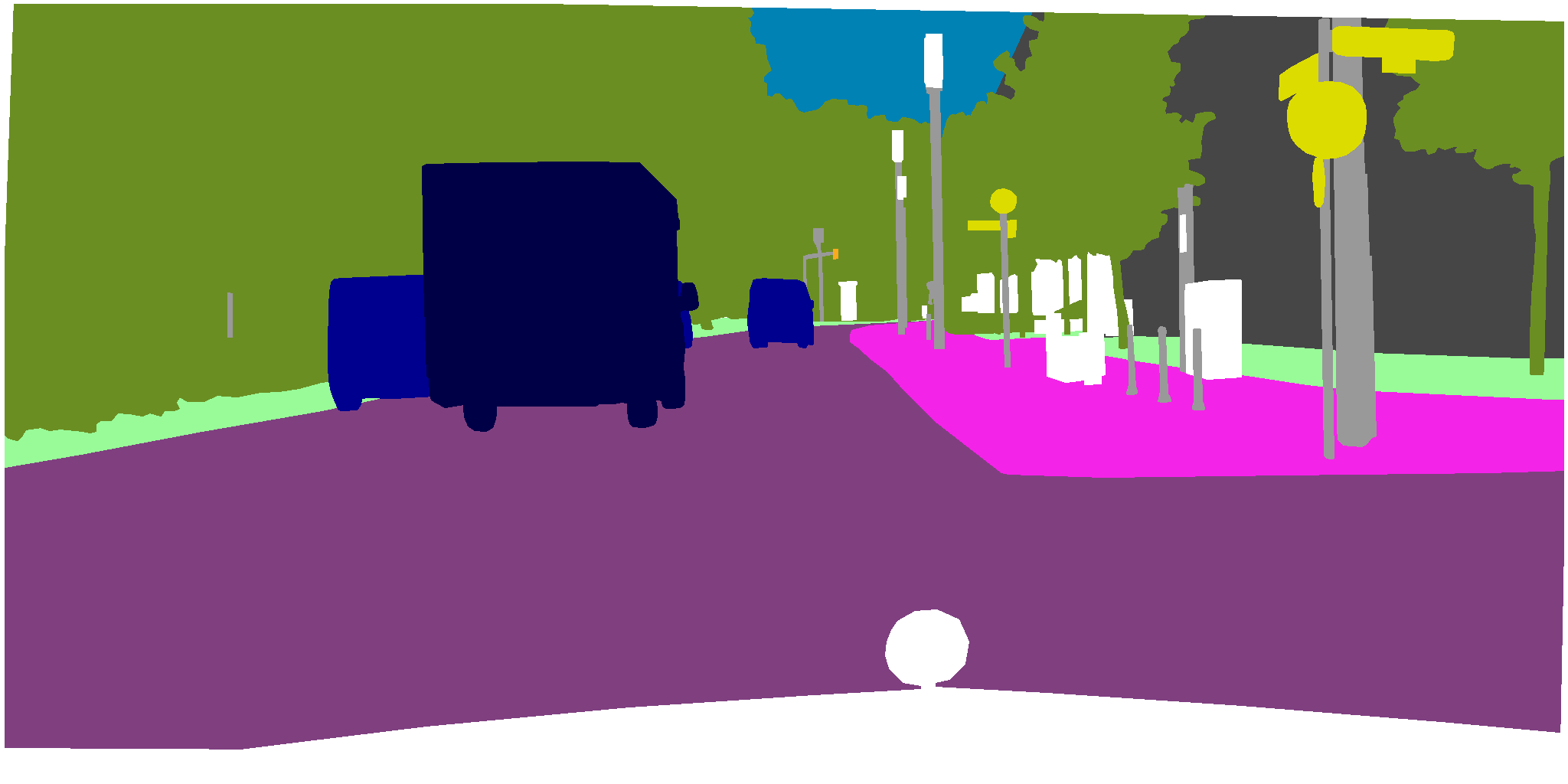}} \\
         
         Image & Before adaptation & After adaptation & Ground-truth
    \end{tabular}}
   \vspace{-0.2cm}
    \caption{Qualitative results of our proposed method.}
    \vspace{-0.5cm}
    \label{fig:examples}
\end{figure*}

\begin{figure}[t]
    \centering
    \begin{tabular}{cc}
         \subfloat{\includegraphics[width=0.45\linewidth]{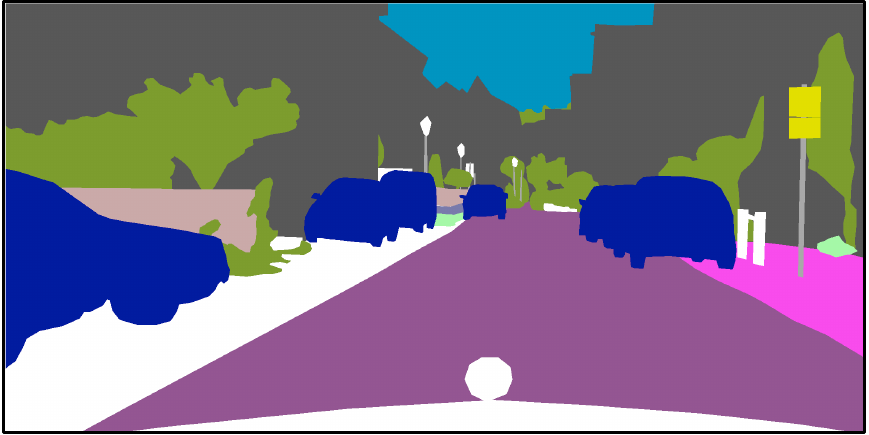}} &
         \subfloat{\includegraphics[width=0.45\linewidth]{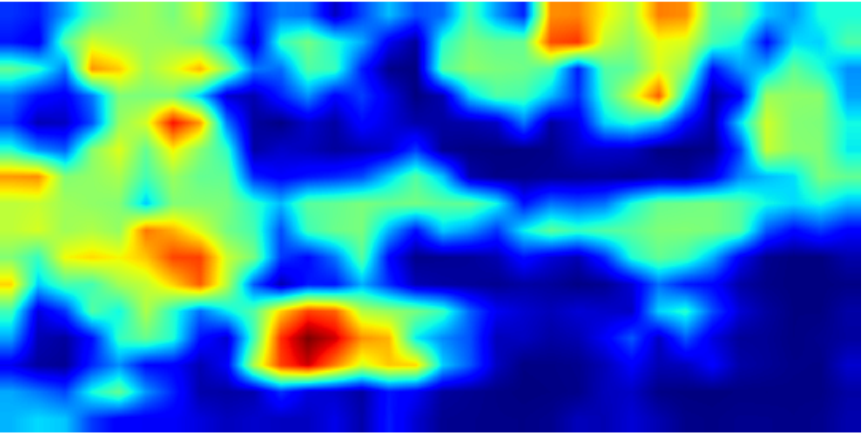}} \\
          Ground-truth & Adv-confidence \\
         \subfloat{\includegraphics[width=0.45\linewidth]{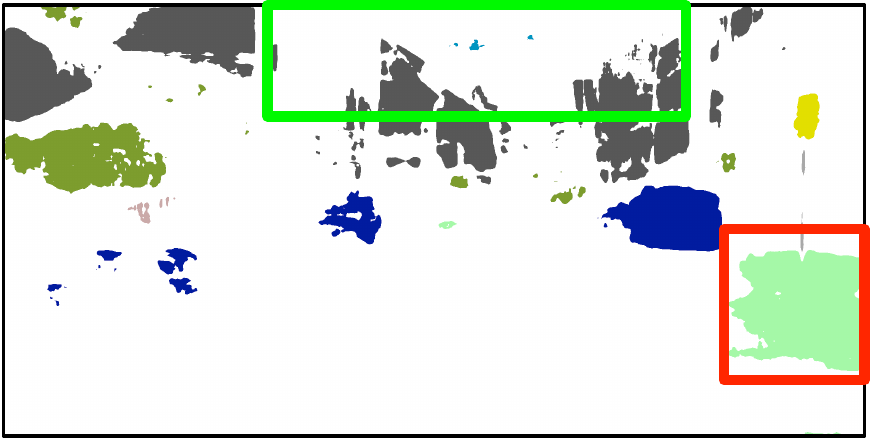}} &
         \subfloat{\includegraphics[width=0.45\linewidth]{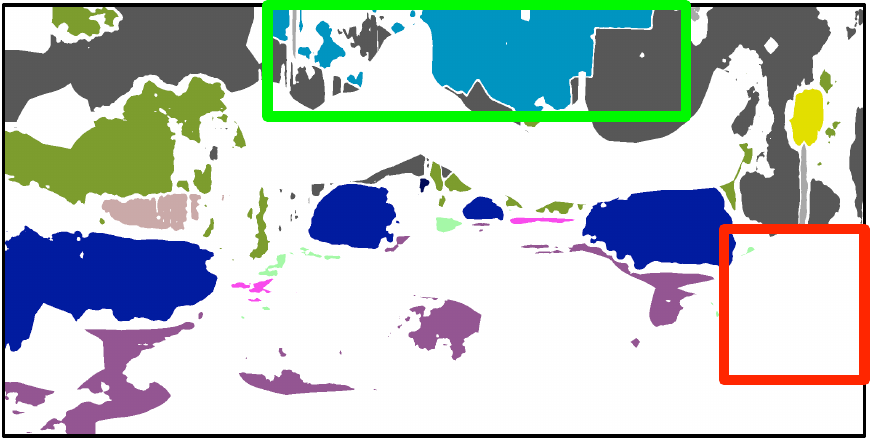}} \\
         Without adv-confidence & With adv-confidence
    \end{tabular}
 \vspace{-0.2cm}
    \caption{Comparison of the proxy with and without using adversarial confidence, which are shown in the bottom. The ground-truth (upper-left) is not available in our setting, which is only used for visualization purposes. The upper-right is the adversarial confidence map.}
    \vspace{-0.5cm}
    \label{fig:ablation}
\end{figure}

\subsection{Datasets}
In this section we consider two popular domain adaptation tasks, GTA5 \cite{richter2016playingGTA} to Cityscapes \cite{Cordts2016Cityscapes} and SYNTHIA \cite{ros2016synthia} to Cityscapes. Cityscapes is a real-world dataset containing high quality images of street scenes from various cities. Images are of size $2048\times1024$. There are 2,975 images for training and 500 images for validation. We use Cityscapes as the target dataset. Both GTA5 and SYNTHIA are synthetic datasets that are generated using computer graphics. GTA5 consists of 24,966 images with the resolution of $1914\times1024$. There are 19 valid classes that are consistent with the Cityscapes dataset. SYNTHIA contains 9,400 images with the resolution of $1280\times760$. We follow the protocal in works \cite{fcnwild, zhang2017curriculum}, which maps 16 common classes to Cityscapes. There is another setting that only takes 13 classes into consideration \cite{Tsai_adaptseg_2018}. These two synthetic datasets are used as the source dataset. We evaluate our method on these two tasks using mean Intersection over Union (mIoU) as the metric, which is a standard metric in semantic segmentation tasks. 

\subsection{Implementation Details}
Our method is implemented in tool box MXNet \cite{mxnet}. To make fair comparison, we adopt different backbones, \ie VGG16 \cite{vgg} and Resnet101 \cite{resnet}. We first train the model on the source dataset, with batch size of 4. The training is optimized with a SGD optimizer \cite{sgd} with momentum of 0.9, weight decay of $5e-4$. The initial learning rate is set to $4e-3$ and decreased by a poly policy. After 80k of iterations, we fine-tune the model with the target dataset and the discriminators involved. The discriminators are optimized with an Adam optimizer \cite{adam}. The learning rate is set to $1e-4$ for both optimizers and fix it during the fine-tuning. The training is performed for 10k iterations. When the pseudo labels are generated, we retrain the model using only the pseudo labels in the target domain. We train the model for 60k iterations with a poly decreasing learning rate starting at $1e-2$. We can also use the previous model as the initialization and fitune it, but we find it is better to train the model directly from the ImageNet pretrained one. 

To include more contextual information in the adversarial learning, we use almost the whole image for each sample. Since different datasets have different resolutions and ratios, for GTA5, we resize the short size to 720 and take the a crop $1280\times720$; for SYNTHIA, we keep the whole image at the original resolution; for Cityscapes, we resize them to $1024\times512$ and use the whole image. For the final training step, as there is only one dataset involved with the proxy labels, we adopt a more standard training protocol, using batch size of 16, crop size of $640\times640$ and random scale 0.7-1.3 of the original size as data augmentation.

\subsection{Overall Performance}
\textbf{GTA5 to Cityscapes.} The results are presented in Table \ref{tab:gta2cs}. Our method outperforms other methods by a large margin. For the VGG-based models, our model trained with source images reaches mIoU of $27.9\%$, which established a strong baseline over other methods. Using our domain adaptation method, we obtain mIoU of $39\%$, which is $3\%$ higher than the previous best model. Using Resnet101 as the backbone, we get mIoU of $38.1\%$ using only the source images, which is also a strong baseline. Adapted by our proposed method, the model reaches to $50.4\%$, that outperforms the previous best model by more than $3\%$. Note that the backbone used in \cite{cbst} ,Wideres38 \cite{wideres}, is more powerful than Resnet101, which has top-1 error of $19.2\%$ on ImageNet compared with $22.1\%$ achieved by Resnet101. Some qualitative results can be found in Figure \ref{fig:examples}.

\textbf{SYNTHIA to Cityscapes.} The results are shown in Table \ref{tab:syn2cs}. Since there are two different evaluation protocols, we use * to indicate the results excluding class Wall, Fence and Pole. Our VGG-based model, trained only with the source images, reaches mIoU of $20.6\%$ and mIoU* of $24.5\%$, which are a bit lower than the other source-only models. Adapted by our method, the model achieves mIoU of $36.8\%$ and $43.0\%$ outperforming the other VGG-based models. Our Resnet101-based model obtains mIoU of $33.4\%$ and mIoU* of $38.3\%$, which are comparable with other models. The adapted model reaches mIoU of $45.5\%$ and mIoU* of $52.3\%$, which are around $3\%$ improvement over the previous best model.

\subsection{Ablation Study}
\textbf{Analysis of different modules}
In this section, we study how each module affects the overall performance. The experiments are based on the Resnet101-based model. Table \ref{tab:ablation} shows the settings of all the experiments. \textbf{Adv} stands for Adversarial training. When multiple discriminators are involved, it is named \textbf{Mul-Adv}. \textbf{Proxy} means the model is trained with proxy labels. Multi-scale inference is utilized as a post-processing step annotated as \textbf{MS}. Starting with the model trained in the source domain without any modules, it reaches mIoU of $38.1\%$. When a single discriminator is added to train the model in an adversarial fashion, it brings the performance to $41.5\%$. By using two discriminators for different levels of feature alignment, we obtain a score of $42.7\%$. We also try adding more discriminators or putting the discriminators at different positions of the feature backbone, but we observe no significant increase in terms of the mIoU. Trained with the proxy labels, the model with a single discriminator achieves the result of 47.1 while the one with an extra auxiliary discriminator reaches to $49.6\%$. The multi-scale inference further increases the result to $50.4\%$.

\begin{table}[htp]
    \centering
    \scalebox{0.796}
    {
        \begin{tabular}{ c|c|c|c|c }
\hline 
\textbf{Adv} & \textbf{Mul-Adv} & \textbf{Proxy} & \textbf{MS} & mIoU\tabularnewline
\hline 
\hline 
 &  &  &  & 38.1\tabularnewline
\hline 
\checkmark &  &  &  & 41.5\tabularnewline
\hline 
 & \checkmark &  &  & 42.7\tabularnewline
\hline 
\checkmark &  & \checkmark &  & 47.1\tabularnewline
\hline 
 & \checkmark & \checkmark &  & 49.8\tabularnewline
\hline 
 & \checkmark & \checkmark & \checkmark & 50.4\tabularnewline
\hline 
\end{tabular}
}
    \caption{Ablation studies of each modules.}
    \vspace{-0.2cm}
    \label{tab:ablation}
\end{table}

\begin{table}[htp]
    \centering 
        \scalebox{0.806}{
        \begin{tabular}{ p{1cm}|p{1cm}|p{1cm}|p{1cm} }
\hline 
$p_1$ & $p_2$ & $p_1\cdot p_2$ & mIoU\tabularnewline
\hline 
0.5 & 0.4 & 0.2 & 46.7\tabularnewline
0.5 & 0.6 & 0.3 & 49.3\tabularnewline
0.5 & 0.8 & 0.4 & 49.7\tabularnewline
\hline 
0.6 & 0.4 & 0.24 & 47.0\tabularnewline
0.6 & 0.5 & 0.3 & 47.9\tabularnewline
0.6 & 0.6 & 0.36 & 49.4\tabularnewline
0.6 & 0.8 & 0.48 & \textbf{49.8}\tabularnewline
\hline 
0.8 & 0.3 & 0.24 & 46.1\tabularnewline
0.8 & 0.4 & 0.32 & 46.9\tabularnewline
0.8 & 0.5 & 0.4 & 47.9\tabularnewline
\hline 
1 & 0.2 & 0.2 & 45.8\tabularnewline
1 & 0.3 & 0.3 & 46.3\tabularnewline
1 & 0.4 & 0.4 & 47.0\tabularnewline
1 & 0.48 & 0.48 & 47.7\tabularnewline
\hline 
\end{tabular}
        }
    \caption{Analysis of the effect of different $p_1$ and $p_2$. When the total proportion of labels is fixed (e.g. $p_1\times p_2=0.3$ or $0.48$), using adversarial confidence (e.g. $p_1=0.6$) and not using adversarial confidence ($p_1=1$) achieve very different results ($46.3$ vs $47.9$ and $47.7$ vs $49.8$). The optimal result is obtained when $p_1=0.6$ and $p_2=0.8$.}
    \vspace{-0.6cm}
    \label{tab:portion}
\end{table}
\textbf{Analysis of proxy generation}
The process of generating proxy labels plays an important role in our approach. We discuss in detail about how the discriminators help select more reliable labels both qualitatively and quantitatively. Figure \ref{fig:ablation} shows an example of different selection strategies. There is a ground-truth mask in the top-left. Please note the ground-truth is not available during the training because it is an image from the target domain, therefore we only use it for visualization purposes to show the quality of our proxy label. In the top-right is the attention map produced by the discriminators that shows how likely the region is from the source domain. In the bottom are two proxy masks generated without and with the attention of the discriminators. The left one is generated using the similar strategy used in \cite{cbst} where pixels are sorted based on the classifier's confidence and kept based a certain portion. The right one is obtained by taking the adversarial attention into consideration because we are more interested in the regions that look more similar to the source domain. The red bounding boxes indicate a region where the classifier has high confidence but it turns out to be bad predictions. With the help of adversarial attention, we could ignore this part and focus more on other reliable regions, such as the region highlighted by the green bounding boxes. 

In the process of proxy label generation, there are two parameters $p_1$ and $p_2$ that control how the labels are selected using the multi-adversarial confidence and classification confidence. Table \ref{tab:portion} describes how the two parameters affect the result. In the third column, $p_1 \cdot  p_2$ indicates the total proportion of the selected pixels. When $p_1=1$, it basically means the adversarial confidence is not used in the picking process. It can be easily seen that when the total proportion of labels is decided (e.g. $p_1\times p_2=0.3$ or $0.48$), using adversarial confidence ($p_1=0.6$) and not using adversarial confidence ($p_1=1$) achieve very different results ($46.3$ vs $47.9$ and $47.7$ vs $49.8$). When $p_1$ is high (e.g. $0.8$ or $1$) indicating that we do not use much information from the adversarial confidence, we get unsatisfactory results, which further shows the importance of using adversarial confidence. The optimal result is obtained when $p_1=0.6$ and $p_2=0.8$.

\section{Conclusion}
In this paper, we propose a novel proxy-based method for unsupervised domain-adaptive semantic segmentation. The core of our method is the generation of high quality proxy labels for the unlabelled target domain. We first train a network in an multi-adversarial style where multiple discriminators are involved. The discriminators not only work as a regularizer to encourage feature alignment, but also provide an alternative signal for sample selection. In the proxy generation process, our algorithm takes multiple confidence maps as inputs, the one provided by the semantic classifier and multiple maps of the discriminators, and selects more confident and reliable predictions as the proxy labels. Relying on the high-quality proxies, our model can be largely boosted. 
{\small
\bibliographystyle{ieee}
\bibliography{egbib}
}

\end{document}


\title{Regularizing Proxies with Multi-Adversarial Training for Unsupervised Domain-Adaptive Semantic Segmentation -- Supplementary Material}

\maketitle

\section{More Details of Proxy Generation}
In Section 2.3 of the main paper, we introduce and discuss the propose proxy generation algorithm with the regularization of multi-adversarial training. 
We provide more details of the proposed proxy generation algorithm in the following. 
Specifically, we summarize the whole proxy generation method (described in Section 2.3) in Algorithm \ref{alg:proxy_gen}. In Algorithm \ref{alg:proxy_gen}, the all operations in Section 2.3 are covered. 
Note that $\{\bD_1^{(n)}\}$ and $\{\bD_2^{(n)}\}$ are the scoremaps with the same size to the images.
In the last step, we obtain the proxy labels for all target images, which is equivalent to the operation defined in Eq. (5) in the main paper.

\begin{algorithm}[htp]
\KwData{Predicted scoremaps for all target images $\{\mathbf{P}^{(n)}\}$; the output of the two discriminators (\ie $\cD_1(\cdot)$ and $\cD_2(\cdot)$) $\{\bD_1^{(n)}\}$ and $\{\bD_2^{(n)}\}$; proportion parameters $p_1$, $p_2$.}
\KwResult{Proxy labels $\{\widehat{\mathbf{Y}}^{(n)}\}$ for all target images.}
-Obtaining \textbf{adversarial confidence} $\bA^{(n)}$:\\
\For{$n=1$ to $N$}{
$\bA^{(n)}$ = (minmaxNorm($\bD_1^{(n)}$) + minmaxNorm($\bD_2^{(n)}$)) / 2\;
}
-\textbf{Confidence refocusing:}\\
Sort the all elements in $[\text{vec}(\bA^{(1)}),...,\text{vec}(\bA^{(N)})]$ and find the $p_1$-th percentile $t_1$\;
\For{$n=1$ to $N$}
{
    Obtain categorical prediction: $\bP_C^{(n)}$ = argmax($\bP^{(n)}$)\;
    Obtain prediction confidence: $\bM_C^{(n)}$ = max($\bP^{(n)}$)\;
    \For{$l=1$ to $L$}
    {
        $\bV=\bM_C^{(n)}(\bP_C^{(n)}=l$ and $\bA^{(n)}>t_1)$\;
        $\bM_l$ = [$\bM_l$, vec($\bV$)]\;
    }
}
-\textbf{Proxy generation with class-balance reweighting:}\\
\For{$l=1$ to $L$ (for all categories)}
    {
        Sort $\bM_l$ and find $p_2$-th percentile $\bt_2^{(l)}$
    }
\For{$n=1$ to $N$ (for all samples)}
{
    \textbf{Scoremap reweighting}: $\widetilde{\bP}^{(n)} = \bP^{(n)}/\bt_2^{(n)}$\;
    Obtain categorical prediction: $\widetilde{\bP}_C^{(n)}$ = argmax($\widetilde{\bP}^{(n)}$)\;
    Obtain prediction confidence: $\widetilde{\bM}_C^{(n)}$ = max($\widetilde{\bP}^{(n)}$)\;
    Obtaining proxy labels (equiv. to Eq. (5) in the main paper): $\widehat{\bY}^{(n)} = \text{Onehot}( \widetilde{\bP}_C^{(n)}(\widetilde{\bM}_C^{(n)}>1$ and $\bA^{(n)}>t_1) )$\;
    
}
\caption{The proposed proxy generation algorithm.}
\label{alg:proxy_gen}
\end{algorithm}


    

\newpage
\section{Visualizations of More Examples in Experiments}
In this section, we visualize more examples in experiments to show the effectiveness of the proposed method. 

\par
In Figure \ref{fig:proxy_exps}, we show the examples of the proxy labels generated with or without the adversarial confidence. As shown in Figure \ref{fig:proxy_exps}, even without the adversarial confidence (referred to as adv-conf in the figure), the method can generate satisfactory proxy labels for the categories with large areas and small domain variance, such as ``road''. However, the method without adv-conf cannot obtain high-quality proxies on the small-scale categories, such as ``car''. Benefiting from the adv-conf and the corresponding ``refocusing'' and ``reweighting'' operations, the proposed method can obtain high-quality and class-balanced proxy labels. 

\par
\begin{figure*}[t]
    \centering
    \scalebox{0.96}{
    \begin{tabular}{cccc}
         \subfloat{\includegraphics[width=0.2\linewidth]{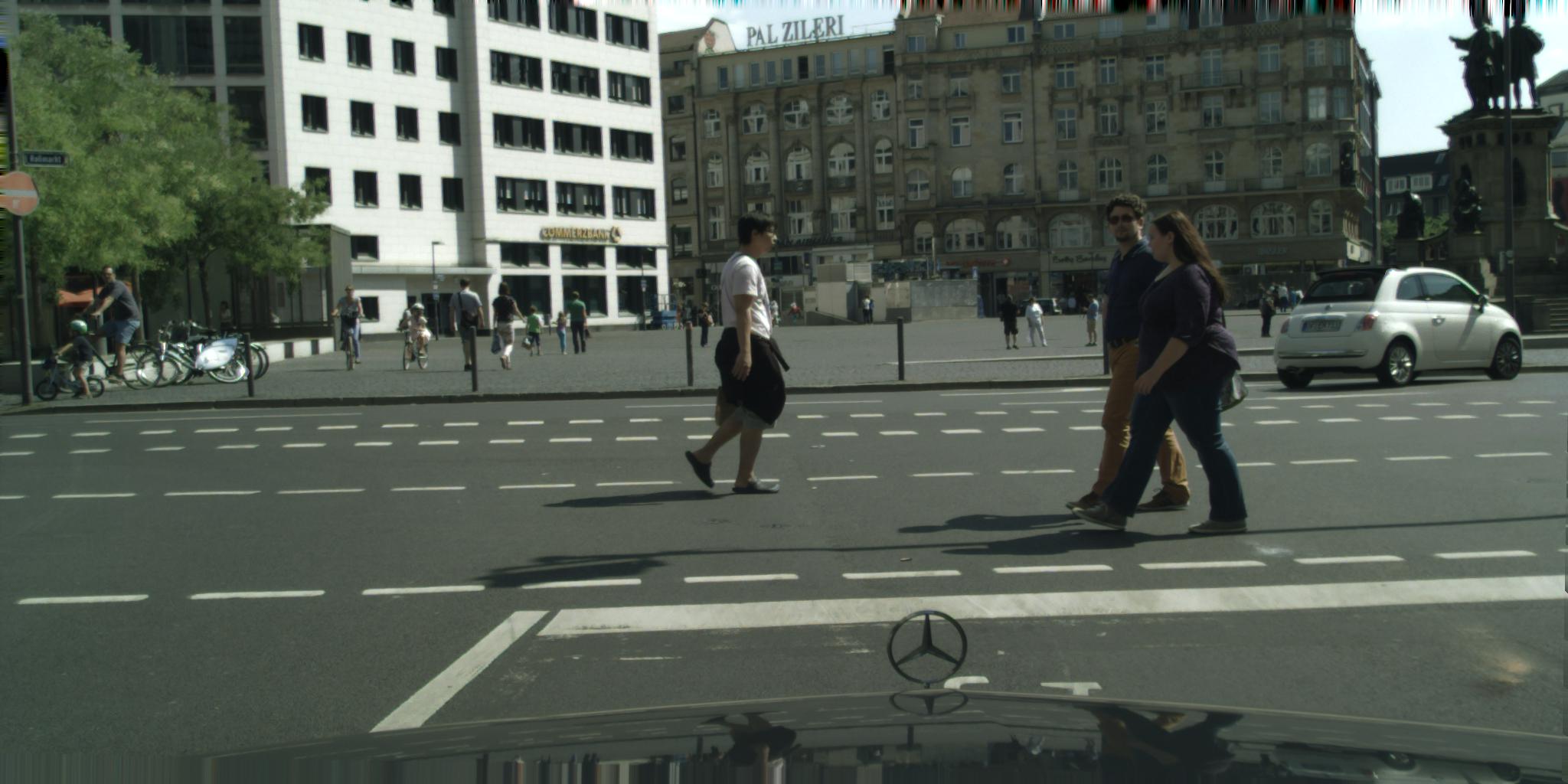}} &
         \subfloat{\includegraphics[width=0.2\linewidth]{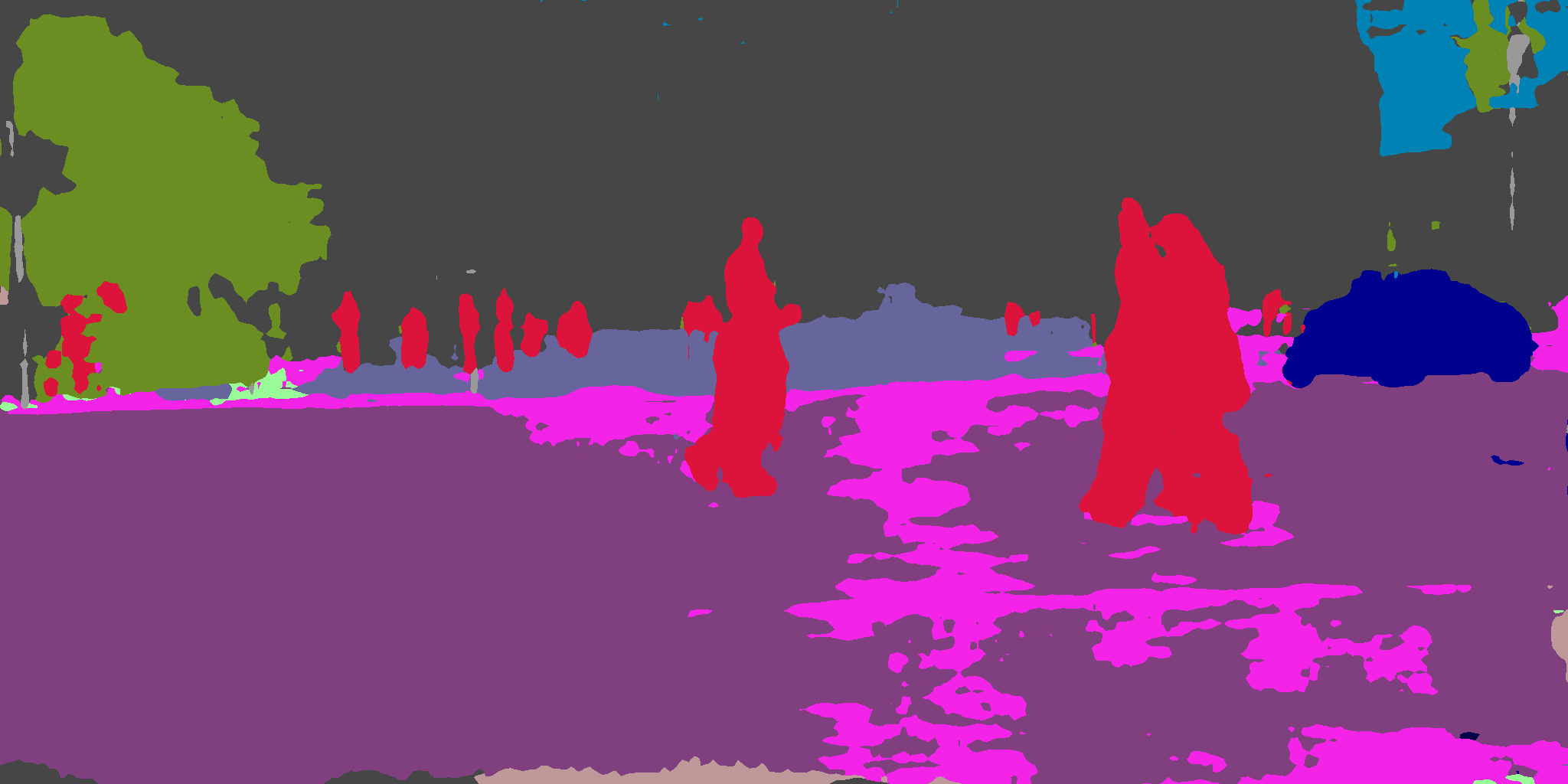}} &
         \subfloat{\includegraphics[width=0.2\linewidth]{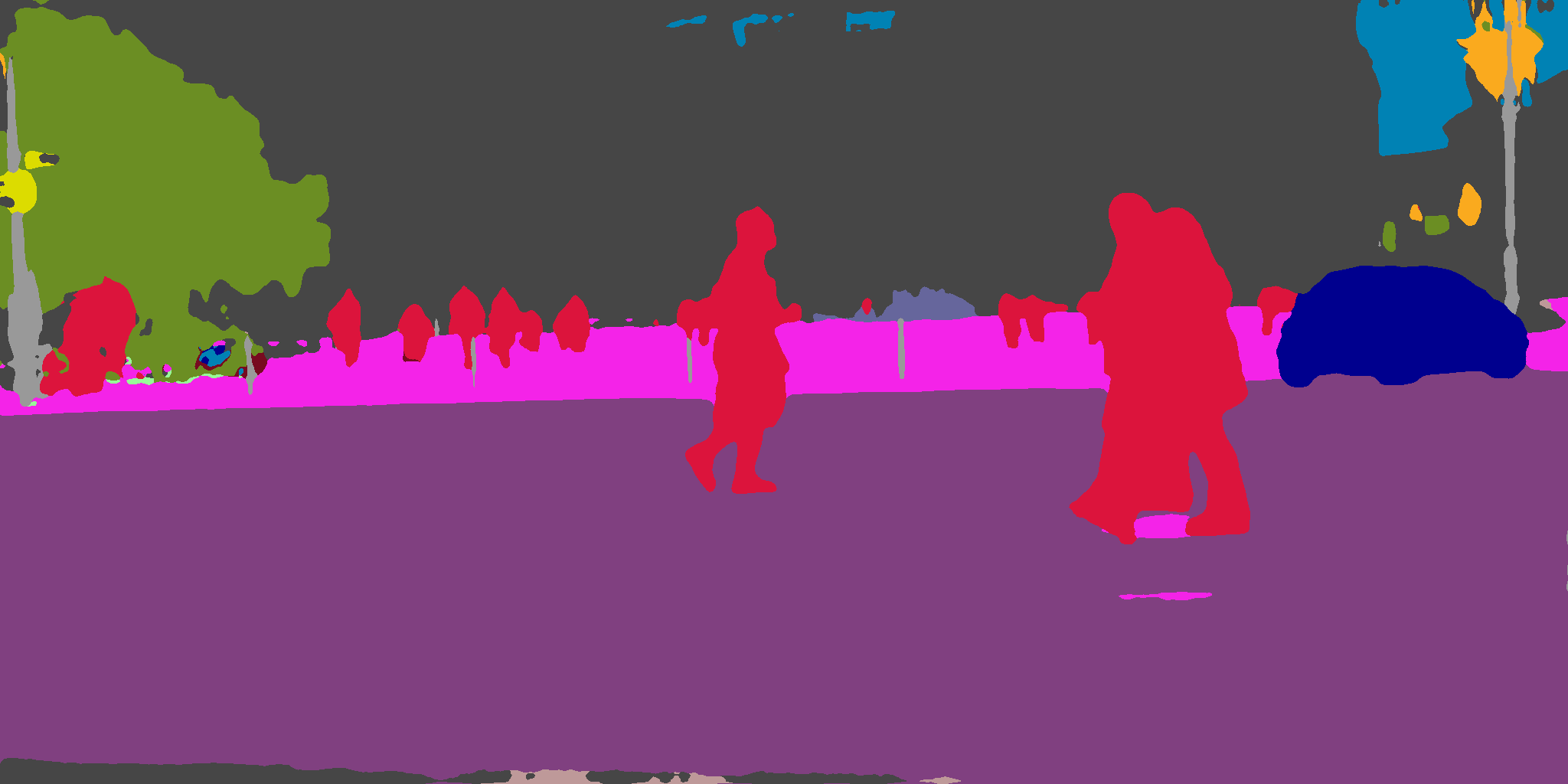}} &
         \subfloat{\includegraphics[width=0.2\linewidth]{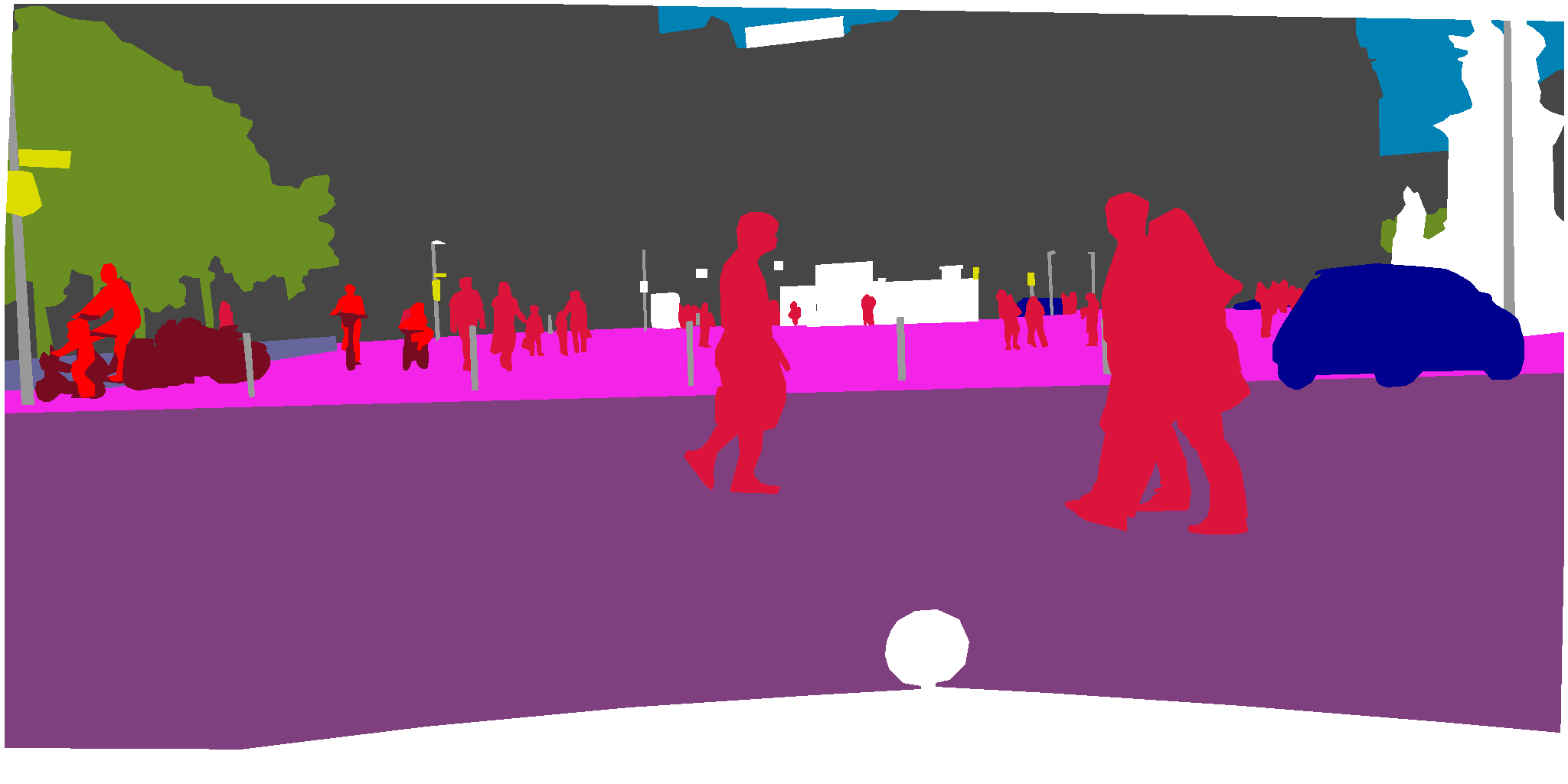}} \\
         
         \subfloat{\includegraphics[width=0.2\linewidth]{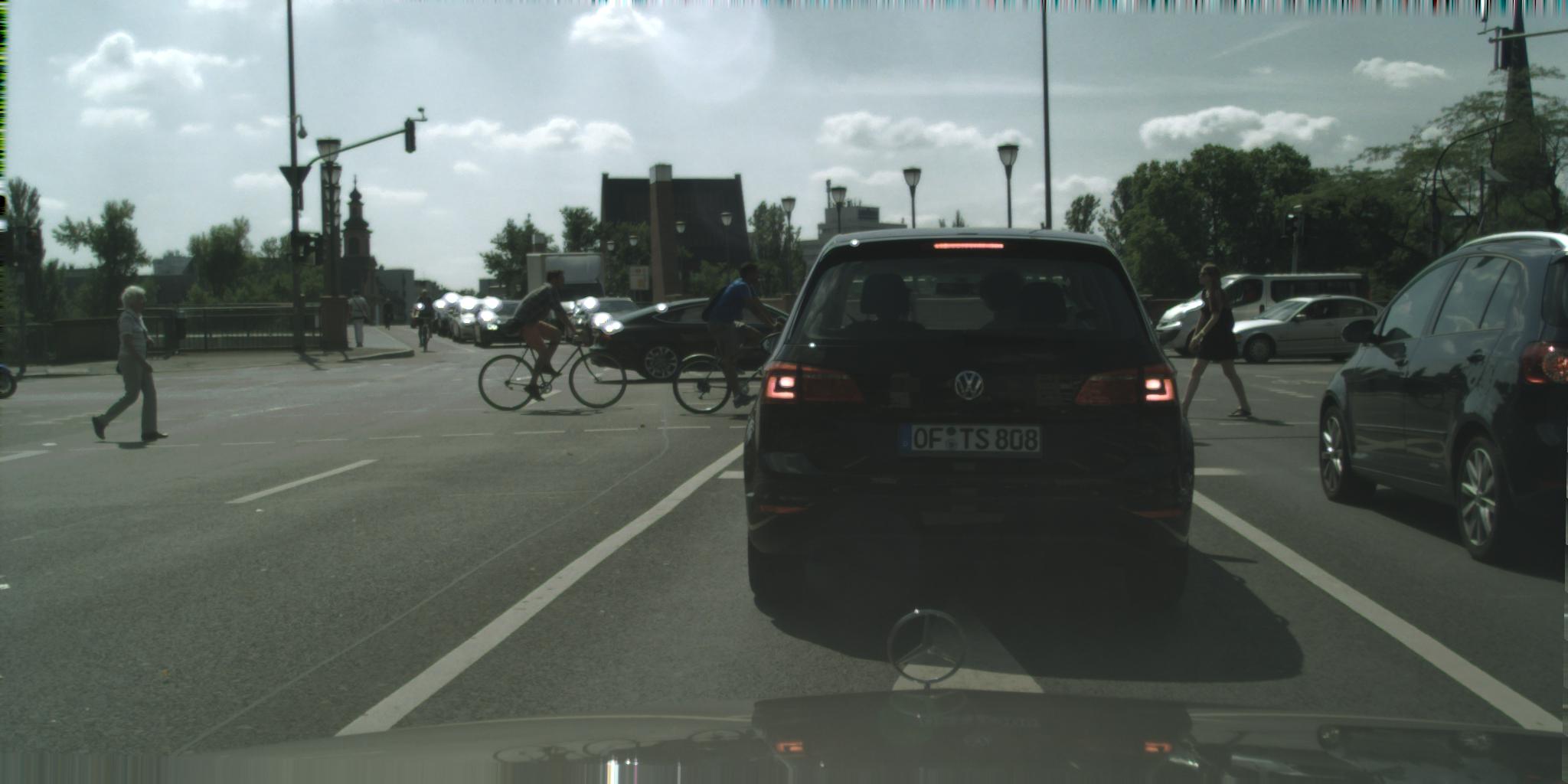}} &
         \subfloat{\includegraphics[width=0.2\linewidth]{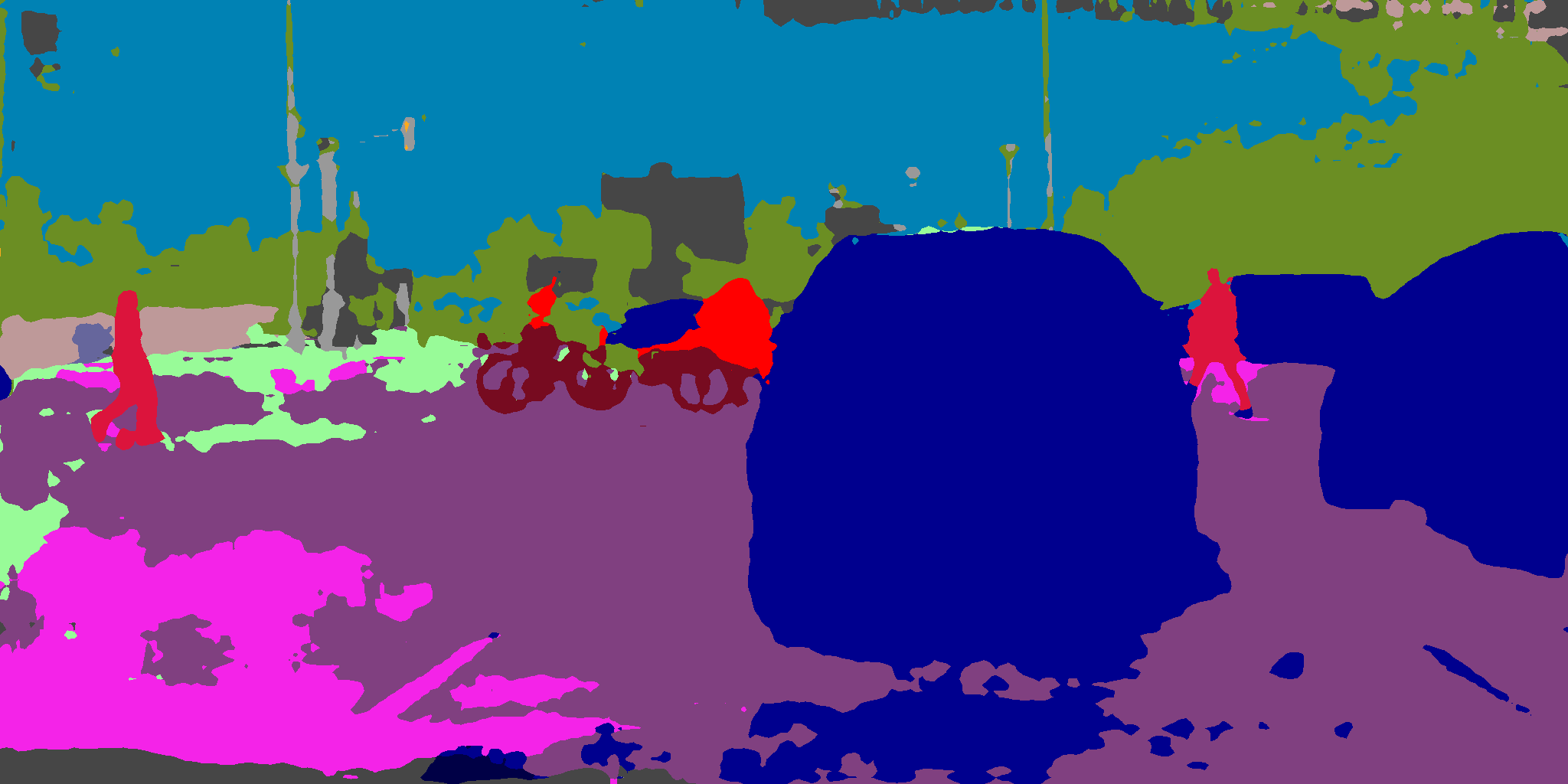}} &
         \subfloat{\includegraphics[width=0.2\linewidth]{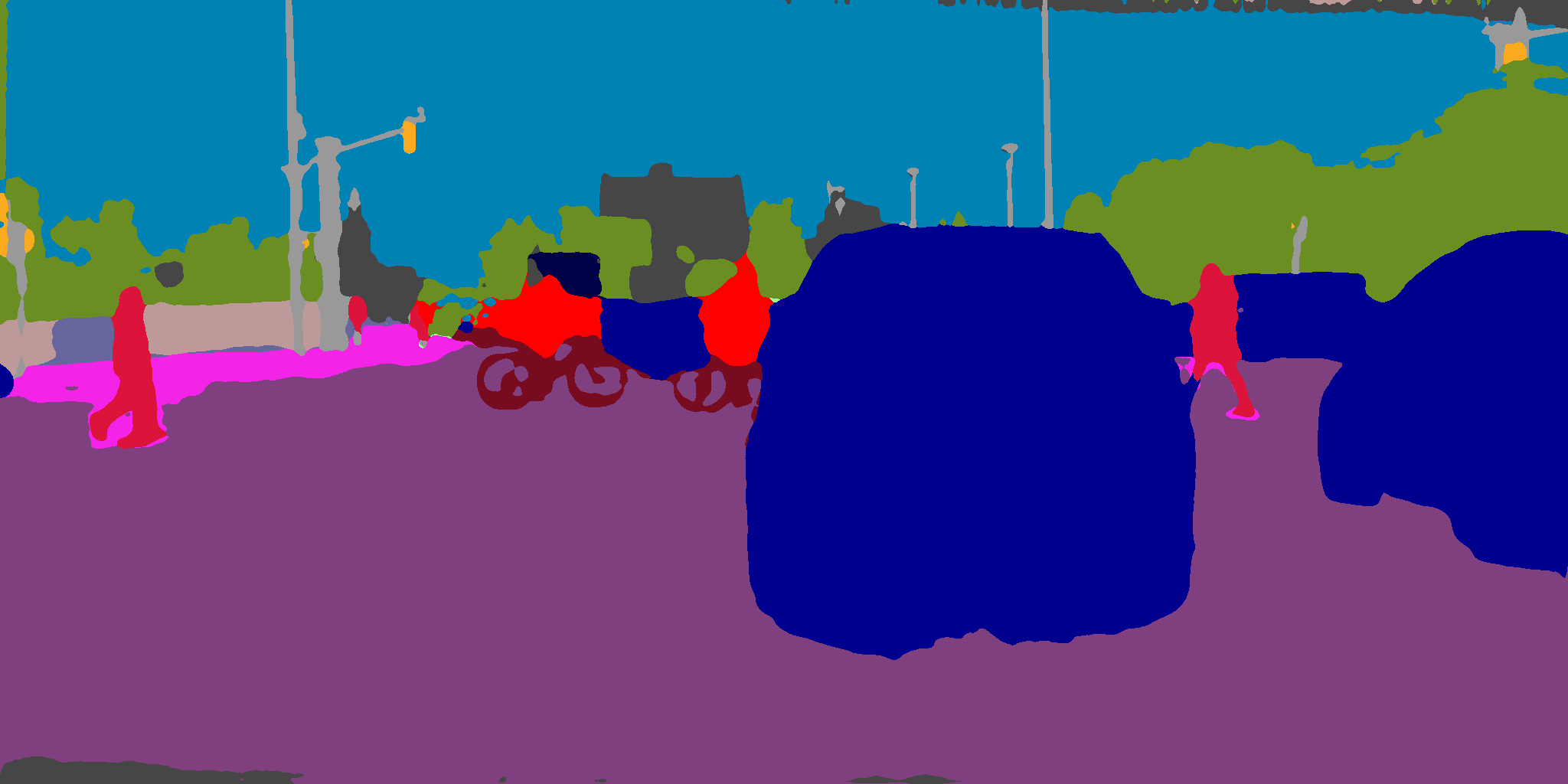}} &
         \subfloat{\includegraphics[width=0.2\linewidth]{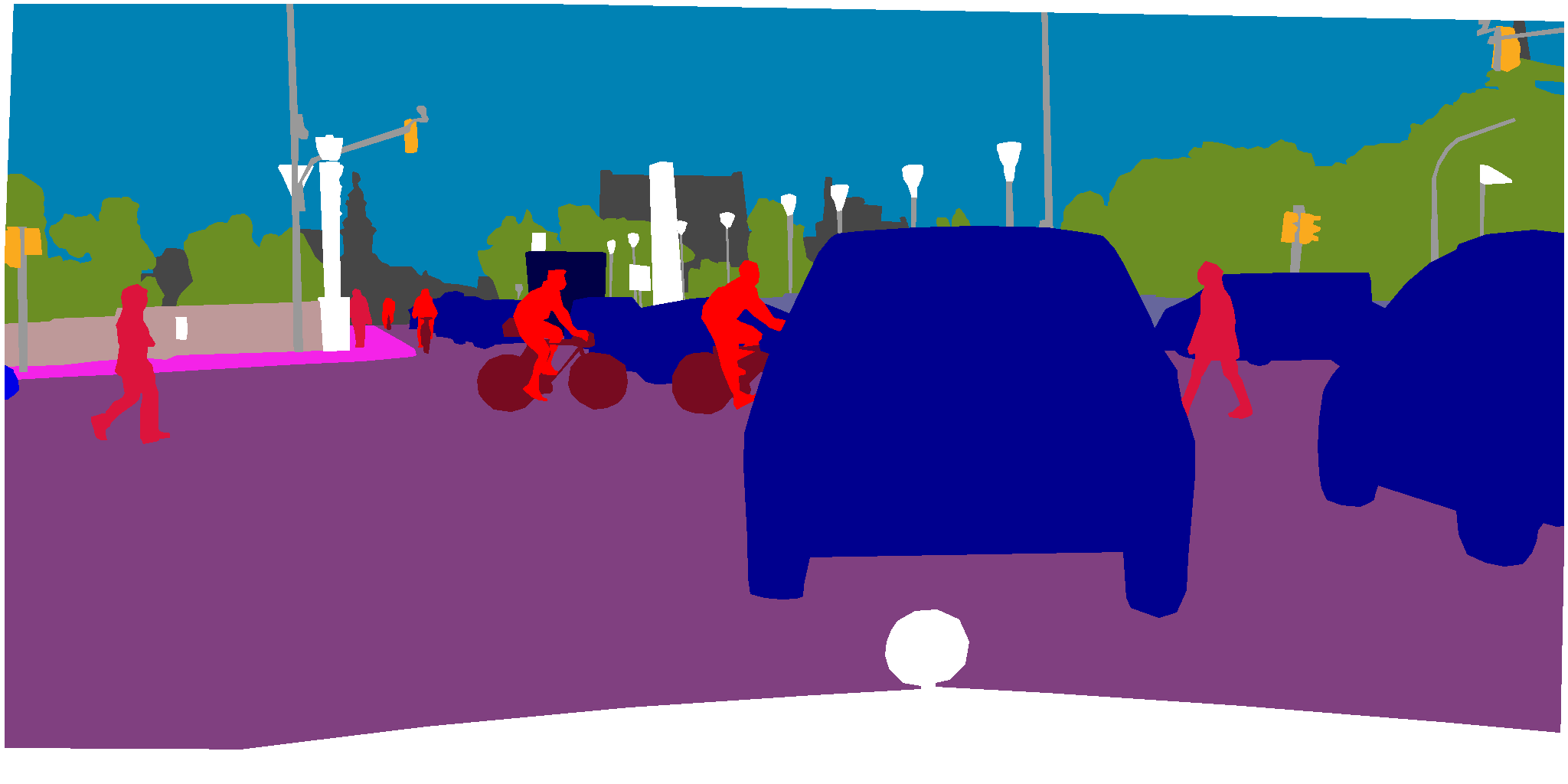}} \\
         
         \subfloat{\includegraphics[width=0.2\linewidth]{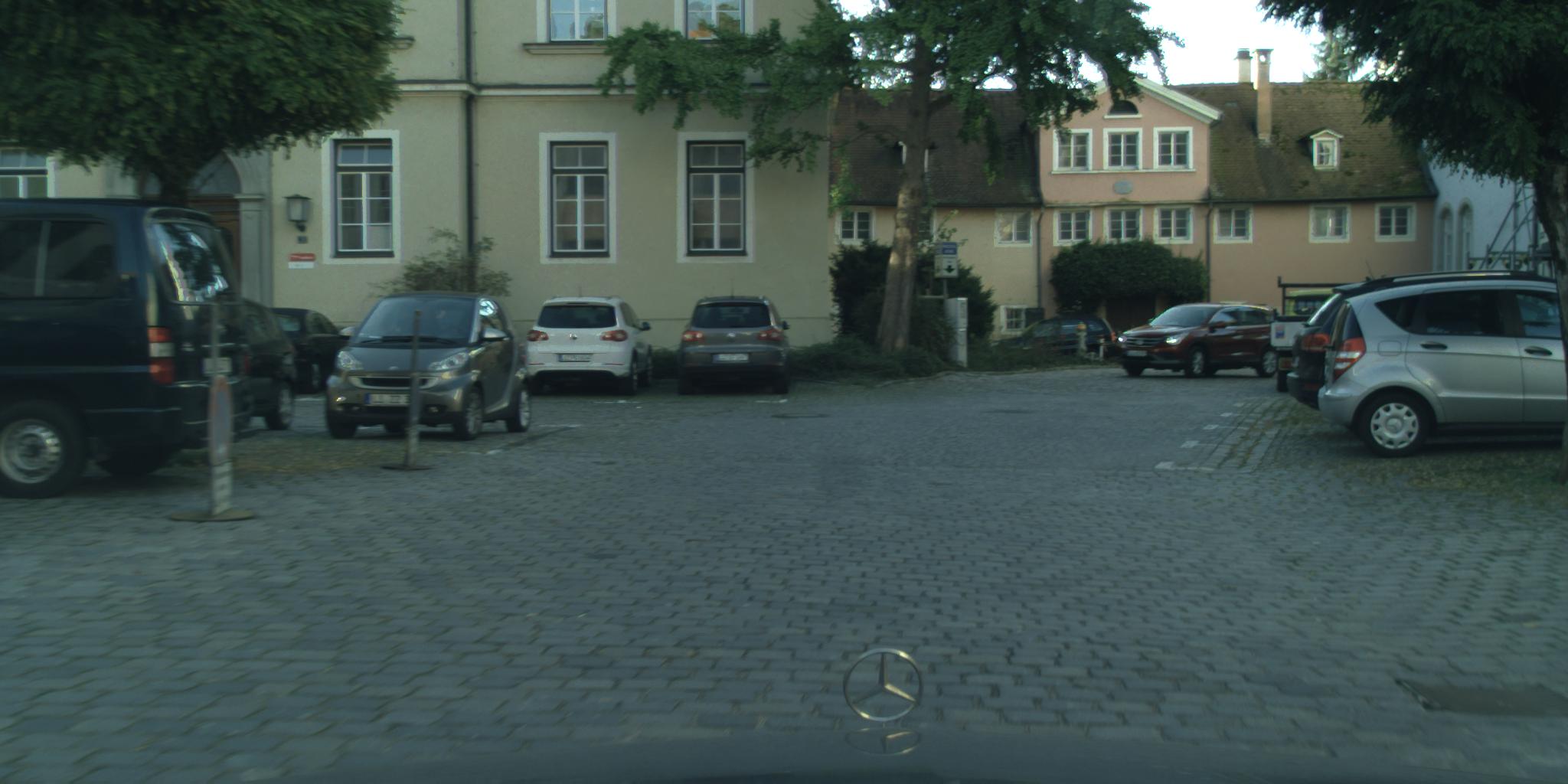}} &
         \subfloat{\includegraphics[width=0.2\linewidth]{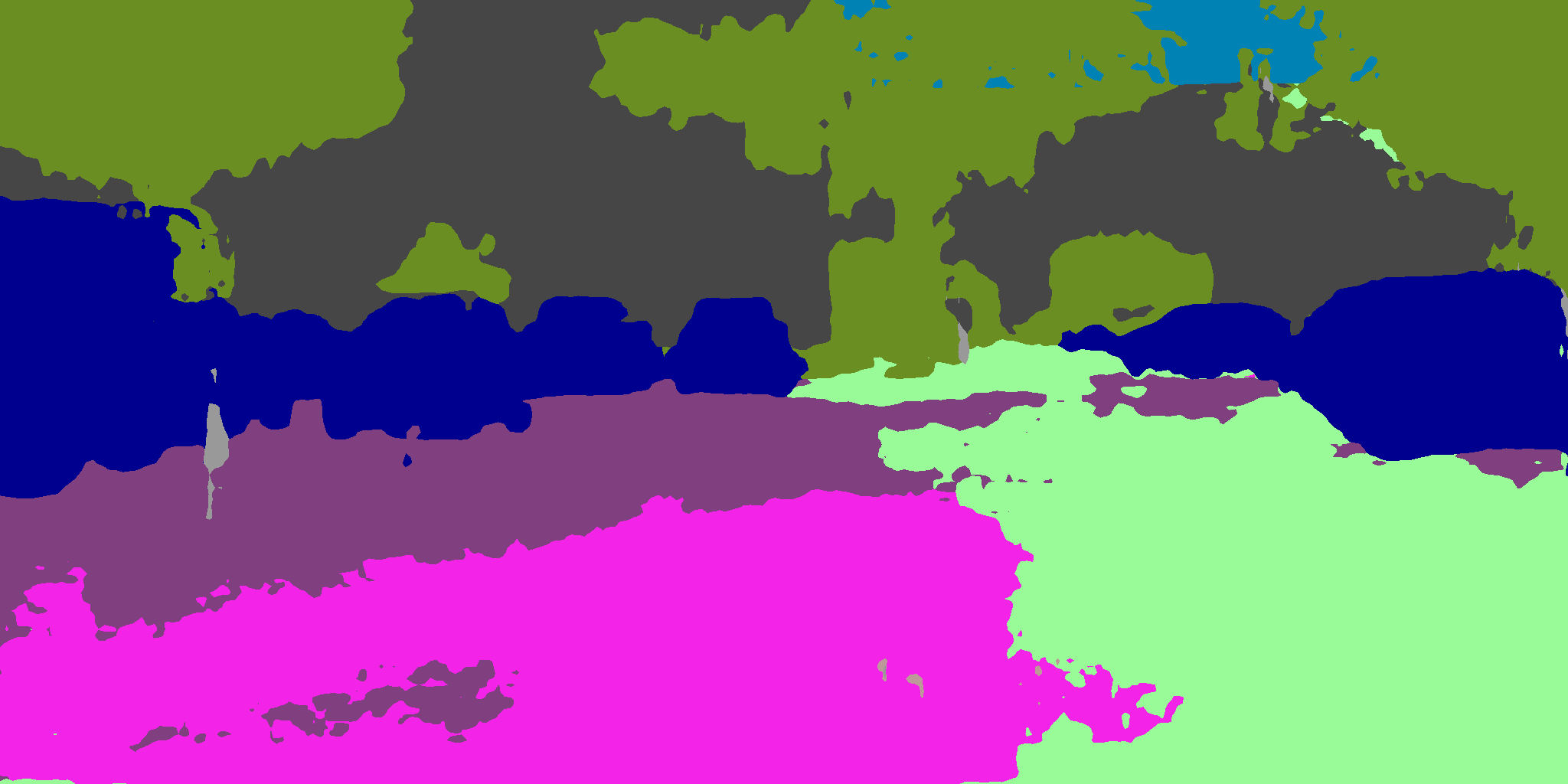}} &
         \subfloat{\includegraphics[width=0.2\linewidth]{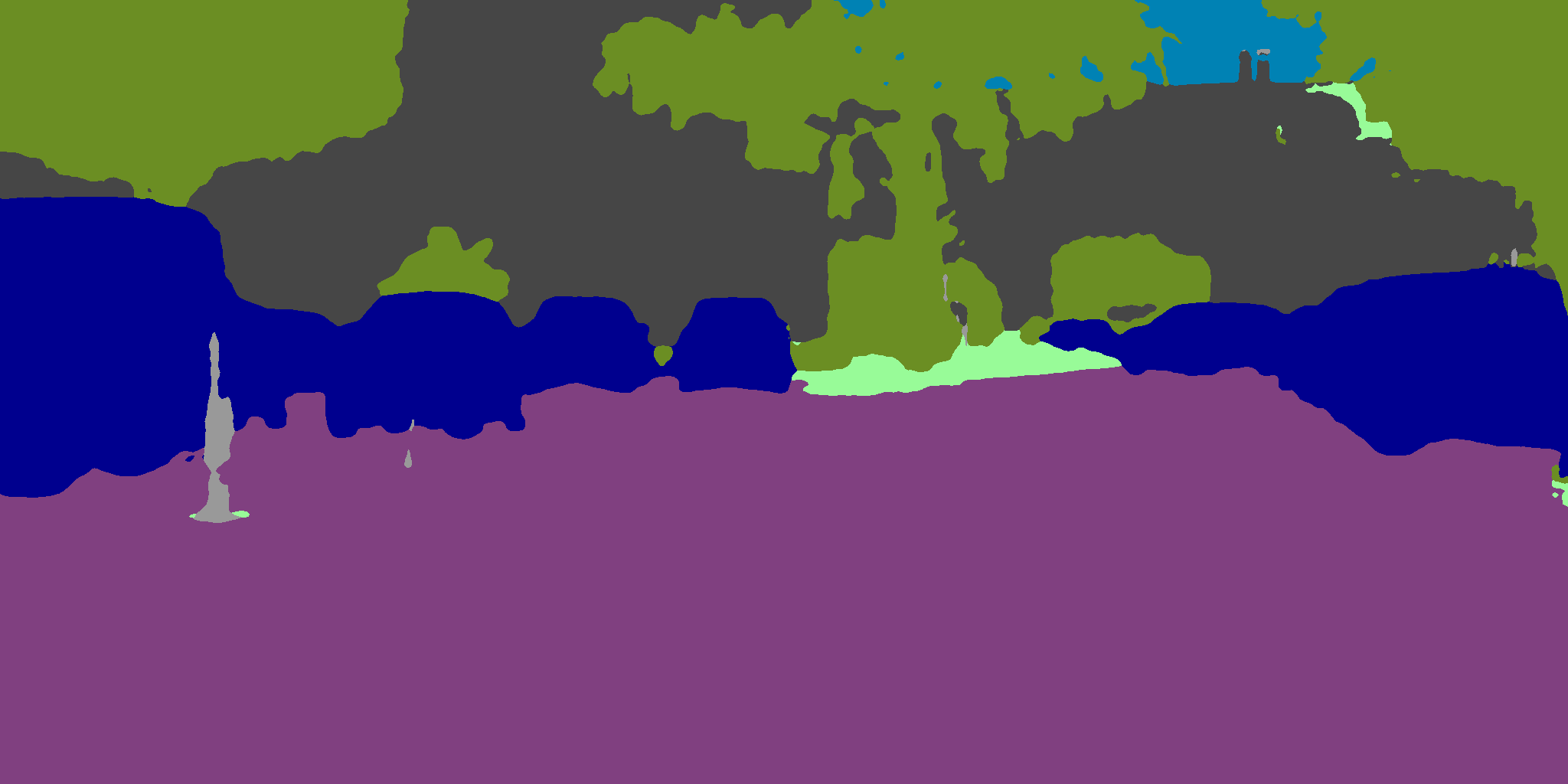}} &
         \subfloat{\includegraphics[width=0.2\linewidth]{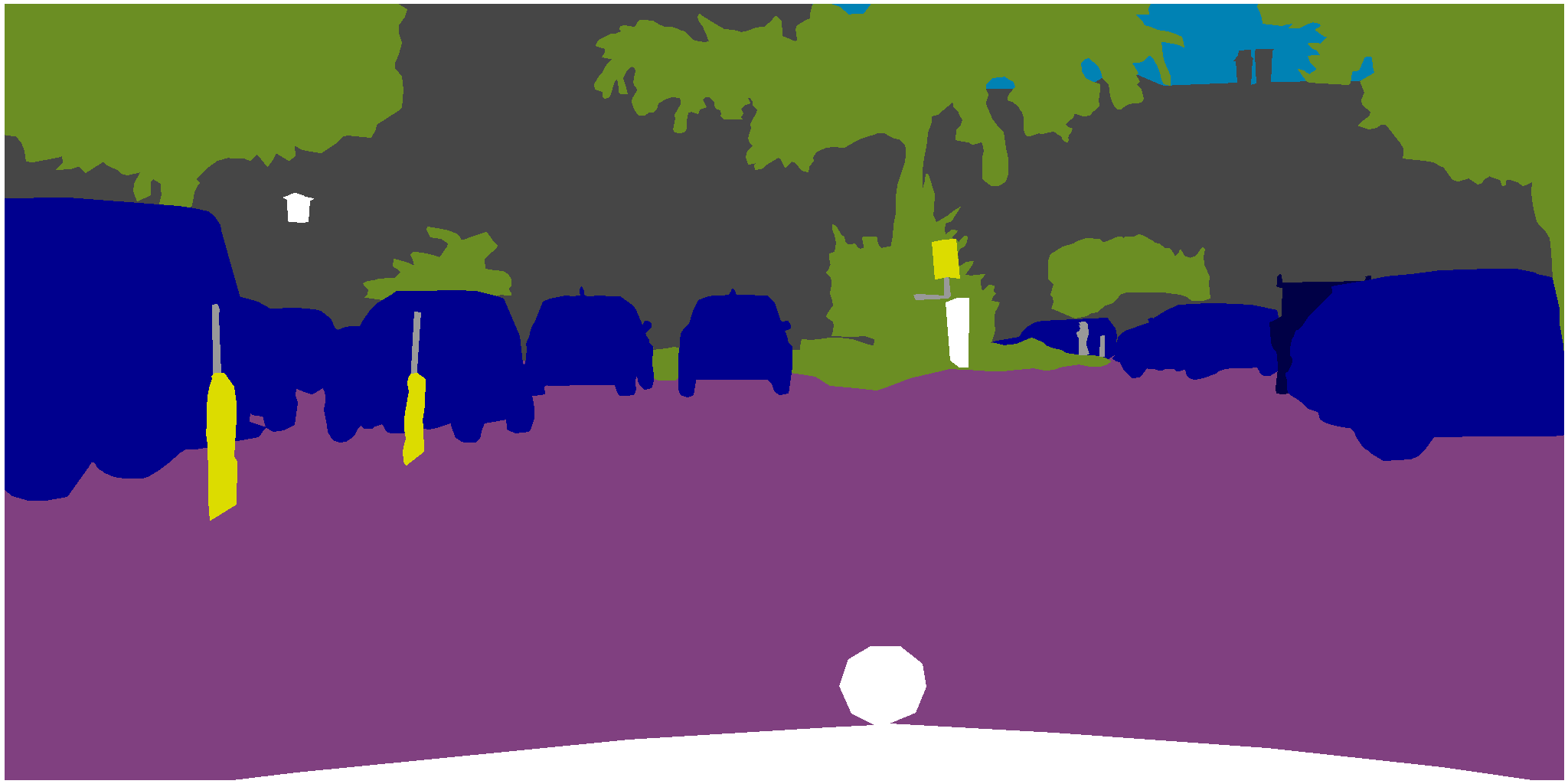}} \\
         
         \subfloat{\includegraphics[width=0.2\linewidth]{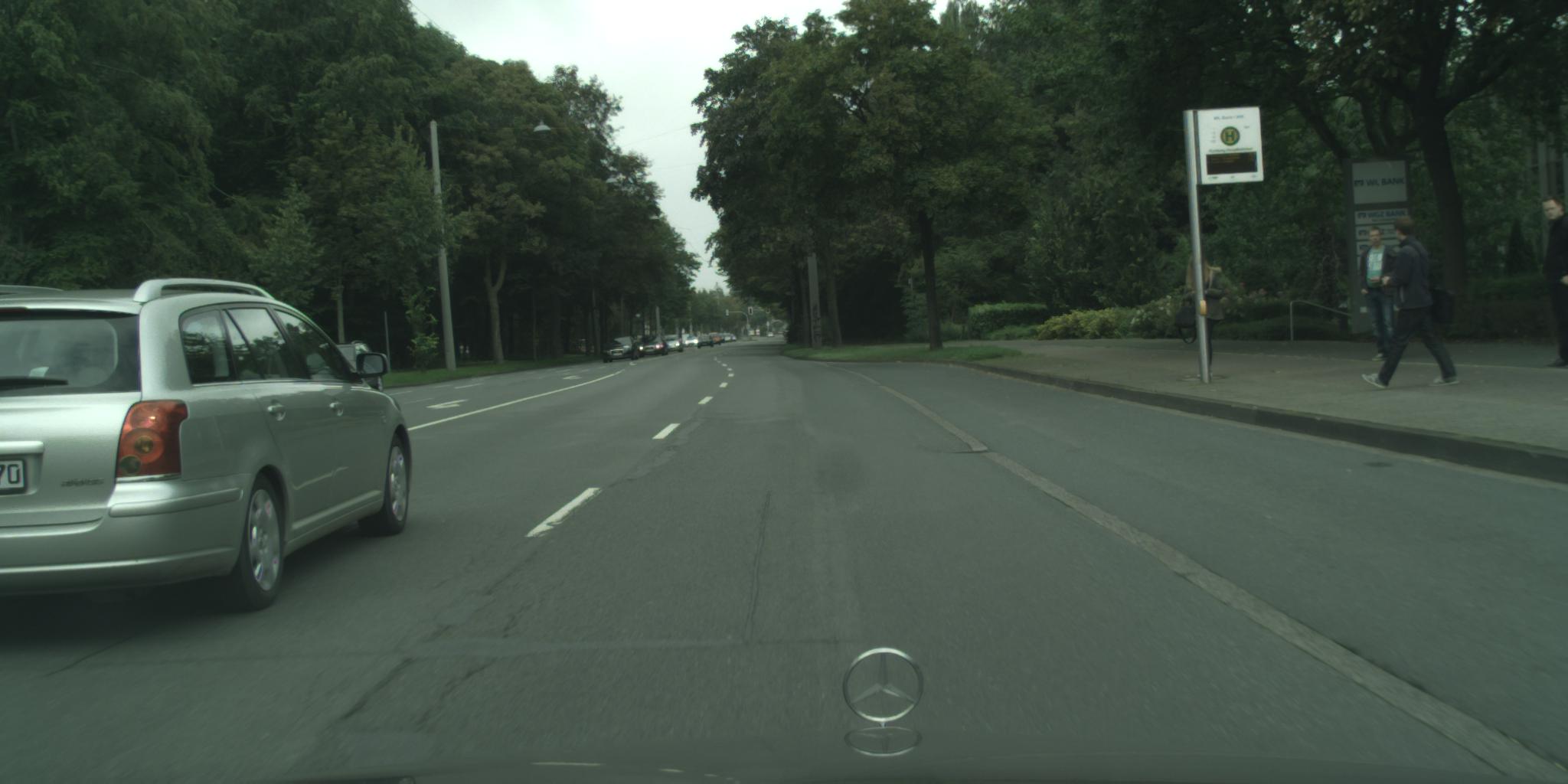}} &
         \subfloat{\includegraphics[width=0.2\linewidth]{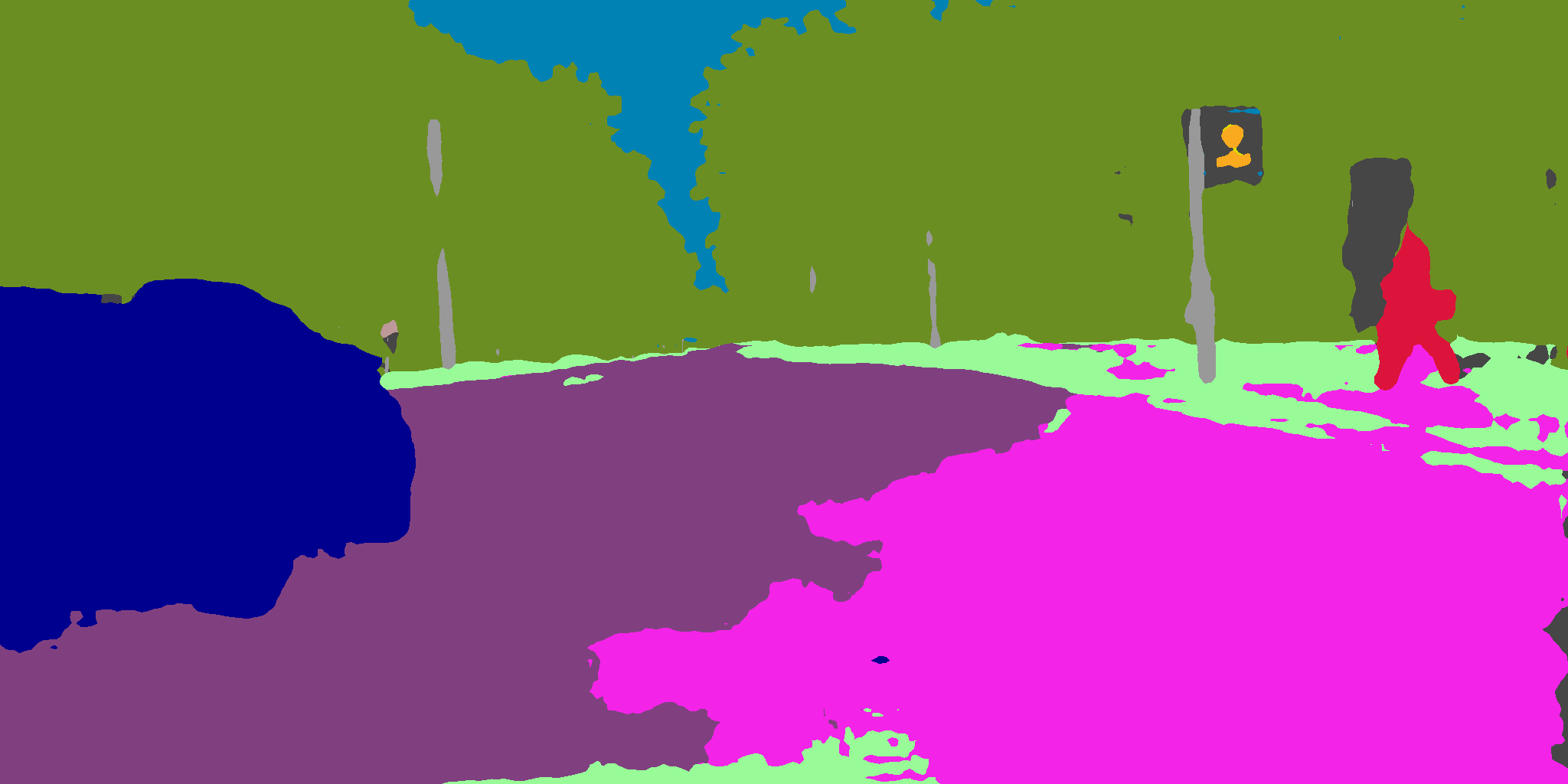}} &
         \subfloat{\includegraphics[width=0.2\linewidth]{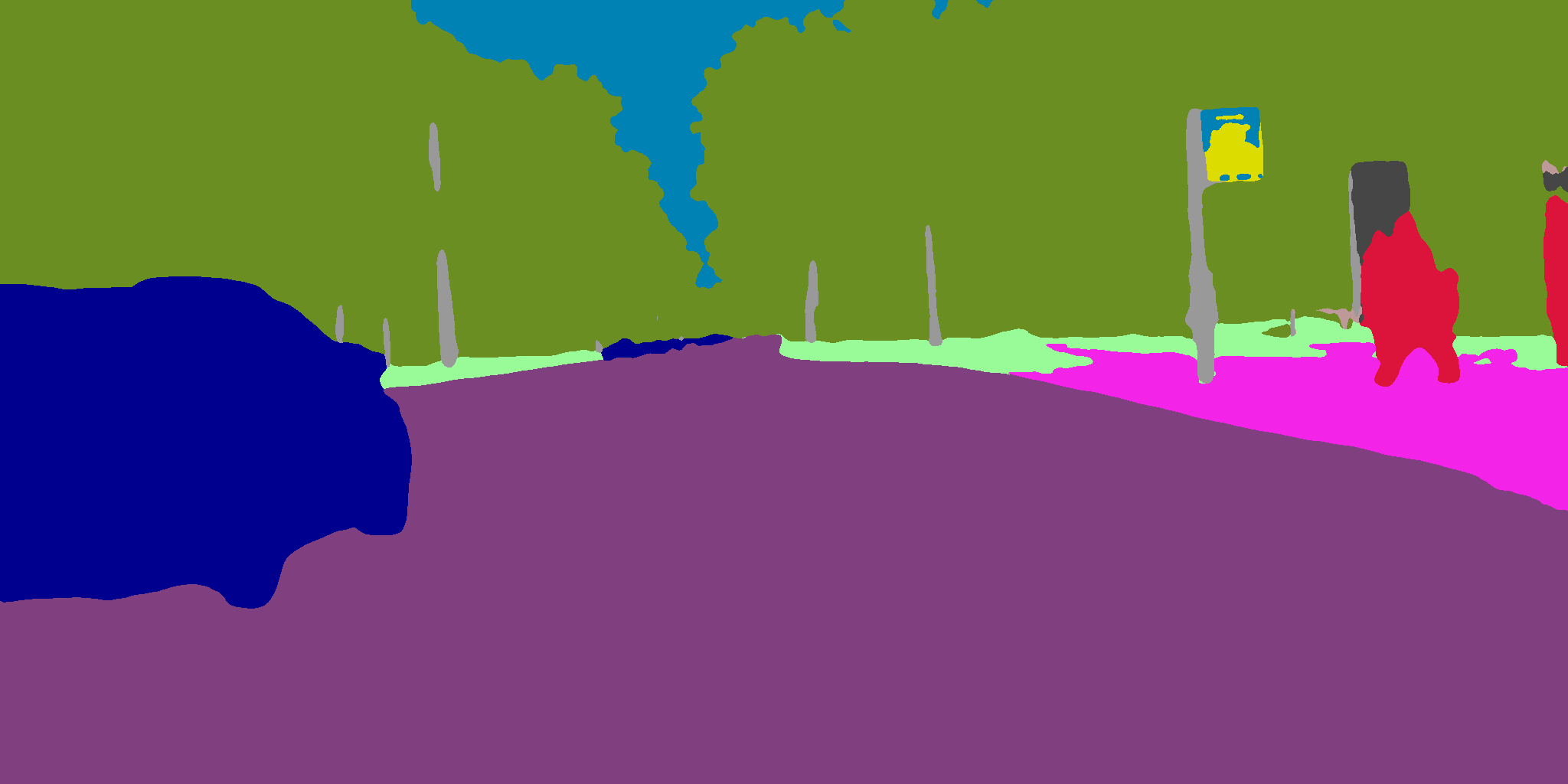}} &
         \subfloat{\includegraphics[width=0.2\linewidth]{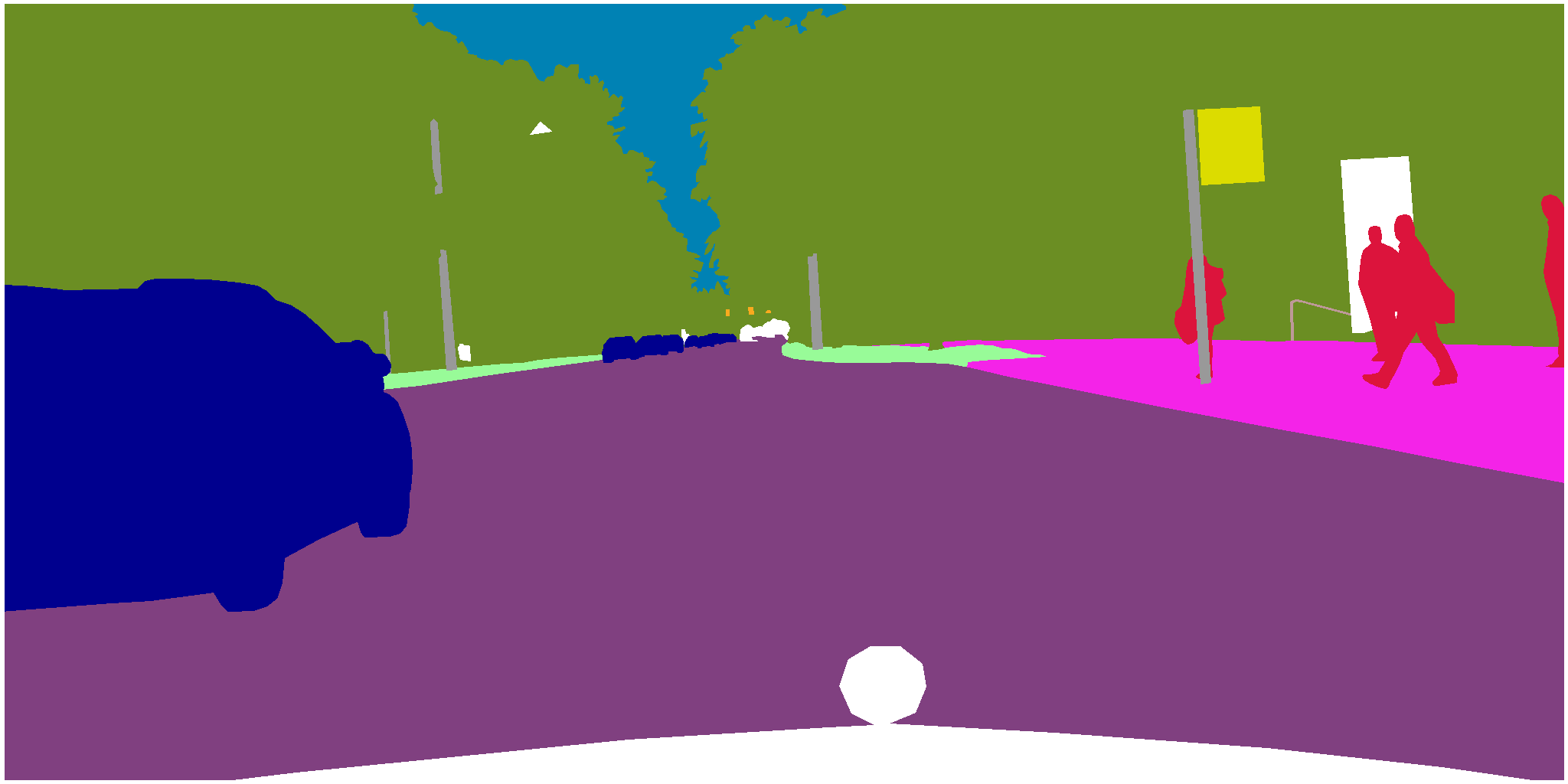}} \\
         
         \subfloat{\includegraphics[width=0.2\linewidth]{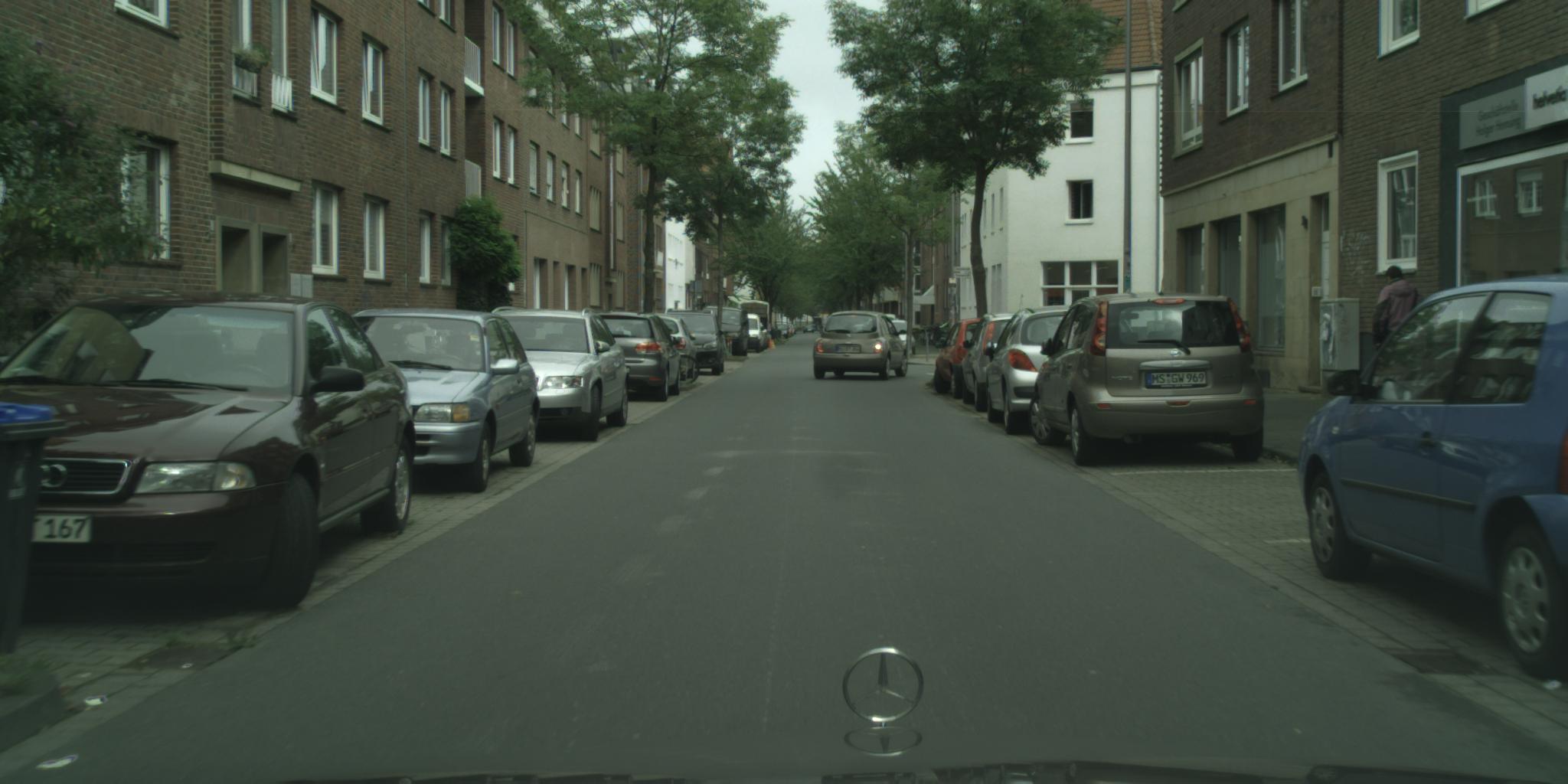}} &
         \subfloat{\includegraphics[width=0.2\linewidth]{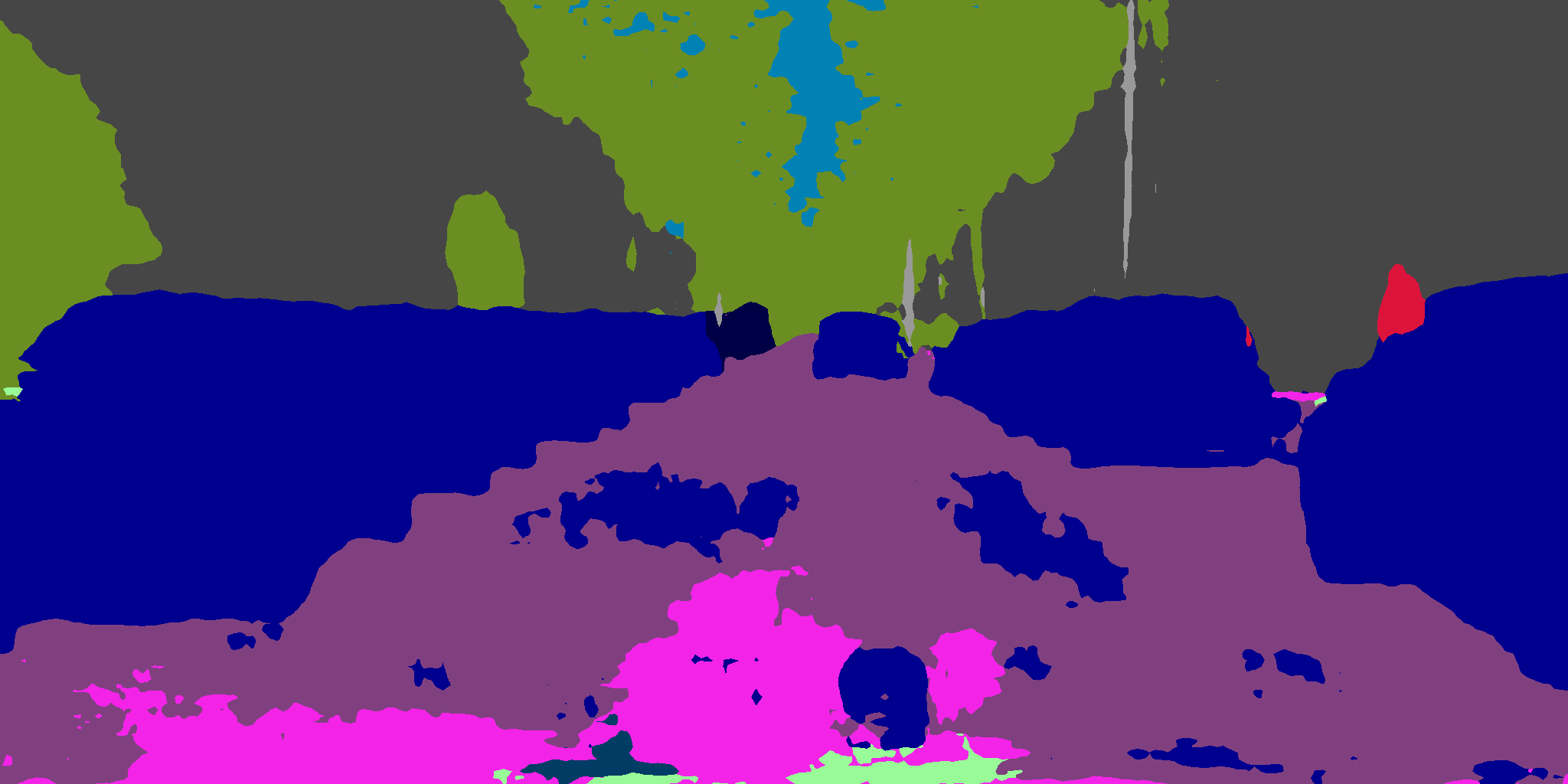}} &
         \subfloat{\includegraphics[width=0.2\linewidth]{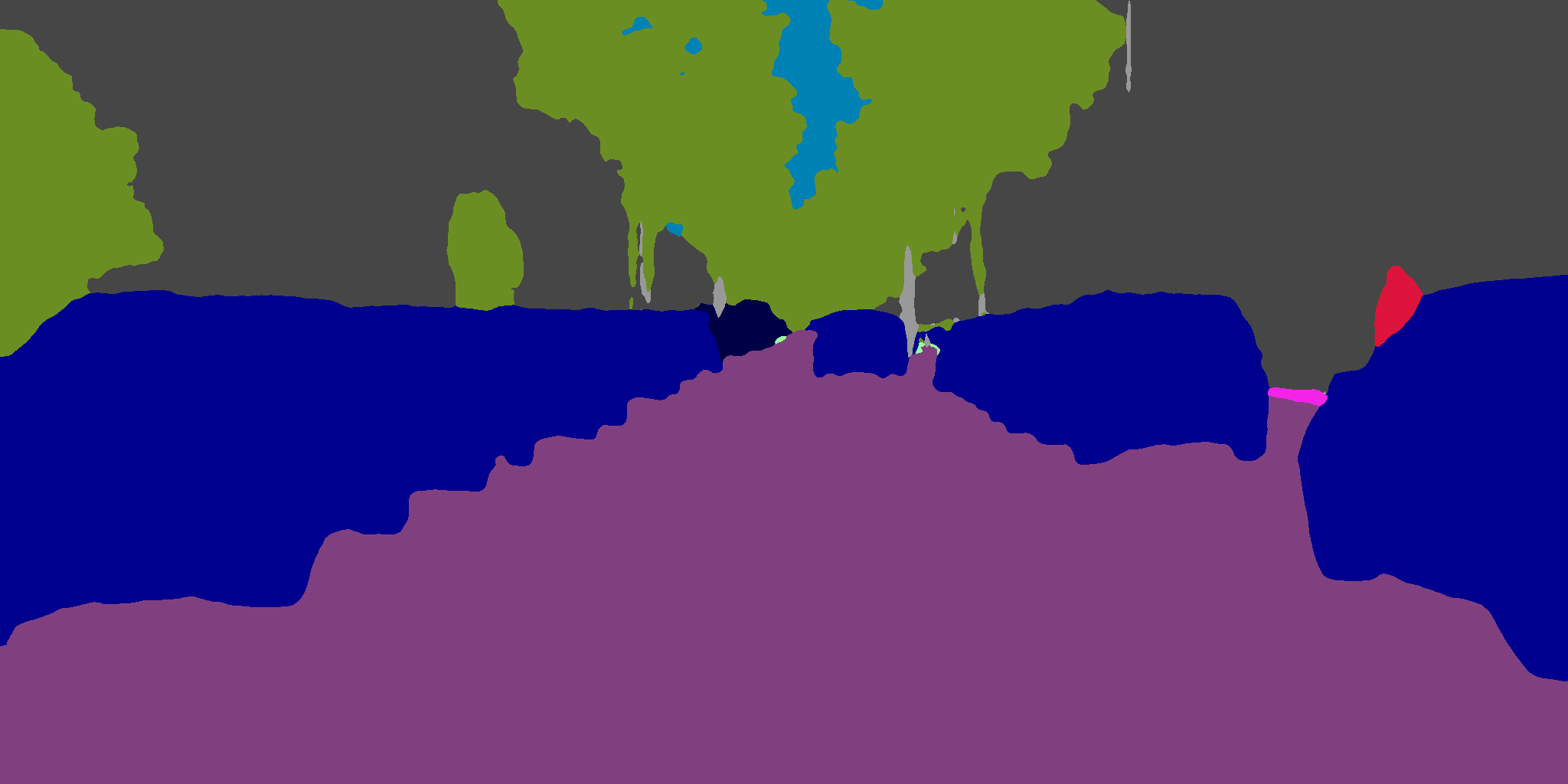}} &
         \subfloat{\includegraphics[width=0.2\linewidth]{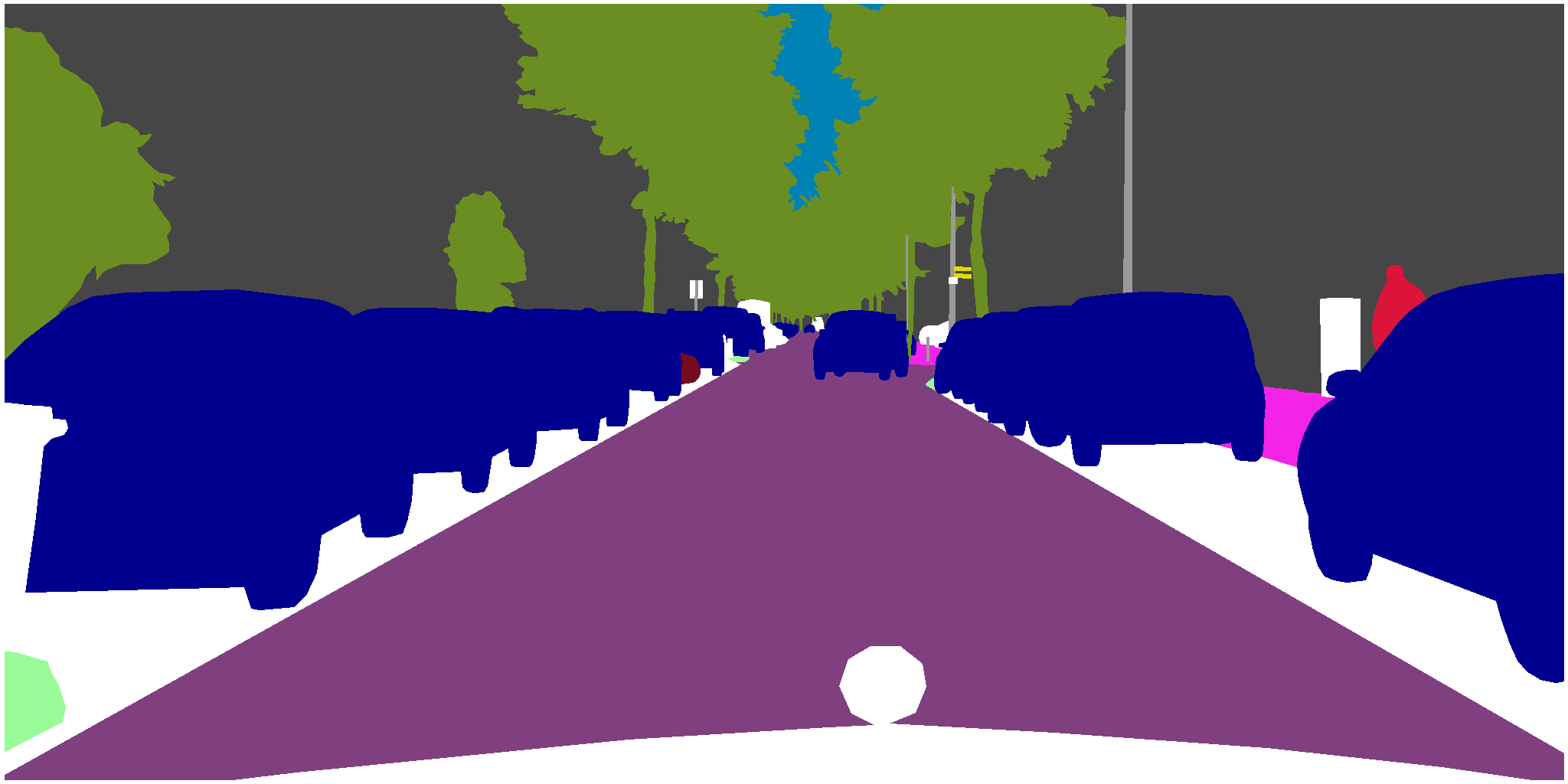}} \\
         
         \subfloat{\includegraphics[width=0.2\linewidth]{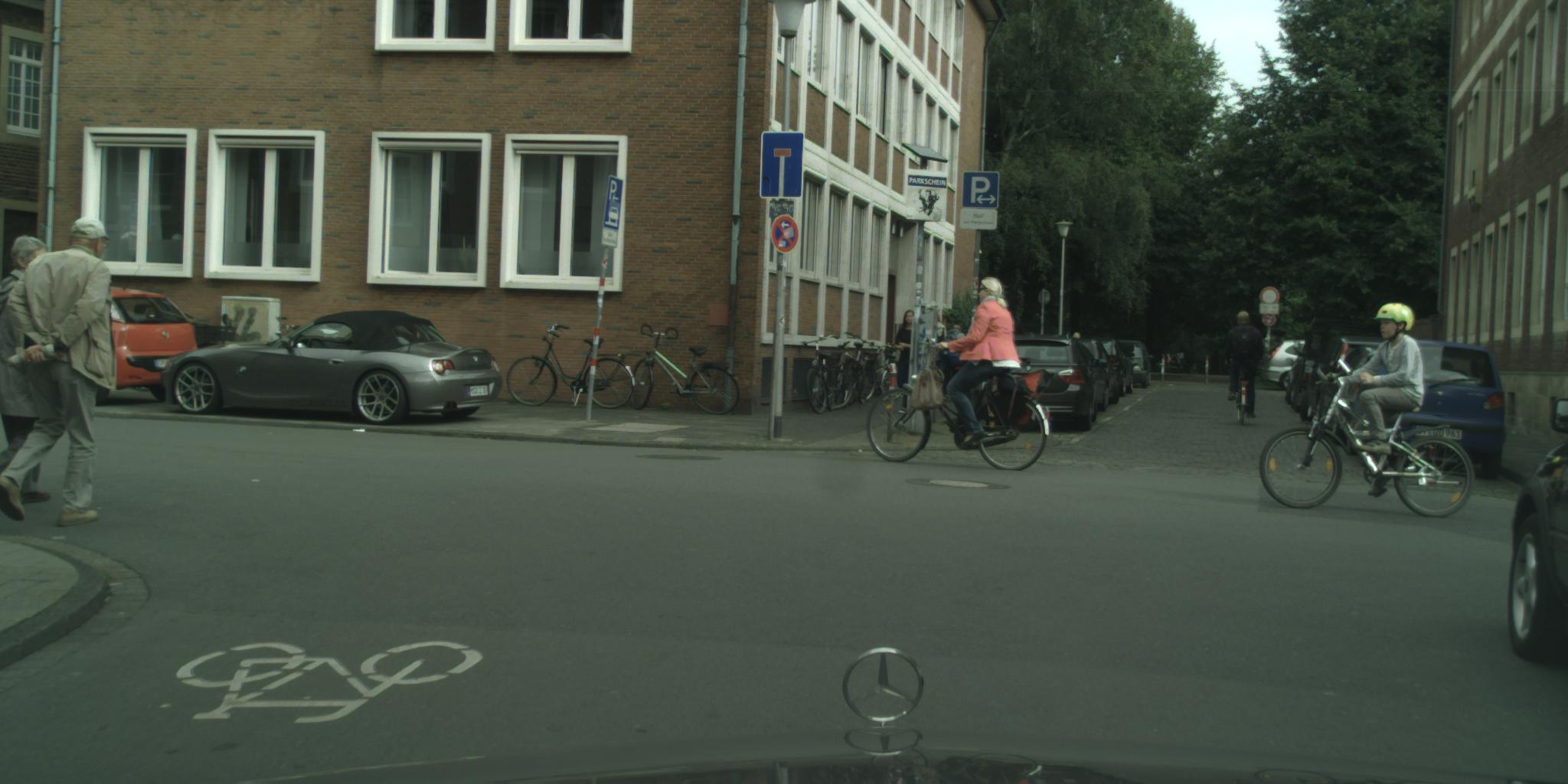}} &
         \subfloat{\includegraphics[width=0.2\linewidth]{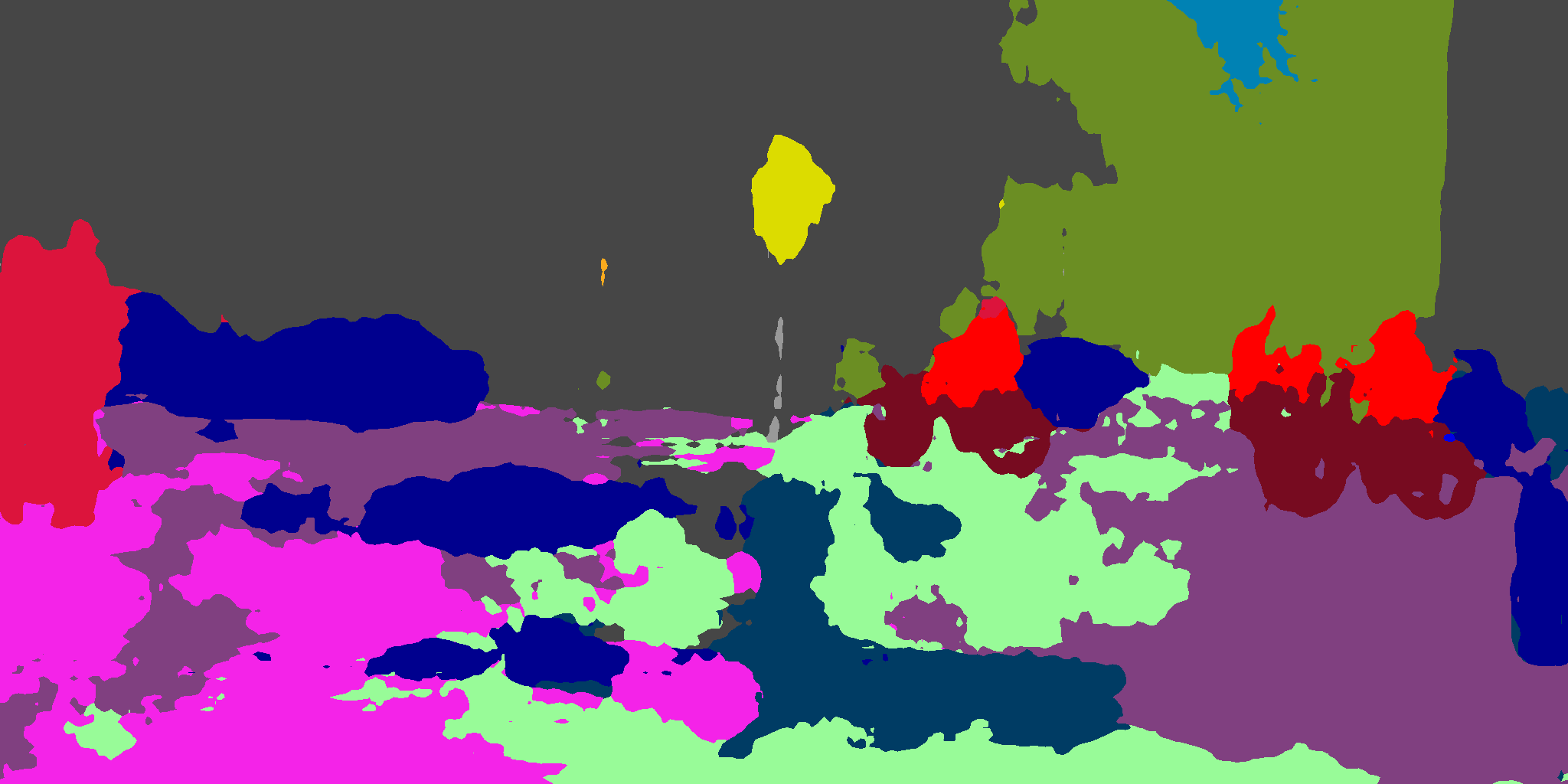}} &
         \subfloat{\includegraphics[width=0.2\linewidth]{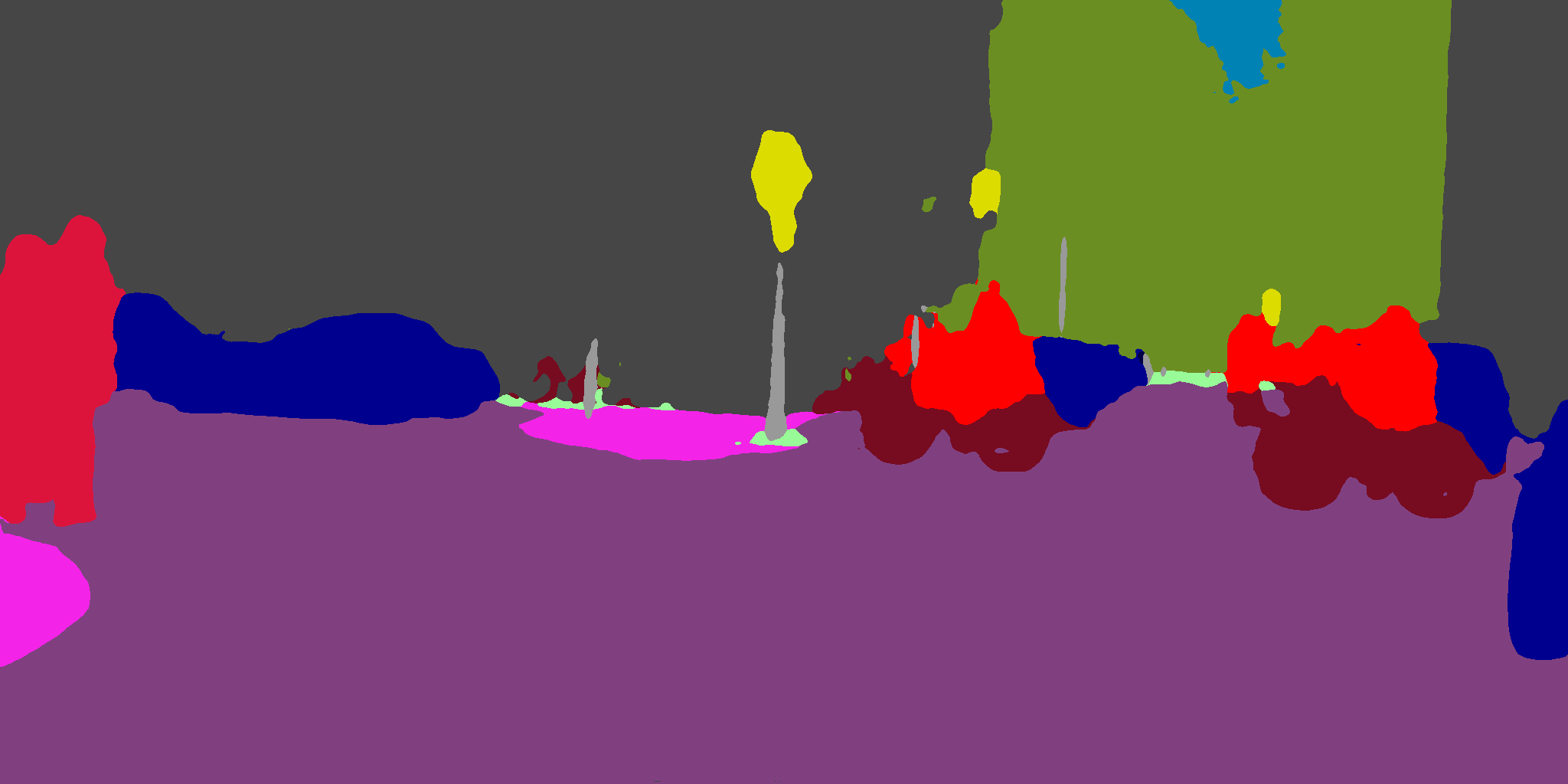}} &
         \subfloat{\includegraphics[width=0.2\linewidth]{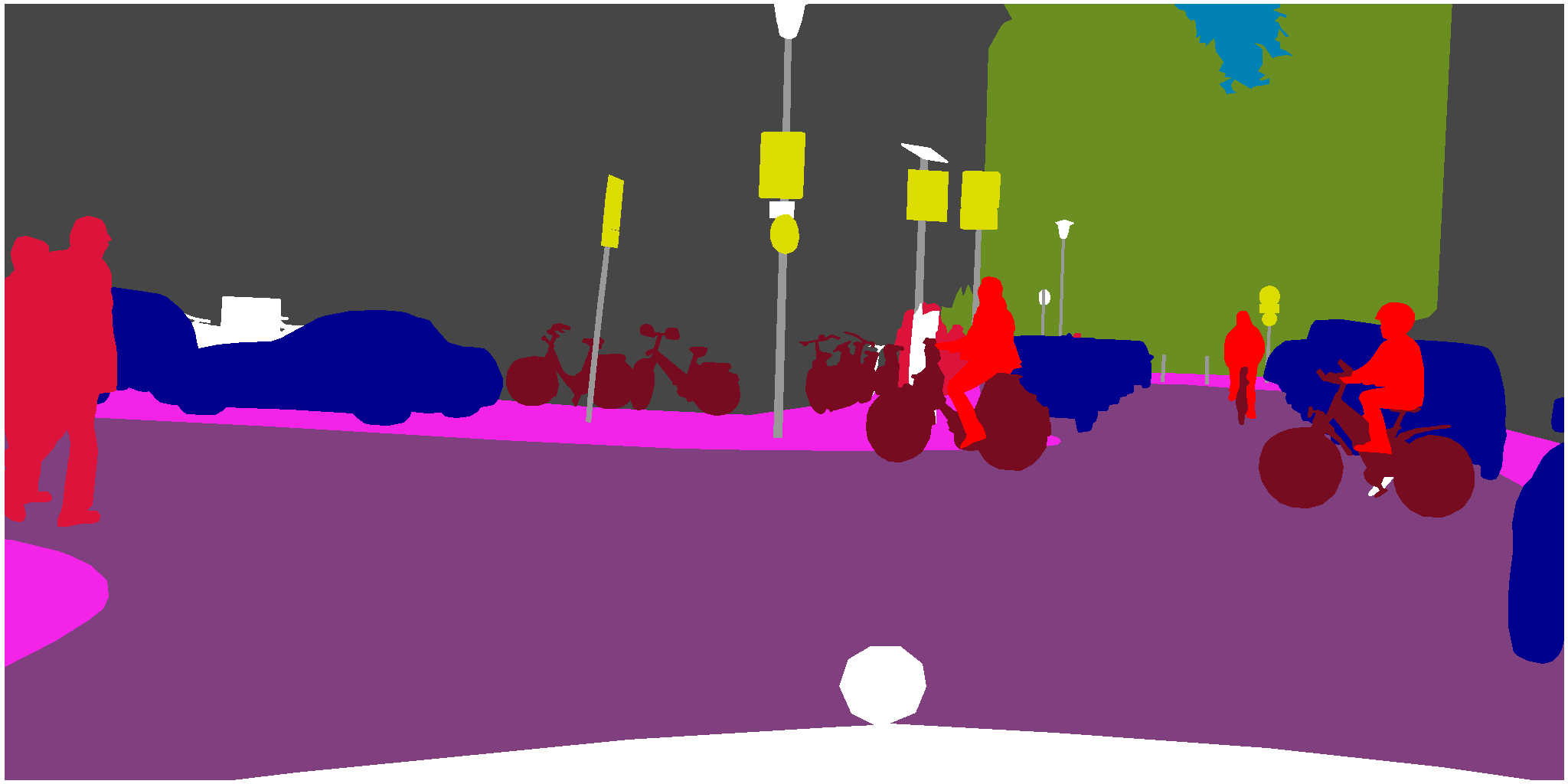}} \\
         
         \subfloat{\includegraphics[width=0.2\linewidth]{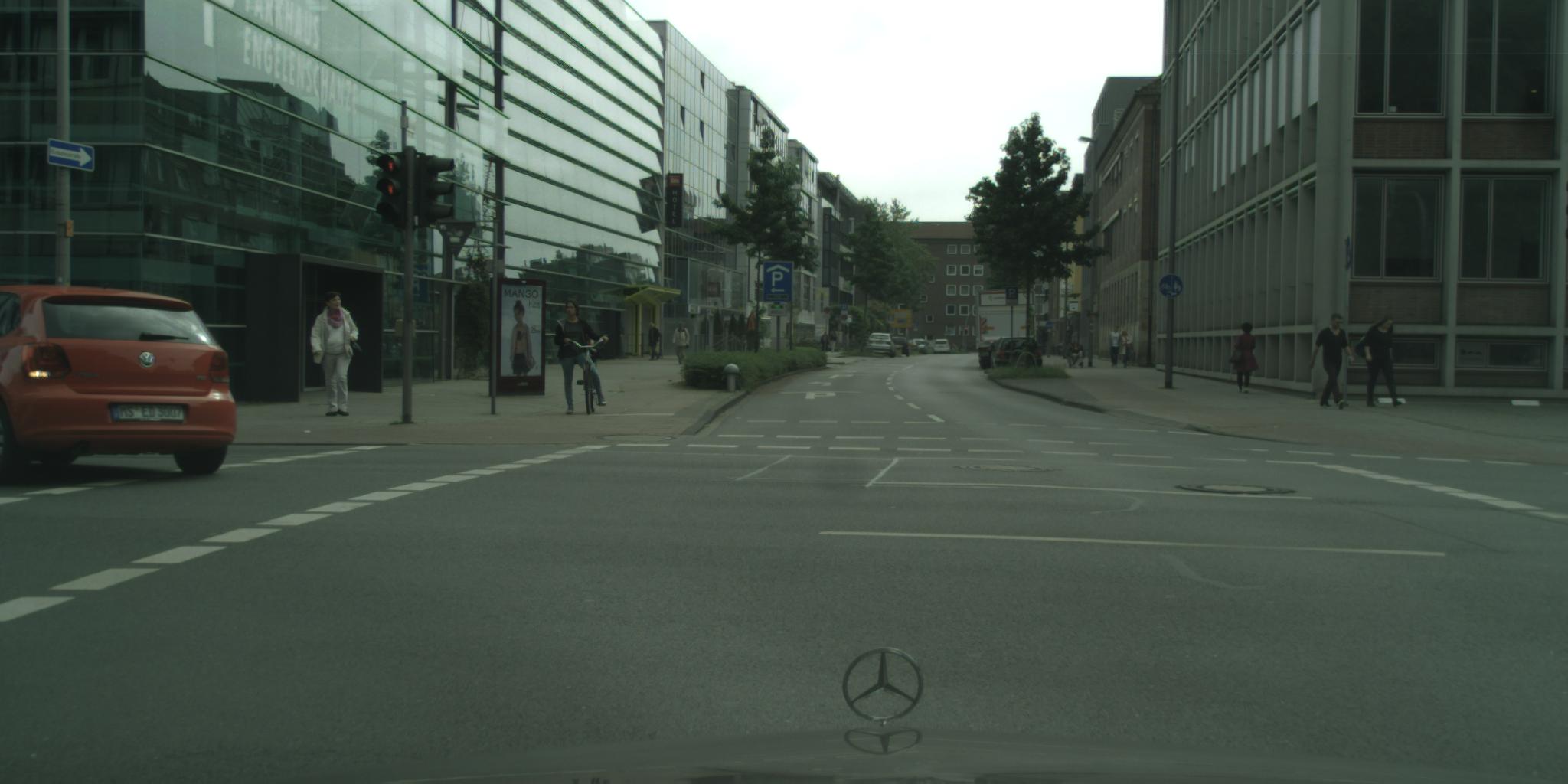}} &
         \subfloat{\includegraphics[width=0.2\linewidth]{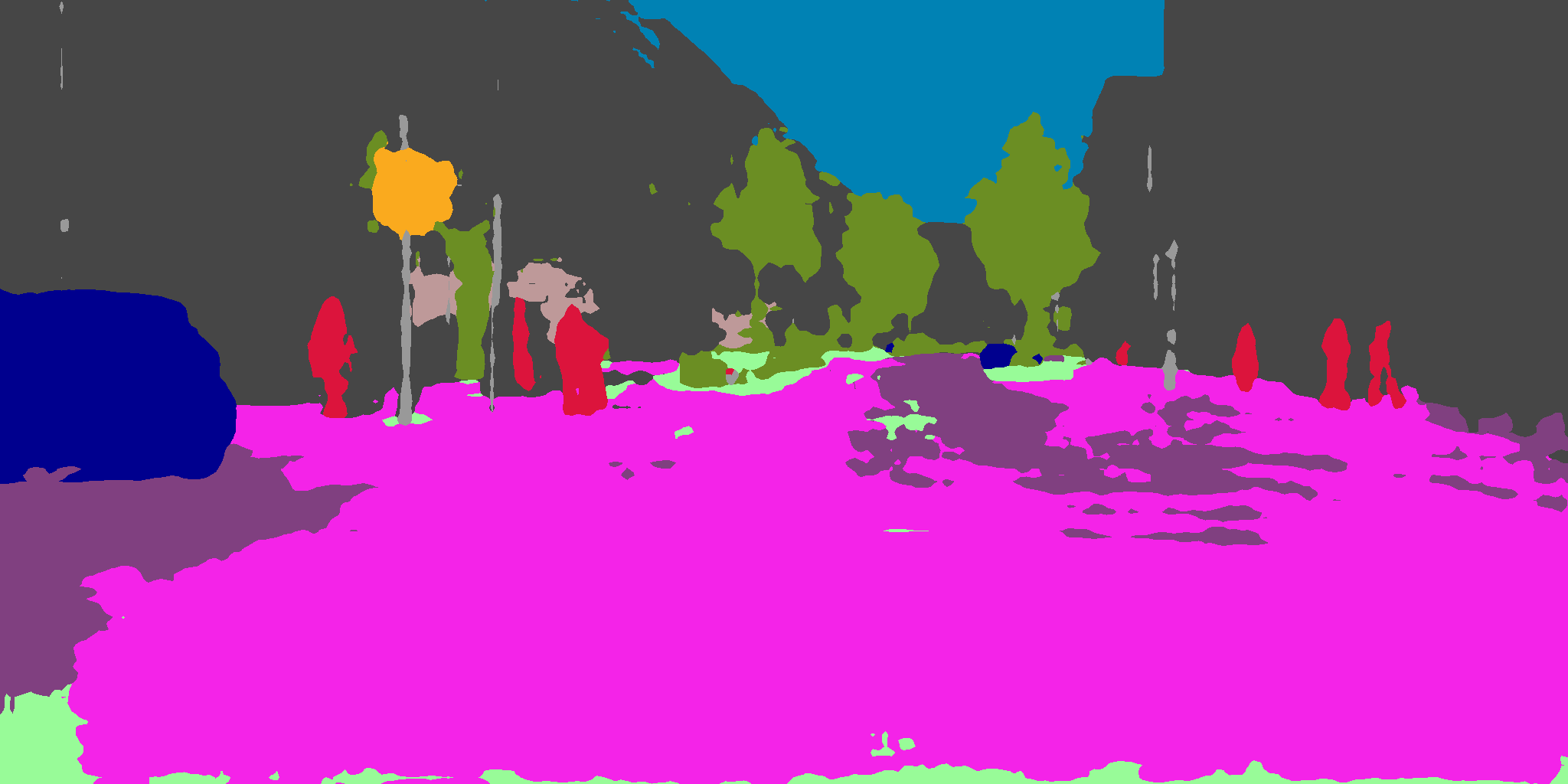}} &
         \subfloat{\includegraphics[width=0.2\linewidth]{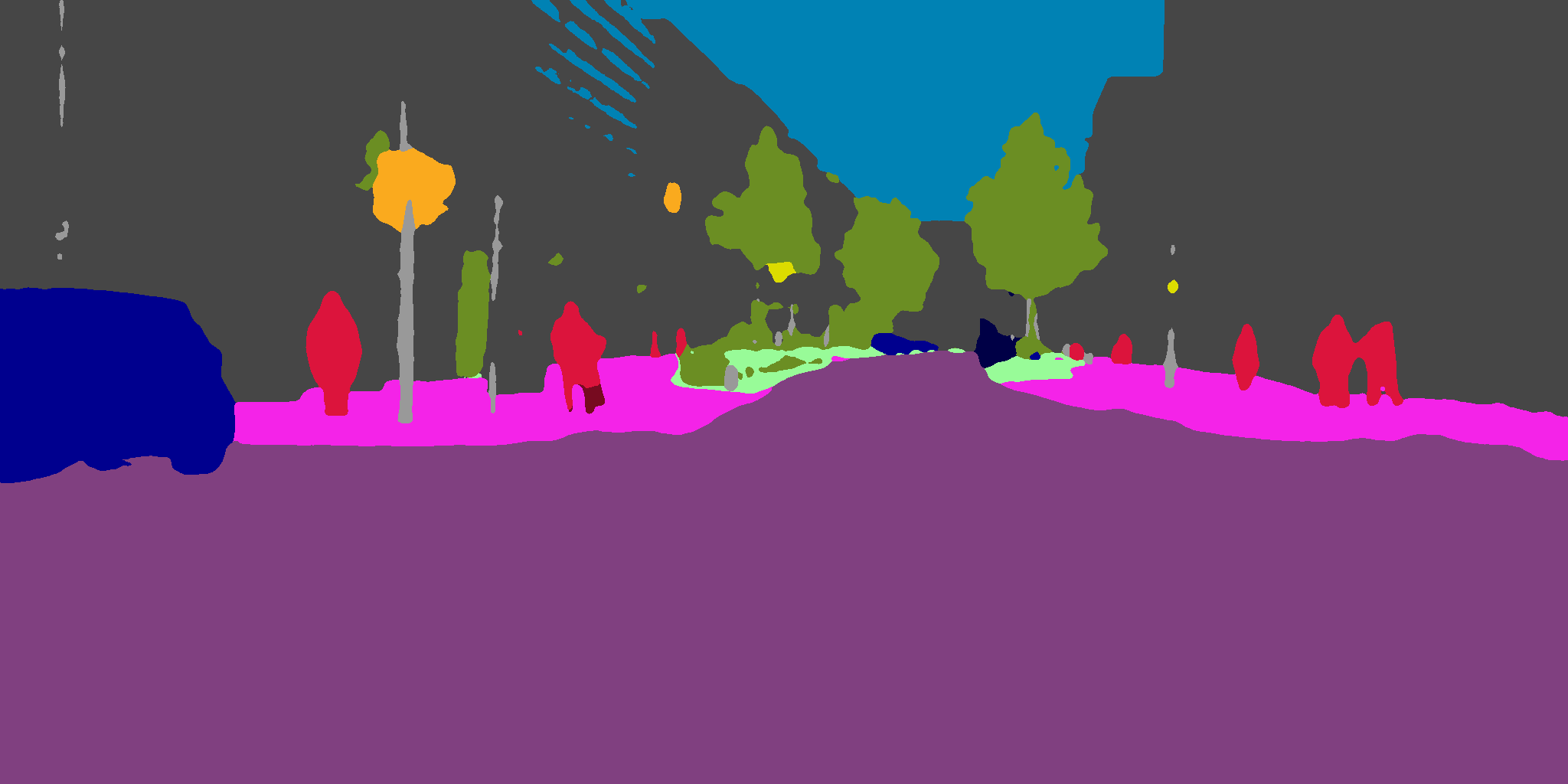}} &
         \subfloat{\includegraphics[width=0.2\linewidth]{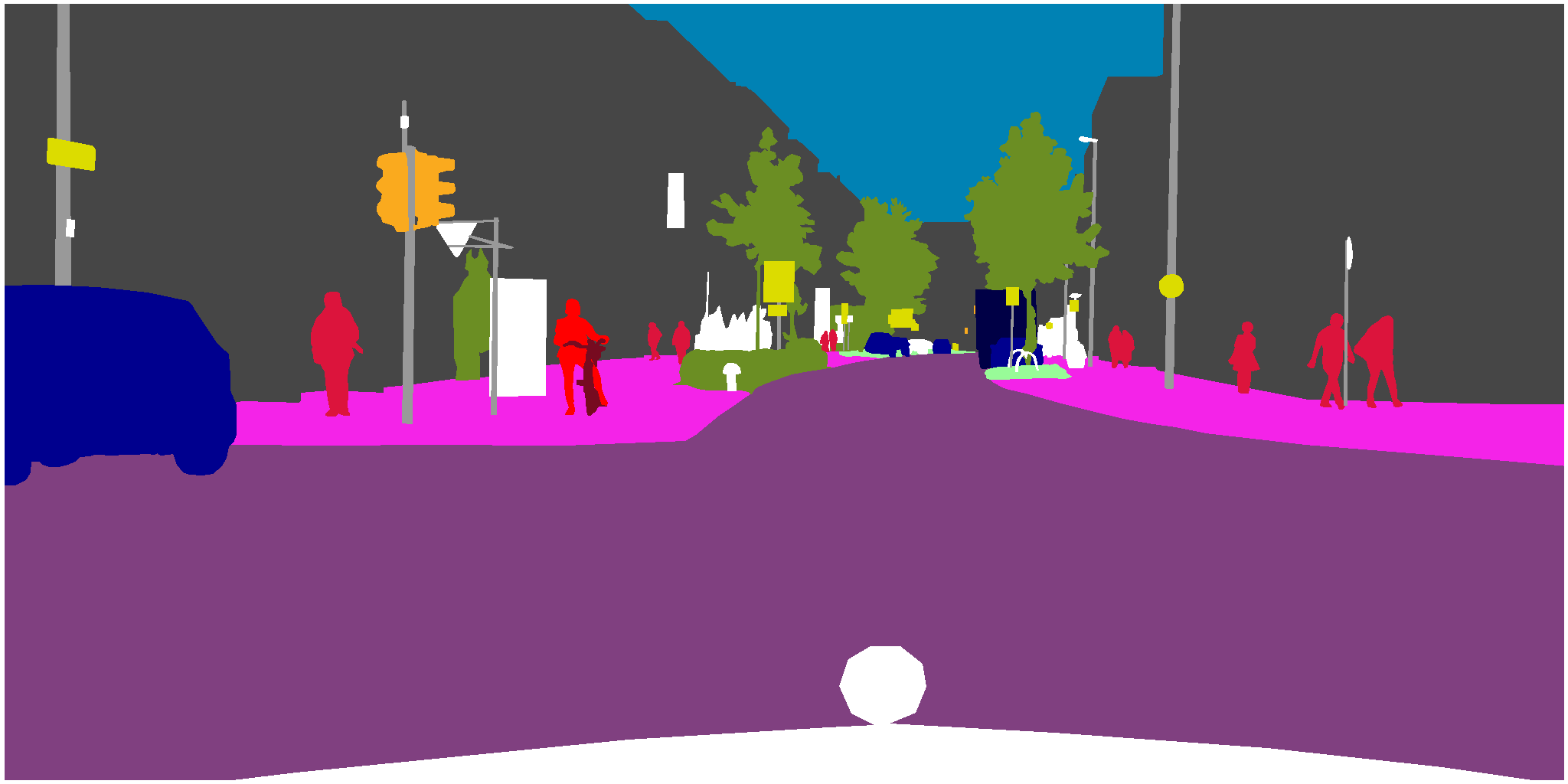}} \\
         
         \subfloat{\includegraphics[width=0.2\linewidth]{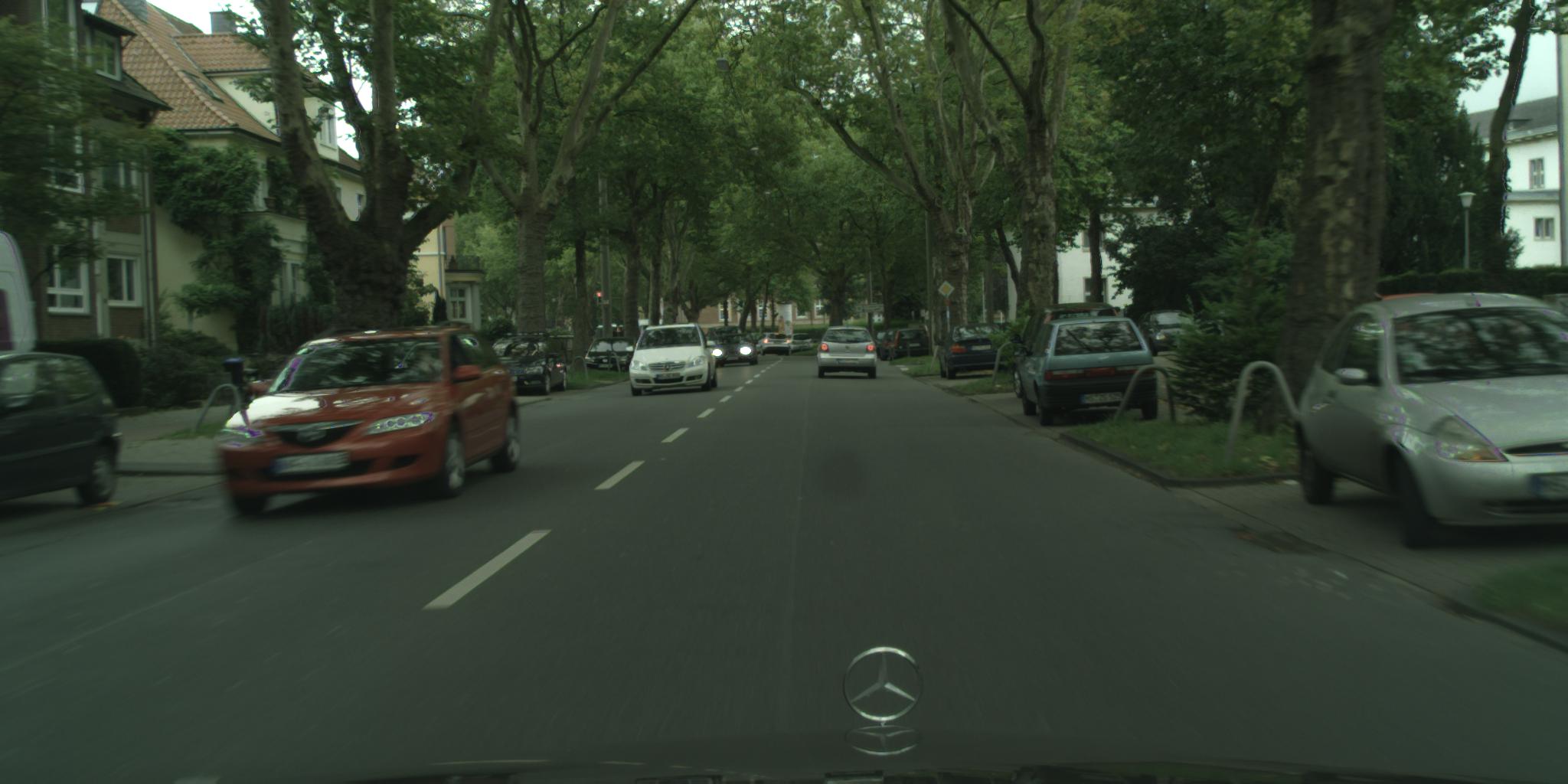}} &
         \subfloat{\includegraphics[width=0.2\linewidth]{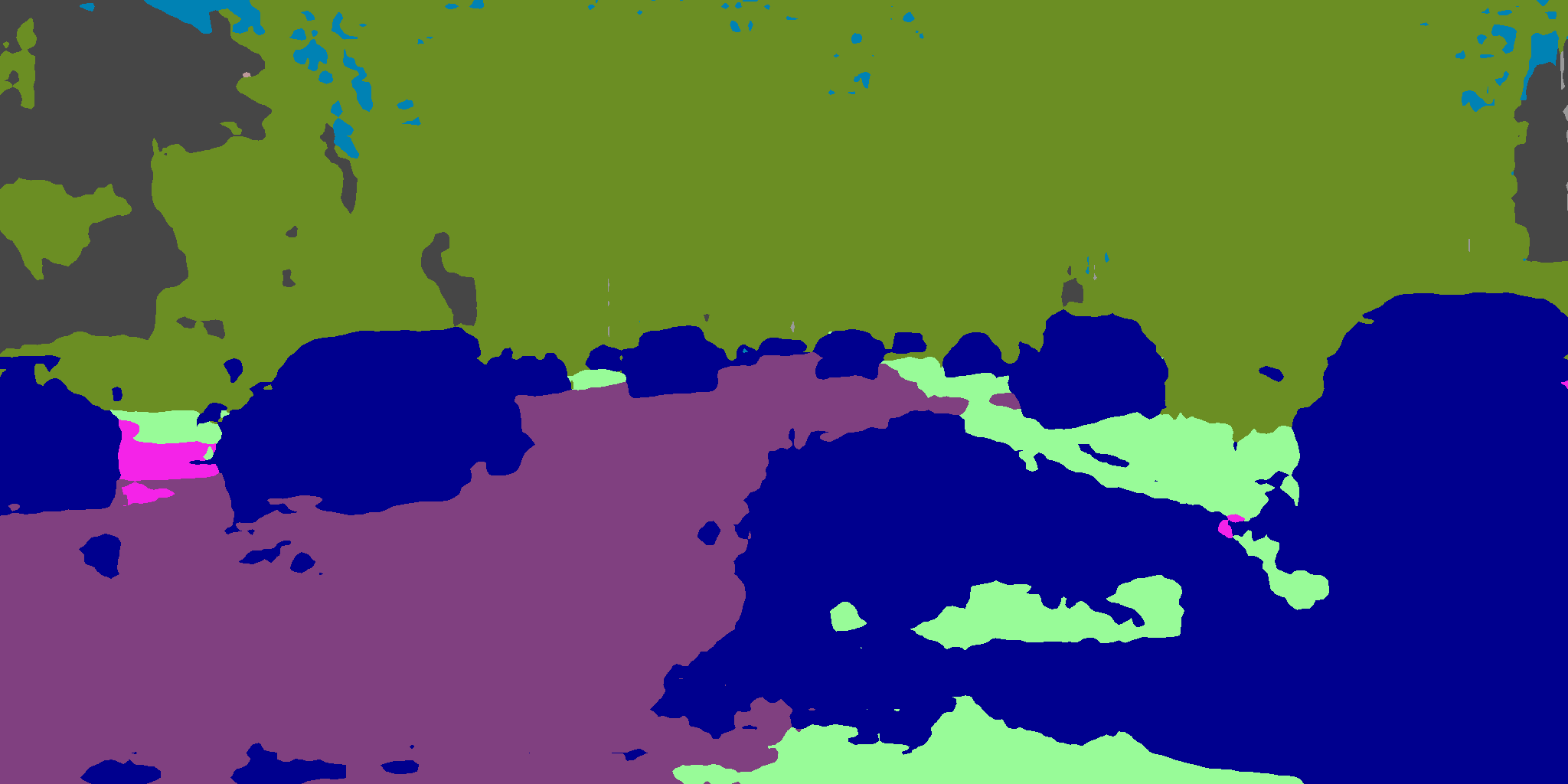}} &
         \subfloat{\includegraphics[width=0.2\linewidth]{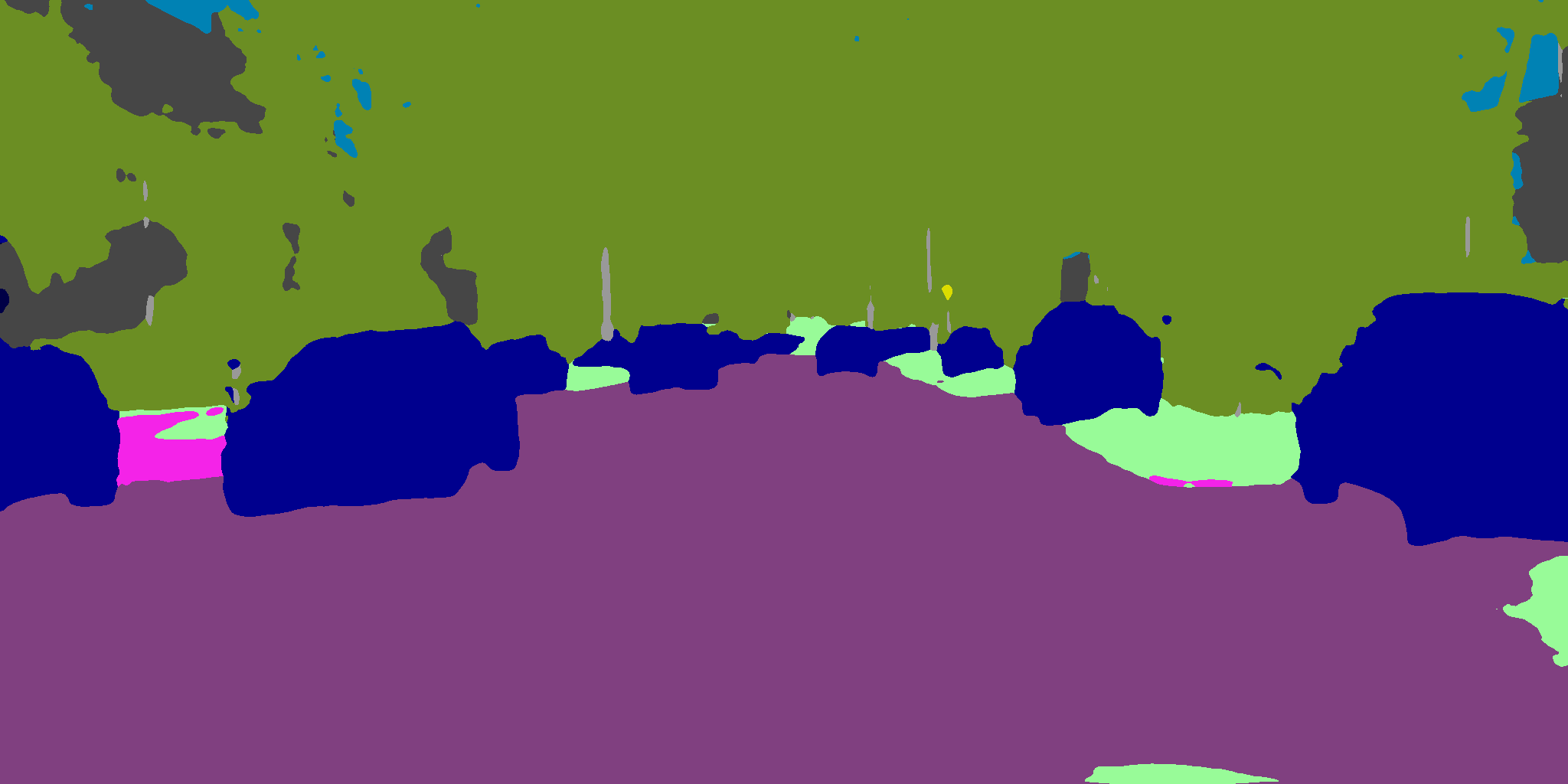}} &
         \subfloat{\includegraphics[width=0.2\linewidth]{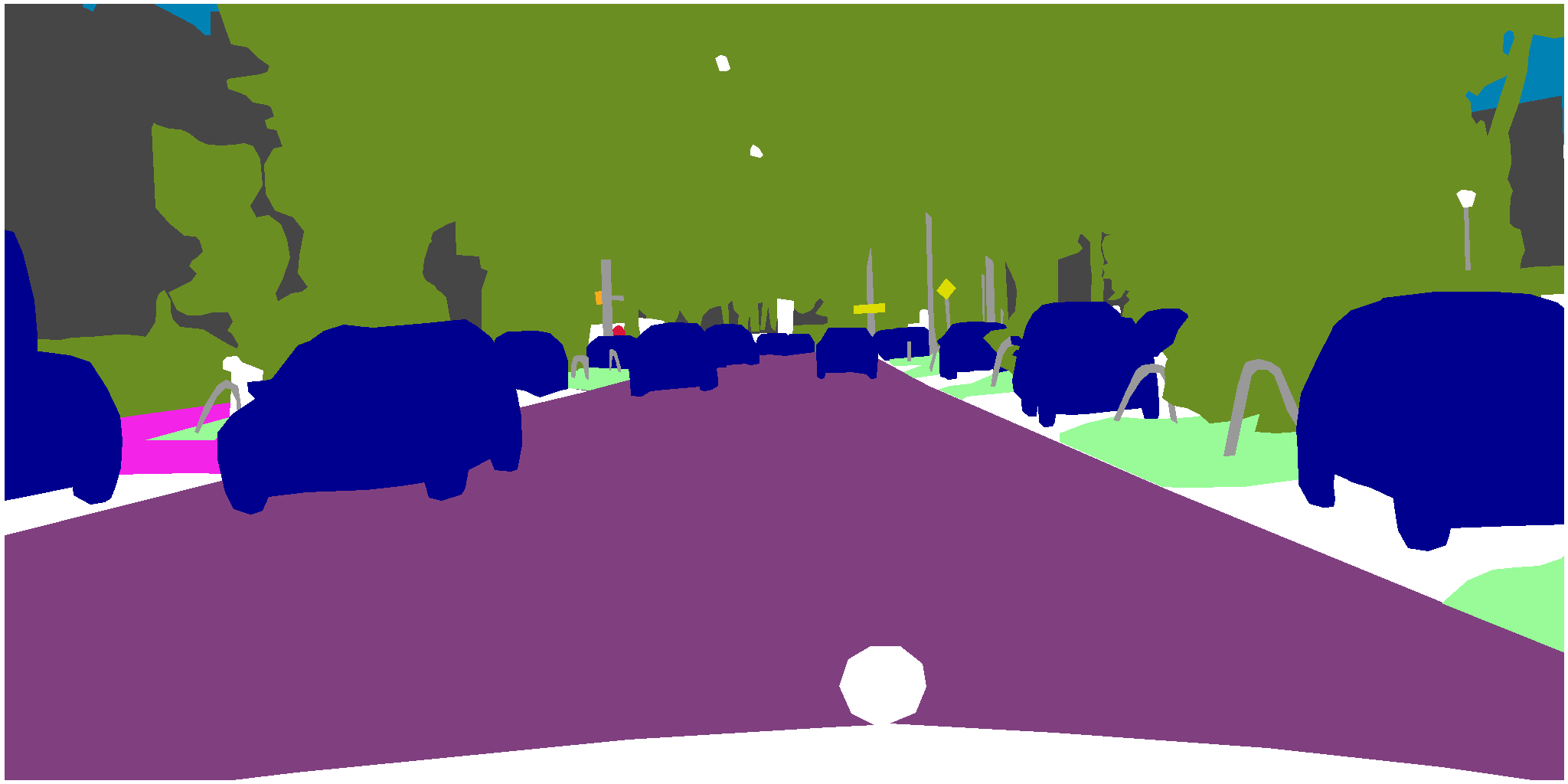}} \\
         
         Image & Before adaptation & After adaptation & Ground-truth
    \end{tabular}}
   \vspace{-0.2cm}
    \caption{More qualitative results of our proposed method.}
    \vspace{-0.5cm}
    \label{fig:examples}
\end{figure*}

After obtaining the high-quality proxy labels, we can train the network for the target domain data in a manner of ``supervised learning''.
As shown in Figure \ref{fig:examples}, the network trained with the generated proxies can produce high-quality semantic labeling results.

\begin{figure*}[htp]
    \centering
    \scalebox{0.99}{
    \begin{tabular}{cccc}
         \subfloat{\includegraphics[width=0.2\linewidth]{figures/proxy_exps/proxy_exp1_im.jpg}} &
         \subfloat{\includegraphics[width=0.2\linewidth]{figures/proxy_exps/proxy_exp1_gt.png}} &
         \subfloat{\includegraphics[width=0.2\linewidth]{figures/proxy_exps/proxy_exp1_ps2.png}} &
         \subfloat{\includegraphics[width=0.2\linewidth]{figures/proxy_exps/proxy_exp1_ps.png}} \\
         \subfloat{\includegraphics[width=0.2\linewidth]{figures/proxy_exps/proxy_exp2_im.jpg}} &
         \subfloat{\includegraphics[width=0.2\linewidth]{figures/proxy_exps/proxy_exp2_gt.png}} &
         \subfloat{\includegraphics[width=0.2\linewidth]{figures/proxy_exps/proxy_exp2_ps2.png}} &
         \subfloat{\includegraphics[width=0.2\linewidth]{figures/proxy_exps/proxy_exp2_ps.png}} \\
         \subfloat{\includegraphics[width=0.2\linewidth]{figures/proxy_exps/proxy_exp3_im.jpg}} &
         \subfloat{\includegraphics[width=0.2\linewidth]{figures/proxy_exps/proxy_exp3_gt.png}} &
         \subfloat{\includegraphics[width=0.2\linewidth]{figures/proxy_exps/proxy_exp3_ps2.png}} &
         \subfloat{\includegraphics[width=0.2\linewidth]{figures/proxy_exps/proxy_exp3_ps.png}} \\
         \subfloat{\includegraphics[width=0.2\linewidth]{figures/proxy_exps/proxy_exp4_im.jpg}} &
         \subfloat{\includegraphics[width=0.2\linewidth]{figures/proxy_exps/proxy_exp4_gt.png}} &
         \subfloat{\includegraphics[width=0.2\linewidth]{figures/proxy_exps/proxy_exp4_ps2.png}} &
         \subfloat{\includegraphics[width=0.2\linewidth]{figures/proxy_exps/proxy_exp4_ps.png}} \\
         \subfloat{\includegraphics[width=0.2\linewidth]{figures/proxy_exps/proxy_exp5_im.jpg}} &
         \subfloat{\includegraphics[width=0.2\linewidth]{figures/proxy_exps/proxy_exp5_gt.png}} &
         \subfloat{\includegraphics[width=0.2\linewidth]{figures/proxy_exps/proxy_exp5_ps2.png}} &
         \subfloat{\includegraphics[width=0.2\linewidth]{figures/proxy_exps/proxy_exp5_ps.png}} \\
         \subfloat{\includegraphics[width=0.2\linewidth]{figures/proxy_exps/proxy_exp6_im.jpg}} &
         \subfloat{\includegraphics[width=0.2\linewidth]{figures/proxy_exps/proxy_exp6_gt.png}} &
         \subfloat{\includegraphics[width=0.2\linewidth]{figures/proxy_exps/proxy_exp6_ps2.png}} &
         \subfloat{\includegraphics[width=0.2\linewidth]{figures/proxy_exps/proxy_exp6_ps.png}} \\
         \subfloat{\includegraphics[width=0.2\linewidth]{figures/proxy_exps/proxy_exp7_im.jpg}} &
         \subfloat{\includegraphics[width=0.2\linewidth]{figures/proxy_exps/proxy_exp7_gt.png}} &
         \subfloat{\includegraphics[width=0.2\linewidth]{figures/proxy_exps/proxy_exp7_ps2.png}} &
         \subfloat{\includegraphics[width=0.2\linewidth]{figures/proxy_exps/proxy_exp7_ps.png}} \\
         
         Image from target domain & Ground-truth & Proxy w/o Adv-conf & Proxy with Adv-conf
    \end{tabular}}
   \vspace{-0.2cm}
    \caption{Examples of proxies generated with and without adversarial confidence. In the label maps, the ignored samples are marked with white color.}
    \vspace{-0.5cm}
    \label{fig:proxy_exps}
\end{figure*}

